\documentclass[10pt,journal,compsoc]{IEEEtran}
\usepackage[colorlinks]{hyperref}
\usepackage{booktabs}
\usepackage[ruled]{algorithm2e}
\usepackage{mathptmx}
\usepackage{graphicx}
\usepackage{times}
\usepackage{epsfig}
\usepackage{epstopdf}
\usepackage{amsmath}
\usepackage{amssymb}
\usepackage{subfigure}
\usepackage{caption}
\usepackage{tabularx}
\usepackage[table]{xcolor}
\usepackage[normalem]{ulem}
\usepackage[noend]{algpseudocode}
\usepackage{booktabs}
\usepackage{multirow}
\usepackage{lipsum}
\usepackage{float}
\usepackage[cmintegrals]{newtxmath}
\usepackage{graphicx}
\usepackage{color}
\usepackage{cases}
\newcommand\etal{et~al. }
\newcommand\eg{e.g. }
\newcommand\ie{i.e. }
\newcommand\wrt{w.r.t. }%
\newcolumntype{L}[1]{>{\raggedright\arraybackslash}p{#1}}
\newcolumntype{C}[1]{>{\centering\arraybackslash}p{#1}}
\newcolumntype{R}[1]{>{\raggedleft\arraybackslash}p{#1}}

\begin{document}

\title{Deep Richardson-Lucy Deconvolution for Low-Light Image Deblurring}


\author{$\text{Liang~Chen}$,
        $\text{Jiawei~Zhang}$,
        $\text{Zhenhua~Li}$,
        $\text{Yunxuan~Wei}$,
        $\text{Faming~Fang}$,
        $\text{Jimmy~Ren}$,
        and~$\text{Jinshan~Pan}$
\IEEEcompsocitemizethanks{\IEEEcompsocthanksitem L. Chen is with The University of Adelaide, SA, Australia. E-mail: liangchen527@gmail.com. This is work is done when L. Chen interned at Sensetime.
\IEEEcompsocthanksitem J. Zhang, Y. Wei, and J. Ren are with Sensetime research, Shenzhen, China, and J. Zhang is the corresponding author. E-mail: zhjw1988@gmail.com. 
\IEEEcompsocthanksitem Z. Li is with Nanjing University of Aeronautics and Astronautics, Nanjing, China. E-mail: zhenhua.li@nuaa.edu.cn. 
\IEEEcompsocthanksitem F. Fang is with East China Normal University, Shanghai, China. E-mail: fmfang@cs.ecnu.edu.cn.
\IEEEcompsocthanksitem J. Pan is with Nanjing University of Science and Technology, Nanjing, China. E-mail: sdluran@gmail.com.}}


\newcommand{\red}[1]{\textcolor{red}{#1}}
\newcommand{\blue}[1]{\textcolor{blue}{#1}}
\IEEEtitleabstractindextext{
\begin{abstract}
Images taken under the low-light condition often contain blur and saturated pixels at the same time.
Deblurring images with saturated pixels is quite challenging. Because of the limited dynamic range, the saturated pixels are usually clipped in the imaging process and thus cannot be modeled by the linear blur model.
Previous methods use manually designed smooth functions to approximate the clipping procedure. Their deblurring processes often require empirically defined parameters, which may not be the optimal choices for different images.
In this paper, we develop a data-driven approach to model the saturated pixels by a learned latent map. Based on the new model, the non-blind deblurring task can be formulated into a maximum a posterior (MAP) problem, which can be effectively solved by iteratively computing the latent map and the latent image.
Specifically, the latent map is computed by learning from a map estimation network (MEN), and the latent image estimation process is implemented by a Richardson-Lucy (RL)-based updating scheme.
To estimate high-quality deblurred images without amplified artifacts, we develop a prior estimation network (PEN) to obtain prior information, which is further integrated into the RL scheme.
Experimental results demonstrate that the proposed method performs favorably against state-of-the-art algorithms both quantitatively and qualitatively on synthetic and real-world images.

\end{abstract}

}

\maketitle
\IEEEdisplaynontitleabstractindextext
\IEEEpeerreviewmaketitle 

\begin{figure*}
\centering
	\begin{minipage}[b]{0.195\linewidth}
		\centering
		\centerline{
			\includegraphics[width =\linewidth]{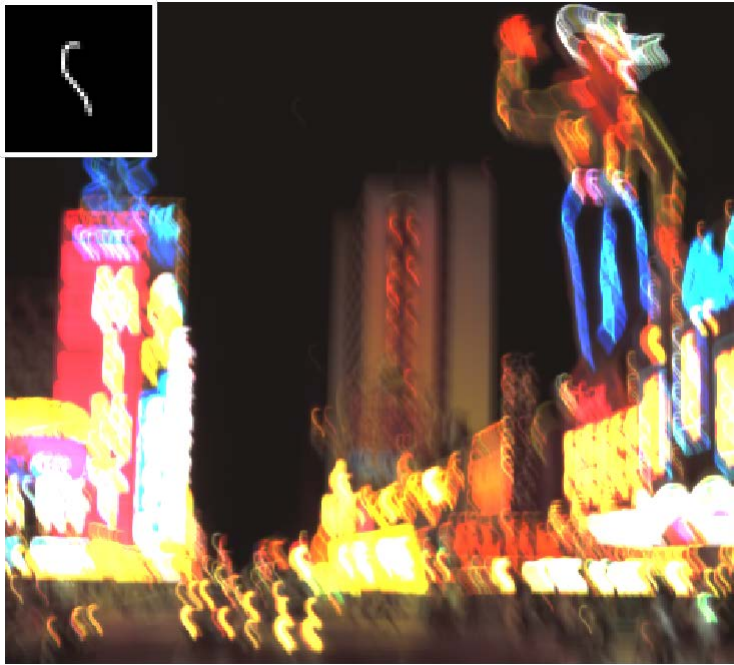}}
		\centerline{(a) Blurry image}
	\end{minipage}
	\begin{minipage}[b]{0.195\linewidth}
		\centering
		\centerline{
			\includegraphics[width =\linewidth]{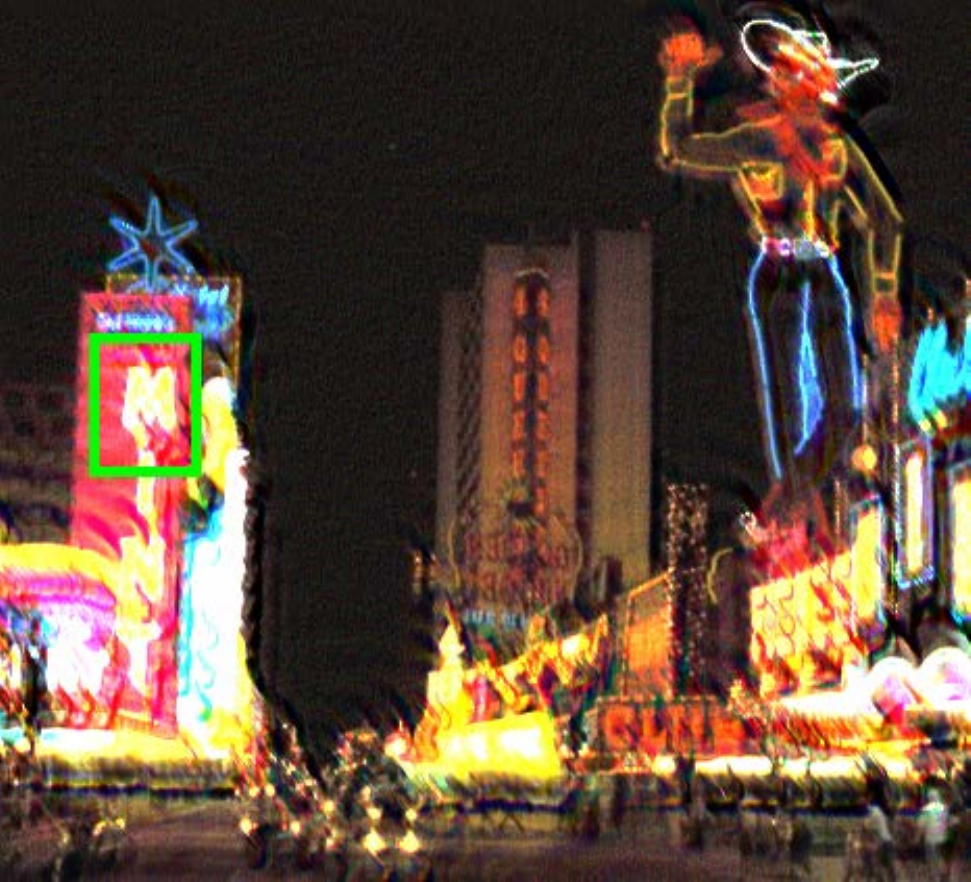}}
		\centerline{(b) Whyte \etal \cite{Whyte14deblurring}}
	\end{minipage}
	\begin{minipage}[b]{0.195\linewidth}
		\centering
		\centerline{
			\includegraphics[width =\linewidth]{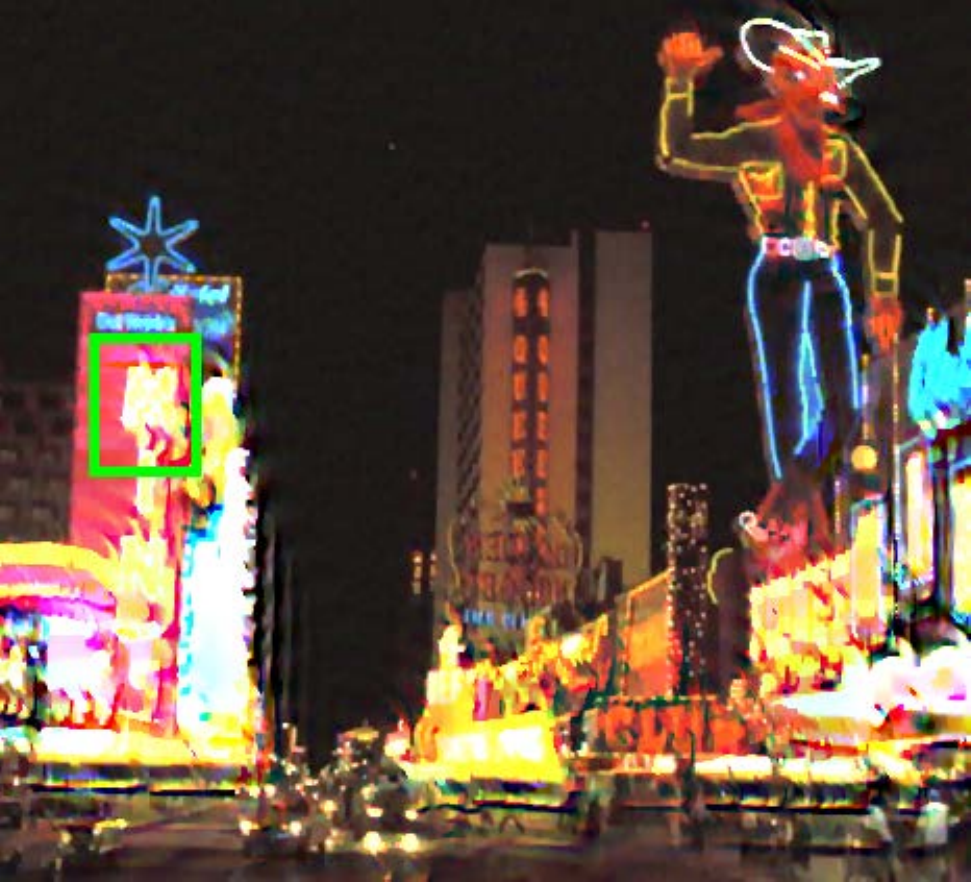}}
		\centerline{(c) Cho \etal \cite{cho2011outlier}}
	\end{minipage}
	\begin{minipage}[b]{0.195\linewidth}
		\centering
		\centerline{
			\includegraphics[width =\linewidth]{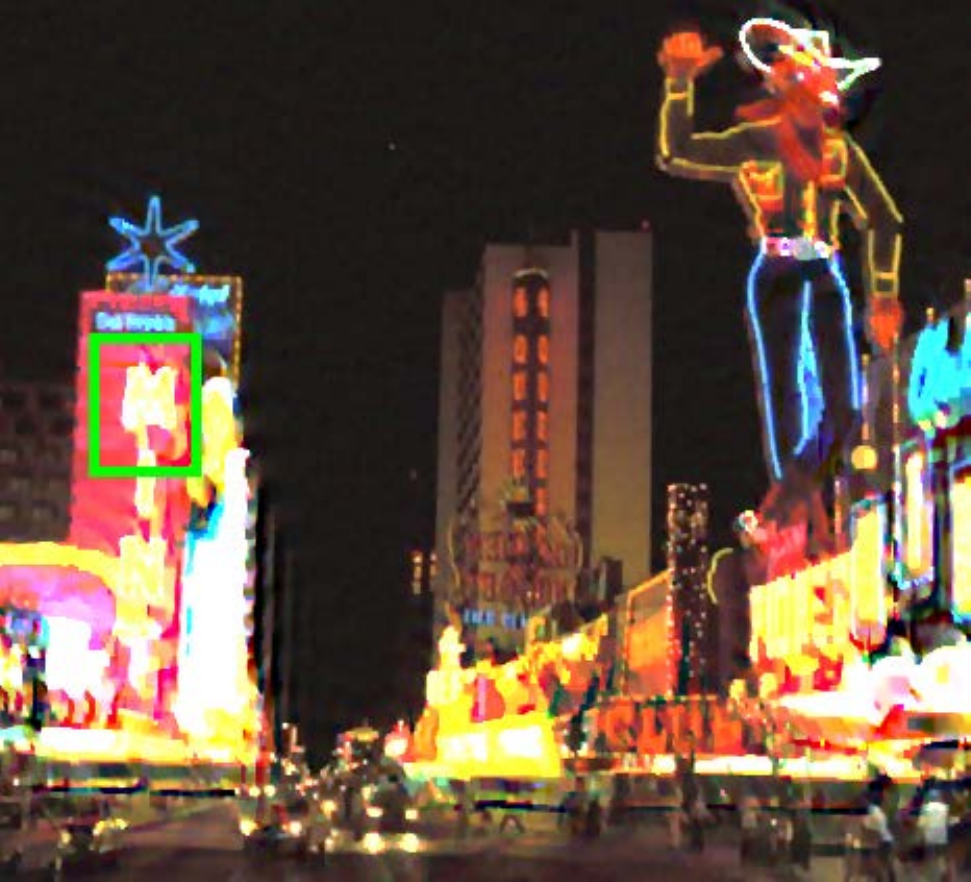}}
		\centerline{(d) Pan \etal \cite{pan2016robust}}
	\end{minipage}
	\begin{minipage}[b]{0.195\linewidth}
		\centering
		\centerline{
			\includegraphics[width =\linewidth]{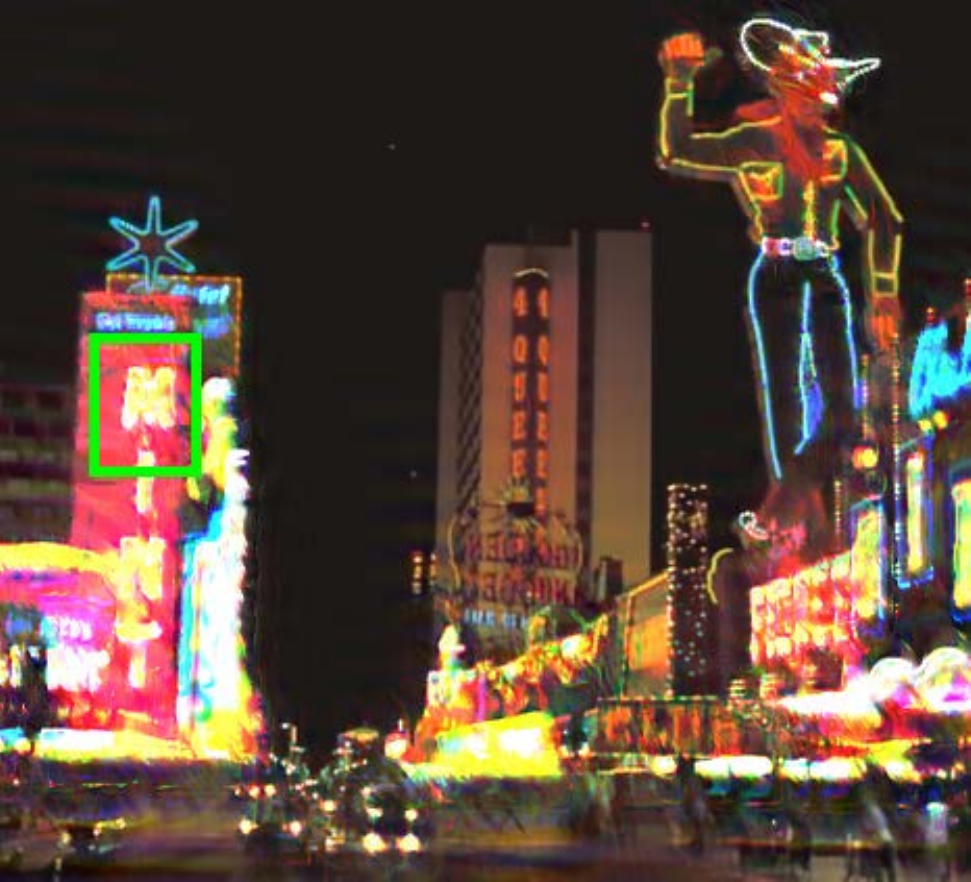}}
		\centerline{(e) NBDN~\cite{chen2021learning}}
	\end{minipage}\\
	\begin{minipage}[b]{0.195\linewidth}
		\centering
		\centerline{
			\includegraphics[width =\linewidth]{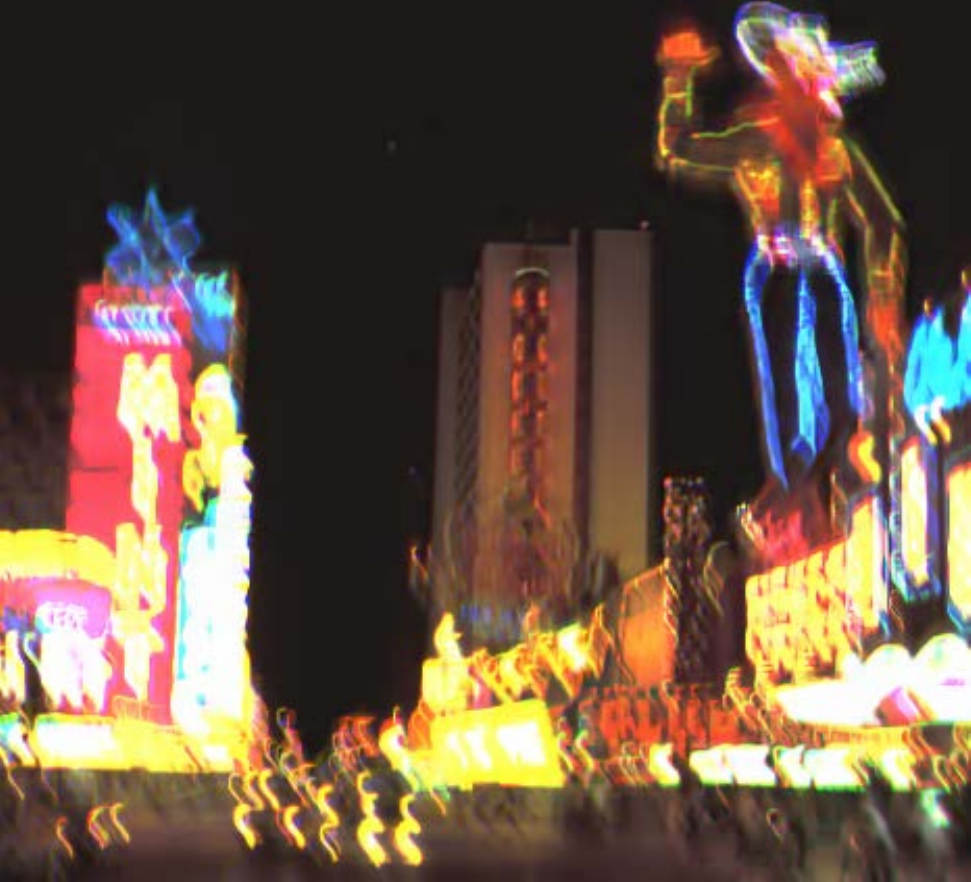}}
		\centerline{(f) SRN~\cite{tao2018scale}}
	\end{minipage}
	\begin{minipage}[b]{0.195\linewidth}
		\centering
		\centerline{
			\includegraphics[width =\linewidth]{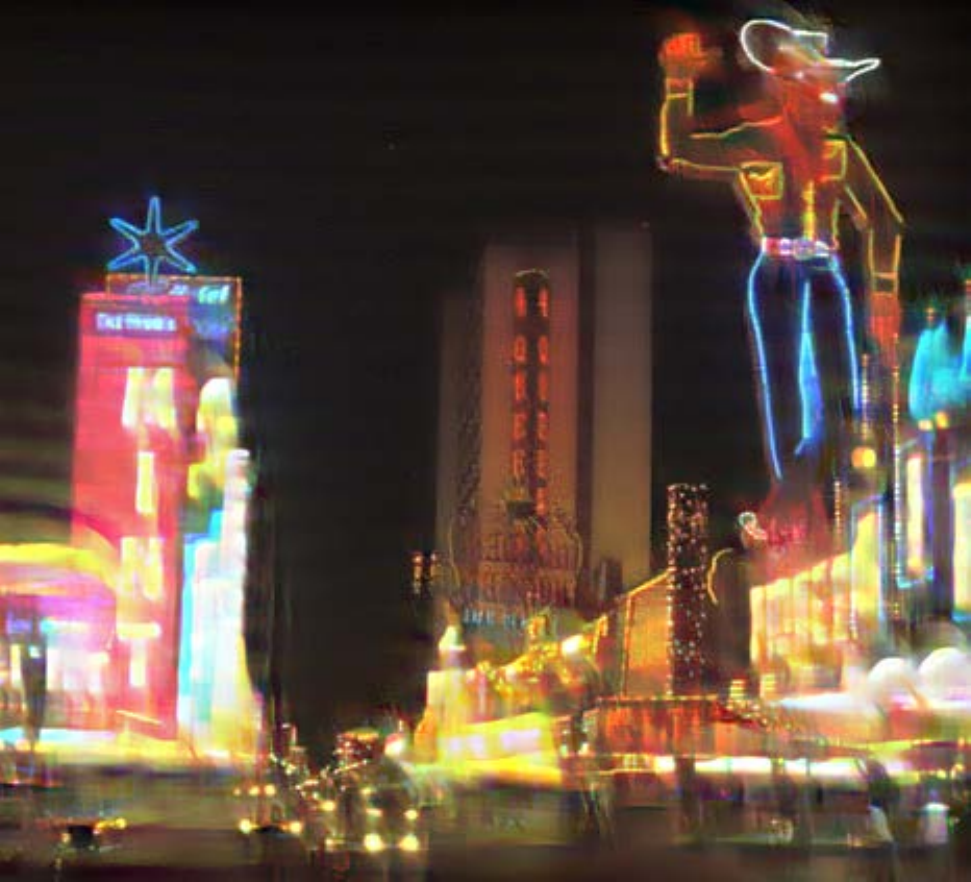}}
		\centerline{(g) FCNN~\cite{Zhang_2017_CVPR}}
	\end{minipage}
	\begin{minipage}[b]{0.195\linewidth}
		\centering
		\centerline{
			\includegraphics[width =\linewidth]{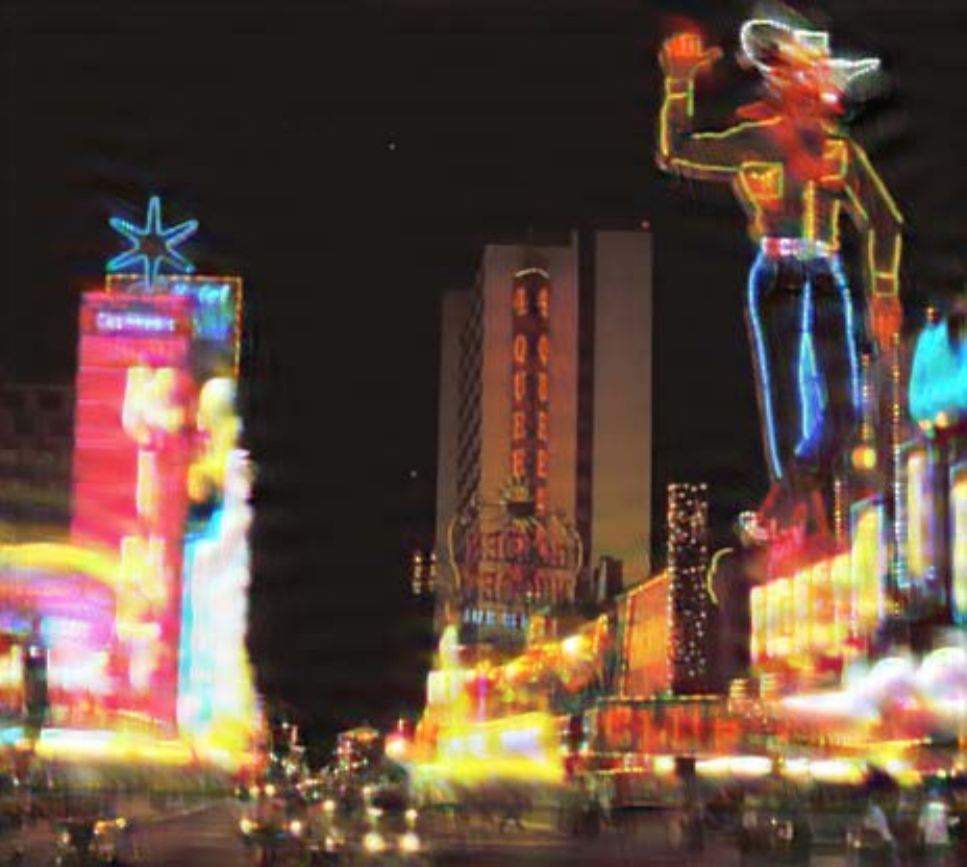}}
		\centerline{(h) DWDN~\cite{dong2020deep}}
	\end{minipage}
	\begin{minipage}[b]{0.195\linewidth}
		\centering
		\centerline{
			\includegraphics[width =\linewidth]{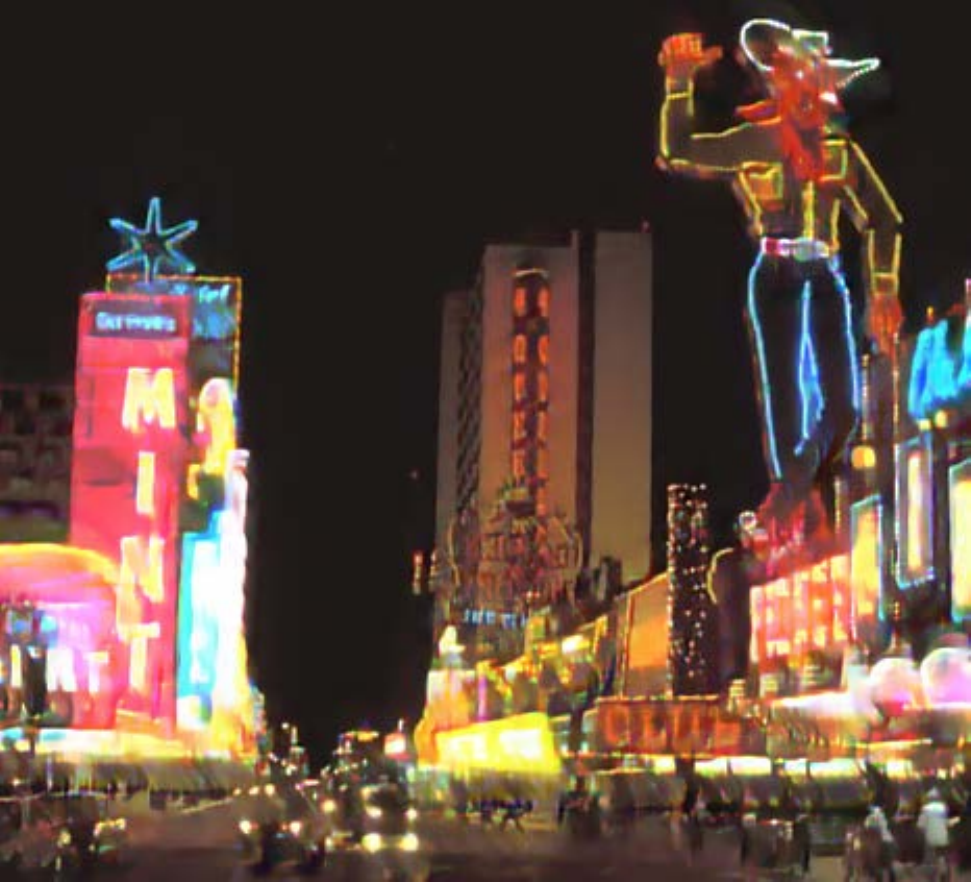}}
		\centerline{(i) Ours}
	\end{minipage}
	\begin{minipage}[b]{0.195\linewidth}
		\centering
		\centerline{
			\includegraphics[width =\linewidth]{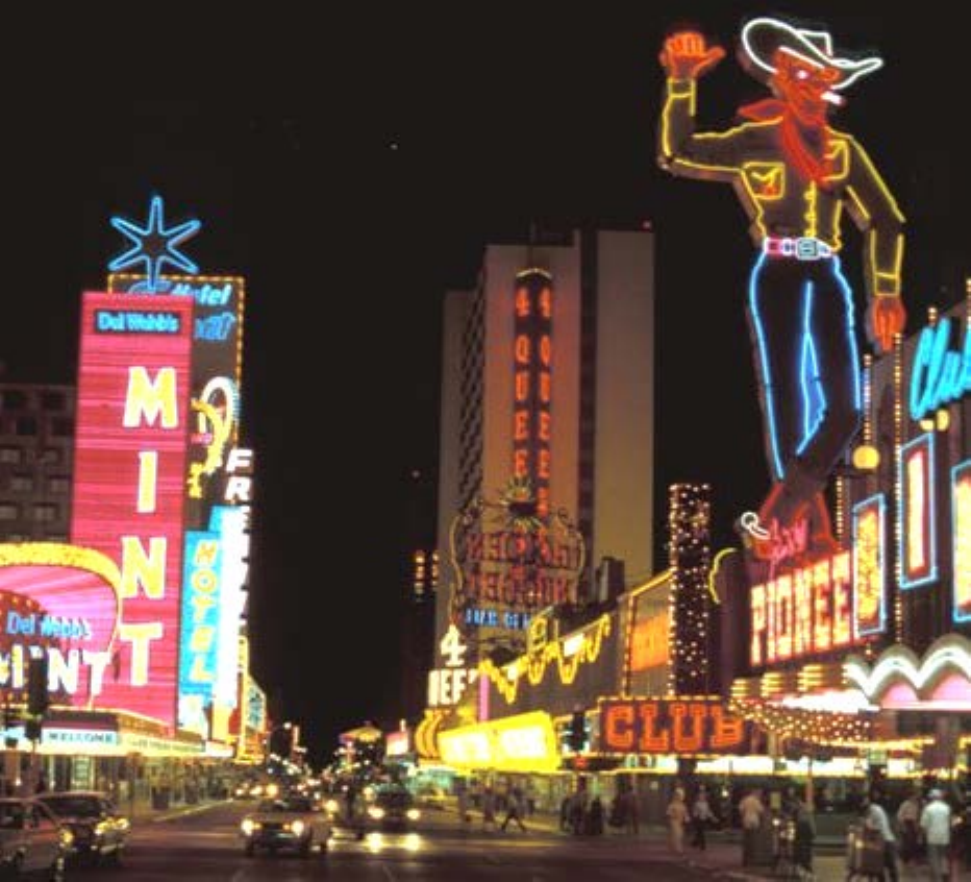}}
		\centerline{(j) GT}
	\end{minipage}
	\caption{Deblurring results of a saturated blurry image.
	The estimated kernel is shown in the white box of (a).
	Methods based on Eq.~\eqref{eq blur}~\cite{Zhang_2017_CVPR,dong2020deep} generate results with many ringings in the saturated regions.
Large blurs still remain in the result from the end-to-end learning-based method~\cite{tao2018scale}.
Saturated pixels cannot be properly handled for the robust models~\cite{cho2011outlier,pan2016robust,chen2021learning,Whyte14deblurring}, and they are ineffective in removing blur and ringings around the saturated regions, as shown in the green boxes.
In comparison, the proposed method can generate a high-quality result with fewer artifacts. Please zoom-in for a better view.}
	\label{fig tease}
\end{figure*}

\section{Introduction}
\label{sec intro}
Non-blind deblurring aims to recover a sharp image given a blurry one and the corresponding blur kernel.
Basically, the blurring process can be modeled by:
\begin{equation}
\label{eq blur}
B = I \otimes K,
\end{equation}
where $B$, $I$, and $K$ are the blurry image, latent image, and the blur kernel, respectively, and $\otimes$ denotes the convolution operation.

This problem is highly ill-posed and has gained considerable attention in recent years~\cite{krishnan2009fast,schmidt2013discriminative,kaiZhang_2017_CVPR,Zhang_2017_CVPR,dong2018learning,gong2018learning,dong2020deep}.
Despite their effectiveness in many cases, most of the above mentioned methods fail to consider the side-effects brought by saturated pixels, which are not rare when images are captured in a high dynamic range scenario at night, as shown in Fig.~\ref{fig tease}~(a).
Without proper handling, saturated pixels will cause severe artifacts in the deblurred results from the above mentioned methods as shown in Fig.~\ref{fig tease} (g) and (h).
The main reason is that saturated pixels cannot be well modeled by the linear blur model in Eq.~\eqref{eq blur}.
To this end, some studies~\cite{cho2011outlier,pan2016robust,hu2014deblurring,chen2020oid,chen2021learning} suggest discarding saturated pixels and using only unsaturated pixels for the deblurring process, which reformulates the blur model into:
\begin{equation}
\label{eq nbdn}
\tilde{M}\circ B = \tilde{M}\circ (I\otimes K),
\end{equation}
where $\tilde{M}$ is the weighting matrix. When $\tilde{M}$ is sufficiently small (\ie often takes value 0), the corresponding pixels will not contribute during the deblurring process, and vice versa for pixels assigning with large $\tilde{M}$ values.
However, separating the saturated and unsaturated regions from the blurry images is not trivial. If the separation is less accurate, the deblurred images may contain significant artifacts around the saturated regions (Fig.~\ref{fig tease} (d)).
In addition, due to the ignoring of saturated regions during the deblurring process, blur in these regions cannot be entirely removed as presented in the green boxes in Fig.~\ref{fig tease} (c), (d), and (e).

Based on the property of the saturated pixels, several algorithms~\cite{cho2011outlier,Whyte14deblurring,chen2020oid} usually formulate the blurring process with saturated pixels by:
\begin{equation}
\label{eq blur_clip}
B_i = C\left((I \otimes K)_i\right),
\end{equation}
where $C(\cdot)$ is the clipping function.
When the pixel $i$ is within the dynamic sensor range, $C((I \otimes K)_i)$ = $(I \otimes K)_i$; otherwise, $C((I \otimes K)_i)$ returns the maximum intensity of the sensor range. Compared to the blur models in Eqs.~\eqref{eq blur} and~\eqref{eq nbdn}, all pixels are included in the imaging process and can be modeled by Eq.~\eqref{eq blur_clip}.

However, the clipping function $C(\cdot)$ involved in Eq.~\eqref{eq blur_clip} is non-differentiable. To solve the problem, Whyte~\etal~\cite{Whyte14deblurring} approximate it with a specially-designed smooth function~\cite{chen1996class}.
To estimate the latent image, they propose a maximum likelihood (ML) framework and use a Richardson-Lucy (RL) updating scheme~\cite{lucy1974iterative,richardson1972bayesian} for optimization.
Although the blur in the saturated regions can be removed to a certain extent, their result in Fig.~\ref{fig tease}~(b) still contains some ringings and artifacts around image edges due to the lack of appropriate image prior information.
Moreover, as the parameters in the smooth function are important for modeling the saturated pixels, only empirical observation to determine these parameters may not model the saturated pixels in different images.
In addition, the computational costs for their method are also high because of the tedious optimization process.

With the development of the convolution neural network (CNN), some methods~\cite{nah2017deep,ren2017video,tao2018scale,kupyn2019deblurgan,wang2022uformer,tsai2022stripformer} utilize the large capacity of neural networks to directly restore the sharp image from the blurry input.
Without using the blur kernel and blur model, they do not effectively remove blur as shown in Fig.~\ref{fig tease}~(f).
The CNN models are also adopted in the non-blind deblurring task~\cite{Zhang_2017_CVPR,kaiZhang_2017_CVPR}. Most of them iteratively deconvolve and denoise the blurry image in a half-quadratic splitting framework, where CNNs are served as denoisers.
However, these methods are less effective for saturated blurry images (Fig.~\ref{fig tease} (g)) as they are based on the linear blur model~Eq.~\eqref{eq blur}.

In this paper, we propose a new method for deblurring saturated images.
First, based on the blur model in Eq.~\eqref{eq blur_clip}, our method uses a learnable latent map $M$, which is determined by the latent image and the blur kernel, to replace the clipping function.
Because both saturated and unsaturated pixels can be well modeled by Eq.~\eqref{eq blur_clip}, our method does not require sophisticated saturated pixel discarding processes~\cite{chen2021learning}, thus avoiding the issues brought by this strategy. We discuss the detailed differences between our work and the model used in NBDN~\cite{chen2021learning} in Sec.~\ref{sec nbdn}.
Meanwhile, compared to the previous model in~\cite{Whyte14deblurring} that adopts a smooth function to approximate the clipping function, the proposed method does not require heuristically designed functions and avoids tedious parameter tuning for different images. Comparisons of our model against that in~\cite{Whyte14deblurring} can be found in Sec.~\ref{sec whyte}

Then, we formulate the deblurring task into a maximum a posterior (MAP) problem and solve it by iteratively computing the latent map and the latent image.
During the latent map estimation step, we use a map estimation network (MEN) to compute $M$ in every iteration.
For the latent image updating step, we use an RL updating strategy to optimize the posterior maximization problem.
Recall that RL method often introduces artifacts in the deblurrd results~\cite{tai2010richardson}, some works~\cite{cho2011outlier,hu2014deblurring} propose to use a sparse prior~\cite{Levin07image} to suppress ringings.
Even though this prior is effective, it may not be the best choice in practice.
Thus, we further propose a prior estimation network (PEN), which is trained from a large amount of synthetic saturated blurry and clean image pairs, and integrate it into the RL updating strategy to facilitate image restoration.
As shown in (Fig.~\ref{fig tease} (i)), our method is able to handle the blurry images with large situated regions and generates a better deblurred result.

\begin{figure*}
\centering
	\begin{minipage}[b]{\linewidth}
		\centering
		\centerline{
			\includegraphics[width =\linewidth]{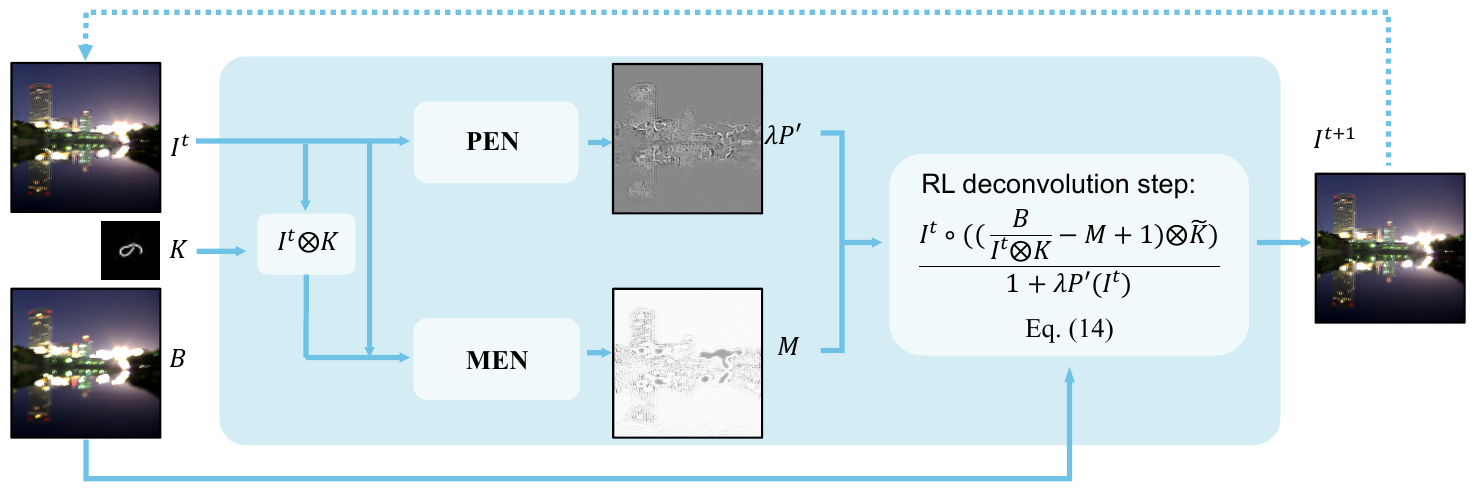}}
	\end{minipage}
	\caption{Overview of our deep Richardson-Lucy (RL) deconvolution method. We obtain the sharp image by iteratively updating the latent image and the latent map. PEN and MEN denote the prior estimation network and map estimation network, respectively. Please see the text for details.}
\label{fig pipeline}
\end{figure*}

The overview of the proposed network is shown in Fig.~\ref{fig pipeline}.
Extensive experimental results demonstrate the effectiveness of our method against state-of-the-arts for synthetic and real saturated blurry images.

\section{Related Work}
\label{sec related}
This section briefly reviews image deconvolution techniques and advances made in the non-blind deblurring task and pioneer arts involving saturation.

\noindent\textbf{Image deconvolution.} Deconvolution is a general image processing technique used to remove the blur caused by a known or estimated blur kernel (which is often obtained by blind deblurring methods~\cite{chen19blind,chen2020enhanced}). Besides applied in the deblurring task \cite{dong2020deep,schuler2013machine,son2017fast,Zhang_2017_CVPR}, image deconvolution techniques are also widely-used in other image restoration tasks, such as image super-resolution \cite{faramarzi2013unified,laghrib2018simultaneous} and enhancement \cite{cannell2006image,osher1990feature}. Before the widespread use of deep learning techniques, deconvolution has been studied by various researchers using different approaches. Those efforts include Weiner filter \cite{wiener1949extrapolation,katkovnik2005spatially}, Richardson-Lucy algorithm \cite{lucy1974iterative,richardson1972bayesian}, and shock filter \cite{osher1990feature,alvarez1994signal}, to name a few. In recent years, with the development of CNN, deconvolution has been integrated with deep leaning skills to further improve the image restoration performance \cite{dong2020deep,kaiZhang_2017_CVPR,zhang2021designing}. Our method also falls into this category. In Sec.~\ref{sec pre}, we provide more details elaborated on the basis of our method.

\noindent\textbf{Deblurring without saturation.} Due to the ill-posed nature of the inverse problem, strong image priors are required to regularize the solution space.
Hyper-laplacian prior, known for its sparsity and fitness of heavy-tail distribution of natural images, is broadly applied in recent deblurring works~\cite{Levin07image,krishnan2009fast,cho2011outlier,pan2016robust}.
In~\cite{roth2005fields}, Roth and Black use a field of experts (FOE) framework to model image priors.
The FOE framework is further extended in~\cite{schmidt2011bayesian,schmidt2013discriminative,xiao2016learning,dong2018learning,ren2019simultaneous} to restore blurry images. Although good results have been achieved, solving models related to FOE bears a large computational burden.

Instead of developing manually-designed regularizers, numerous image restoration algorithms turn to CNNs~\cite{schuler2013machine,xu2014deep,Zhang_2017_CVPR,kaiZhang_2017_CVPR,dong2018learning,ren2018deep}.
Schuler~\etal~~\cite{schuler2013machine} first impose a regularized inversion of the blur in the Fourier domain and then propose to remove the artifacts using a learned multi-layer perceptron (MLP).
In~\cite{Zhang_2017_CVPR} and~\cite{kaiZhang_2017_CVPR}, they use the variable substitution method to decouple the non-convex restoring task into a quadratic deconvolution step and a denoising problem which can be effectively solved by a learned denoiser.
Instead of stacking learned regularizers in a deblurring chain, Gong~\etal~\cite{gong2018learning} suggest learning a universal projector for different updating steps.
Due to the neglecting of side-effects brought by saturating, although effective in most cases, these methods are ineffective in dealing with saturated blurry images.

\noindent\textbf{Deblurring with saturation.} Saturation has not received wide attention in the non-blind deblurring literature.
Cho~\etal~\cite{cho2011outlier} consider saturated pixels as outliers, and they propose an EM-based algorithm to iteratively identify and exclude outliers in the blurry image.
This idea is broadly extended in some recent works~\cite{hu2014deblurring,pan2016robust,gong2017self,chen2020oid,chen2021learning}.
In~\cite{pan2016robust}, saturated pixels are detected by computing the residual between the blurry image and the convolution output (\ie $B-I\otimes K$).
Then they develop a specially-designed fidelity term, where smaller weights are applied to the detected saturated pixels, for optimization.
Moreover, based on this saturated pixel detection, Chen~\etal suggest using more faithful ways to detect saturated pixels, which are based on the maximum entropy rule~\cite{chen2020oid} or implemented using CNN~\cite{chen2021learning}.
%
%
To better handle the saturated regions, Hu~\etal~~\cite{hu2014deblurring} combine the works from~\cite{Whyte14deblurring} and~\cite{cho2011outlier}, and they further propose an EM-based regularized RL deconvolution method.

Different from previous works that explicitly exclude saturated pixels, some works suggest altering the clipping function in Eq.~\eqref{eq blur_clip} so that all pixels are included in the deblurring process.
Whyte~\etal~\cite{Whyte14deblurring} extend the RL algorithm with an approximation function~\cite{chen1996class} to model the property of saturated pixels. But their approaches require empirical parameter settings during optimization, and the deblurring process is also time-consuming.
Recently, Chen \etal~\cite{chen2021blind} propose to use a latent map to replace the clipping function, where the latent map is naively determined by the reverse of the convolving result.
However, their work requires an empirically defined threshold. This setting works well for the blind deblurring task since a modest latent image can also lead to a precisely estimated kernel.
When it comes to the non-blind deblurring task, the empirically defined threshold may jeopardize the final result.
Differently, we use a learning-based approach to compute the latent map, which avoids the empirical parameter tuning and naive settings.

There are also works suggest using CNN to directly deal with the low-light condition. Xu~\etal~\cite{xu2014deep} train an end-to-end network to handle blurry images with outliers, but their work demands fine-tuning for every blur kernel.
Ren~\etal~extend~\cite{ren2018deep} to handle any blur kernel by a generalized low-rank approximation.
However, their method does not work well for motion blur.

\section{Preliminary}
\label{sec pre}
This section describes the basic formulations of a standard non-blind deblurring method and the original Rishcardson-Lucy (RL) deconvolution algorithm.

\subsection{Non-Blind Deblurring}
\label{sec nonblind}
The blurring process formulated in Eq.~\eqref{eq blur} often involves noise that is often modeled as following either a Poisson or Gaussian distribution. For both cases, the non-blind deblurring process can be formed as the following minimization problem:
\begin{equation}
\label{eq minimize}
\min_I \mathcal{L}(B, I\otimes K) + \lambda P(I),
\end{equation}
where the fidelity term $\mathcal{L}$ measures the distance between the blurry image and the convolving result between the estimated sharp image and the blur kernel. Assuming the noise involved in the blurring process follows a known distribution, we can reformulate the fidelity term into the negative log-likelihood: $\mathcal{L}(B, I\otimes K) = -\log \mathcal{P}(B\vert I\otimes K)$; $P(I)$ is the prior term for $I$; The scalar weight $\lambda$ balances the contribution of these two terms.

If the noise follows a Gaussian distribution, we can model $\mathcal{L}(B, I\otimes K) = -\log (e^{-\frac{\Vert B - I \otimes K\Vert^2}{2\sigma^2}})$, where $\sigma$ is the standard deviation of the Gaussian noise mode. With the defined fidelity term and prior term, solving Eq.~\eqref{eq minimize} is not difficult, and it is typically solved using standard linear least-squares algorithms, such as conjugate gradient descent.

If the noise follows a Poisson distribution, we can model $\mathcal{L}(B, I\otimes K) = -\log (\frac{e^{-I \otimes K} (I \otimes K)^B}{B !})$. Note that
image noise is assumed to be independent and identically
distributed (i.i.d.). Hence, $\mathcal{L}(B, I\otimes K)$ is accessed per pixel.
The classic deblurring method for deblurring images with Poisson noise is the RL algorithm~\cite{lucy1974iterative,richardson1972bayesian}, and we give detailed derivation steps in Sec.~\ref{sec rl}.
Following practices from existing deblurring works that aim for saturated images~\cite{hu2014deblurring,Whyte14deblurring,chen2021blind}, we also assume the noise involved in the blurring process to follow a Poisson distribution.

\subsection{Richardson-Lucy Deconvolution Algorithm}
\label{sec rl}
The RL algorithm~\cite{lucy1974iterative,richardson1972bayesian} is well known for deconvolving blurry images involved in Poisson processes. In this section, we present the deviation of the basic RL for solving Eq.~\eqref{eq minimize} when the noise in the blurring process is assumed to follow a Poisson distribution. Same as the original idea in~\cite{lucy1974iterative,richardson1972bayesian}, we only consider the fidelity term in the function.
Given the likelihood $\mathcal{P}(B\vert I\otimes K) = \frac{e^{-I \otimes K} (I \otimes K)^B}{B !}$, we can reformulate the fidelity term from Eq.~\eqref{eq minimize} into:
\begin{equation}
\label{eq minrl}
\min_I I \otimes K - \log (I\otimes K) \circ B,
\end{equation}
where $\circ$ is the Hadamard product. Taking the derivative of the equation \wrt $I$ and setting it to be zero, we obtain
\begin{equation}
\label{eq basic_rl}
\textbf{1}\otimes \widetilde{K} - \frac{B}{I\otimes K} \otimes \widetilde{K} = 0,
\end{equation}
where $\textbf{1}$ is an all-one image; $\widetilde K$ can be obtained by flipping $K$ upside-down and then left-to-right. Recall that $\textbf{1}\otimes \widetilde{K} = \textbf{1}$, from Eq.~\eqref{eq basic_rl} we obtain the fixed point iteration by: 
\begin{equation}
\label{eq fix}
\frac{I^{t+1}}{I^t} = \textbf{1} = \frac{B}{I^t\otimes K}\otimes K \Rightarrow I^{t+1} = I^t \circ (\frac{B}{I^t\otimes K}\otimes K),
\end{equation}
where $t$ is the iteration index.
The updating scheme in Eq.~\eqref{eq fix} is known as RL deconvolution algorithm. The convergence property is also analyzed in~\cite{lucy1974iterative}. Please refer to the original papers~\cite{lucy1974iterative,richardson1972bayesian} for detailed descriptions.

\section{Proposed Method}
The overview of the proposed network is shown in Fig.~\ref{fig pipeline}. Our method simultaneously estimates the latent map for modeling saturated pixels, the prior information as the image prior, and the latent clean image in an iterative optimization framework.
Inputs of our model include a saturated blurry image $B$ and the corresponding blur kernel $K$.
For every iteration step $t$, given the latent image $I^t$ updated from the previous iteration, we first use the proposed prior estimation network (PEN) and map estimation network (MEN) to estimate the prior information $\lambda P'(I^t)$ and the latent map $M$.
Then, the latent image $I^{t+1}$ can be updated through a Richardson-Lucy-based optimization scheme, which is further used as the input for the next iteration.
We use the blurry image $B$ as the initialization of the latent image $I^0$.
We present details of our model in the following.

\subsection{Proposed Blur Model}
\label{sec ourmodel}
The degrading process in Eq.~\eqref{eq blur_clip} involves a non-linear clipping function, which is also non-differentiable.
To make the problem more tractable, instead of finding complicated approximations to model the degrading process, we propose a learnable latent map $M$ to replace the clipping function. Assume the imaging process follows the Poisson distribution, then the observed noisy blurry image $B$ of Poisson distribution with mean $M\circ (I\otimes K)$ obeys the conditional probability:
\begin{equation}
\begin{aligned}
\label{eq our}
\mathcal{P}(B\vert M\circ (I\otimes K)) = &\prod_i \frac{e^{-{(M\circ (I\otimes K))_i}} ((M\circ (I\otimes K))_i)^{B_i}}{B_i !},
\\&~~\text{s.t.}~~M = \mathcal{F}(I, K)
\end{aligned}
\end{equation}
where $i$ denotes a pixel location, and $\mathcal{F}(\cdot)$ represents a function that determines $M$ based on $I$ and $K$.

It is noteworthy that $M$ has a similar effect to the clipping function that can keep the blurry result within the sensor range.
With the help of $M$, saturated pixels can be well modeled by the proposed blur model. Meanwhile, compared to the blur model in Eq.~\eqref{eq blur_clip}, the proposed model in Eq.~\eqref{eq our} is differentiable \wrt $I$ and $K$, which facilitates the following optimization steps.
Note our model is different from that in~\cite{chen2021blind}. Because their blind deblurring task does not require a high quality latent image during the kernel estimation step, thus they can use a naive form of $\mathcal{F}$ (\ie assuming a uniform threshold $v$ in the image, for every pixel location $i$, $M_i=1$ if $(I\otimes K)_i<v$, or $M_i=v/(I\otimes K)_i$ if $(I\otimes K)_i>v$) to obtain a roughly estimated $I$.
However, our non-blind deblurring task requires a much more accurate $I$. Considering that determining proper thresholds for all situations is rather challenging, thus their empirically selected value (\ie $v$ is fixed as $0.9$ in their implementation for all images) is far from convincing in our setting.
Differently, we suggest using CNN to approximate $\mathcal{F}$ which can adapt to different situations given its large learning capacity.
We further present detailed comparisons between our model and that from~\cite{chen2021blind} in Section \ref{sec chen}.

\subsection{Optimization}
By formulating the optimization task in a Maximum a Posteriori (MAP) scheme, our objective is to find the optimal $\{I^{\ast}$, $M^{\ast}\}$ so that
\begin{equation}
\label{eq map}
\{I^{\ast}, M^{\ast}\}  = \arg\max_{I,M} \log \mathcal{P}(I, M\vert B, K),~~\text{s.t.}~~M = \mathcal{F}(I, K)
\end{equation}
Based on the Bayesian theorem, we obtain
\begin{equation}
\label{eq bayes}
\begin{aligned}
\{I^{\ast}, M^{\ast}\} = \arg\max_{I,M} \{&\log \mathcal{P}(B\vert K, I, M) + \log \mathcal{P}(I, M)\},
\\&~~\text{s.t.}~~M = \mathcal{F}(I, K)
\end{aligned}
\end{equation}
We can solve the above problem by the alternate minimization approach that iteratively updates $I$ and $M$. Thus, for each iteration step $t$, we can obtain the solutions by solving the following two sub-problems,
\begin{numcases}{}
\small
M = \mathcal{F}(I^{t}, K) \label{eq m_obj},\\
I^{t+1} = \arg\max_I \log\mathcal{P}(B\vert K, I, M) + \log\mathcal{P}(I). \label{eq img_obj}
\end{numcases}
Eq.~\eqref{eq m_obj} is deduced from the hard constraint in Eq.~\eqref{eq bayes} that any feasible solutions of $M$ are completely determined by $I$ and $K$, which is also the projection of the solution of $M$ in Eq.~\eqref{eq bayes}. The above updating scheme is similar to the projected alternating minimization (PAM) method in~\cite{Perrone14total}.

\noindent\textbf{Solving the sub-problem \wrt $M$.}
Defining a specific form of the function $F(\cdot)$ is not trivial. Current methods~\cite{chen2020oid,chen2021blind} resort to empirical findings for the definition, which may not suit different images.
Given the strong approximation ability of deep neural networks, we develop a map estimation network (MEN) to directly approximate the function. MEN uses the current estimated image $I^t$, and convolving result $I^t \otimes K$ as inputs and outputs $M$. Directly solving Eq.~\eqref{eq m_obj} using MEN, we can avoid the definition of $F(\cdot)$ and the possible heuristical settings brought by it.

\noindent\textbf{Solving the sub-problem \wrt $I$.}
Putting Eq.\eqref{eq our} into Eq.\eqref{eq img_obj} and defining the image prior $\mathcal{P}(I)$ as $\mathcal{P}(I) = \exp(-\lambda P(I))$, where $\lambda$ is the weight parameter, the objective in Eq.\eqref{eq img_obj} can be reformulated as,
\begin{equation}
\small
\label{eq img_reform}
\begin{aligned}
I^{t+1} &= \arg\max_I \log\mathcal{P}(B\vert K, I, M) + \log\mathcal{P}(I)\\
&= \arg\max_I \log \prod_i \frac{e^{-{(M\circ (I\otimes K))_i}} ((M\circ (I\otimes K))_i)^{B_i}}{B_i !} - \lambda P(I)\\
&= \arg \min_I M\circ (I\otimes K) - \log(M\circ (I\otimes K))\circ B + \lambda P(I).
\end{aligned}
\end{equation}
We use the RL updating scheme to solve Eq.~\eqref{eq img_reform} as:
\begin{equation}
\label{eq rl}
I^{t+1} = \frac{I^t\circ ((\frac{B}{I^t\otimes K} - M + \textbf{1}) \otimes \widetilde K)}{\textbf{1}+\lambda P'(I^t)},
\end{equation}
where $P'(\cdot)$ denotes the derivative of $P(\cdot)$.
Here the divisions are both element-wise operations. Please refer to the appendix for detailed derivations.

Generally, $P(I)$ can be the form of the classical sparse prior~\cite{Levin07image}. But it may not be the best choice for all scenarios.
In this paper, we propose a prior estimation network (PEN) to directly estimate the derivative of the regularization term (\ie $\lambda P'(\cdot)$).
%
PEN takes $I^t$ as inputs and outputs $\lambda P'(I^t)$, it implicitly learns the prior information \wrt $I^t$ and can be efficiently plugged into the optimization process in Eq.~\eqref{eq rl}.

\begin{algorithm}[tp]
\caption{Deep Richardson-Lucy (RL) deconvolution for saturated image deblurring}
\KwIn {Blurry image $B$, blur kernel $K$.}
\KwIn {The number of iterations $Q$.}
\KwOut {Sharp latent image $I^{Q}$.}
Initialize $I^0=B$. \\
t=0. \\
\While {$t<Q$}
{
$M=$MEN$(I^t, K)$\\
$\lambda P'(I^{t})=$ PEN $(I^{t})$\\
Update $I^t$ using Eq.~\eqref{eq rl}\\
$t\gets t+1$\\
}
\label{alg 1}
\end{algorithm}

\section{Network Details}
By integrating MEN and PEN, which are trained from a large amount of training data, into the proposed framework, our method can avoid the heuristically defined functions and manually parameter tuning in the prior arts.
In this section, we present the network configurations and the implementation details.

\subsection{Network Designs}
The inputs of our method include the blurry image and the corresponding blur kernel.
During each iteration stage, we perform the latent map (\ie $M$) estimation via MEN and prior estimation (\ie $\lambda P'(I)$) via PEN.
The updated latent image can be obtained with the updated $M$ and $\lambda P'(I)$ by performing the deblurring step given in Eq. \eqref{eq rl}. The detailed algorithm is shown in Algorithm \ref{alg 1}.

\noindent\textbf{Network architecture of MEN.}
In existing methods, \eg~\cite{Whyte14deblurring,chen2021blind}, different hyper-parameter settings may need to be tuned for different images, and a heuristically defined approximation function may not be the best choice in practice.
Instead of seeking manually designed representations, we suggest directly learning to estimate the latent map $M$ by the MEN from numerous training data.

MEN takes $I^t$ and $I^t \otimes K$ as inputs and outputs $M$. It is constructed with six res-blocks, and each block contains two convolution layers to generate 32 features. We add a rectified linear unit (ReLU) after the first convolution layer in every block, and a sigmoid layer is attached at the end.

\noindent\textbf{Network architecture of PEN.}
Instead of using a manually designed image prior and empirically selecting the weight parameter (\ie $\lambda$), we suggest directly estimating the prior information from the intermediate estimated latent image.

PEN uses $I^t$ as the input and outputs the prior information $\lambda P'(I^t)$. It is implemented by a 3-scale lightweight U-net.
Specifically, each scale in the U-net contains two convolutions, and each convolution layer is attached with a ReLU layer for activation.
The features from the first to the last scale are 8, 16, and 32, respectively.

The proposed network is fully differentiable and can be trained in an end-to-end manner.
Both MEN and PEN share weights through different iterations.
%

\subsection{Network Training and Implementation Details}
To make the training process more stable, we train both PEN and MEN by minimizing the difference between the estimated deblurred image $I^t_n$ from every stage with the ground truth $I^{gt}$ as:
\vspace{-0.3 cm}
\begin{equation}
\vspace{-0.1 cm}
\label{eq 8}
\mathcal{E} = \frac{1}{NQ}\sum_{n=1}^N \sum_{t=1}^Q\Vert I^t_n - I^{gt}\Vert_1,
\end{equation}
where $\Vert\cdot \Vert_1$ denotes the $L_1$ norm loss, $N$ is the number of training samples in every batch, and $Q$ is the maximum updating stages during training.
We first train PEN without considering MEN for all updating stages where $M$ is fixed as $\textbf{1}$.
Then both PEN and MEN are optimized until they are converged.

Our implementation is based on PyTorch~\cite{paszke2019pytorch}, we initialize our network according to~\cite{he2015delving}.
The training is carried by ADAM optimizer~\cite{kingma2014adam} with $\beta_1=0.9$, $\beta_2=0.999$, $\epsilon=10^{-8}$ and learning rate as 0.0001.
We set the batch size as 4, image size as 256 $\times$ 256, and the number of iterations $Q$ as 30, respectively.

\begin{table*}[t]
\centering
\caption{Quantitative evaluations on the given saturated blurry testing set.}
    \centering
    \scalebox{0.87}{
    \begin{tabular}{C{0.5 cm}C{0.9 cm}C{0.9 cm}C{0.9 cm}C{0.9 cm}C{0.9 cm}C{0.9 cm}C{1.1 cm}C{1.1 cm}C{0.9 cm}C{0.9 cm}C{0.9 cm}C{0.9 cm}C{0.9 cm}C{0.9 cm}}
    \toprule
    & Cho \cite{cho2011outlier}
    & Hu \cite{hu2014deblurring}
    & Whyte \cite{Whyte14deblurring}
    & Pan \cite{pan2016robust}
    & Chen \cite{chen2020oid}
    & SRN \cite{tao2018scale}
    & Uformer \cite{wang2022uformer}
    & Stripformer \cite{tsai2022stripformer}
    & FCNN \cite{Zhang_2017_CVPR}
    & IRCNN \cite{kaiZhang_2017_CVPR}
    & RGDN \cite{gong2018learning}
    & DWDN \cite{dong2020deep}
    & NBDN \cite{chen2021learning}
    & Ours\\
    \midrule
    \multicolumn{14}{c}{Results with GT blur kernels}\\
    \midrule
    PSNR & 20.27 &23.26 & 23.75 & 24.68 &24.85 & 22.95 &24.20 &24.48 & 24.80 & 19.34 & 20.94 &25.02 &25.25 & \textbf{25.66}\\ 
    SSIM & 0.7410 &0.7755 &0.8206 & 0.8550 &0.7713 & 0.7701 &0.8206 &0.8293 & 0.8470 & 0.6755 & 0.7530 &0.8515 &0.8544 & \textbf{0.8595}\\ 
    \midrule
    \multicolumn{14}{c}{Results with estimated kernels from~\cite{hu2014deblurring}}\\
    \midrule
    PSNR & 20.06 &23.01 & 23.20 & 23.66 &24.39 & 22.95 &24.20 &24.48 & 23.77 & 19.22 & 20.79 &24.51 &24.40 & \textbf{25.11}\\ 
    SSIM & 0.7318 &0.7657 &0.7965 & 0.8239 &0.8202 & 0.7701 &0.8206 &0.8293 & 0.8090 & 0.6728 & 0.7512 &0.8295 &0.8291 & \textbf{0.8367}\\ 
    \bottomrule
    \end{tabular}}
    \label{tab 1}
\end{table*}

\begin{figure*}\footnotesize
\centering
	\begin{minipage}[b]{0.16\linewidth}
		\centering
		\centerline{
			\includegraphics[width =\linewidth]{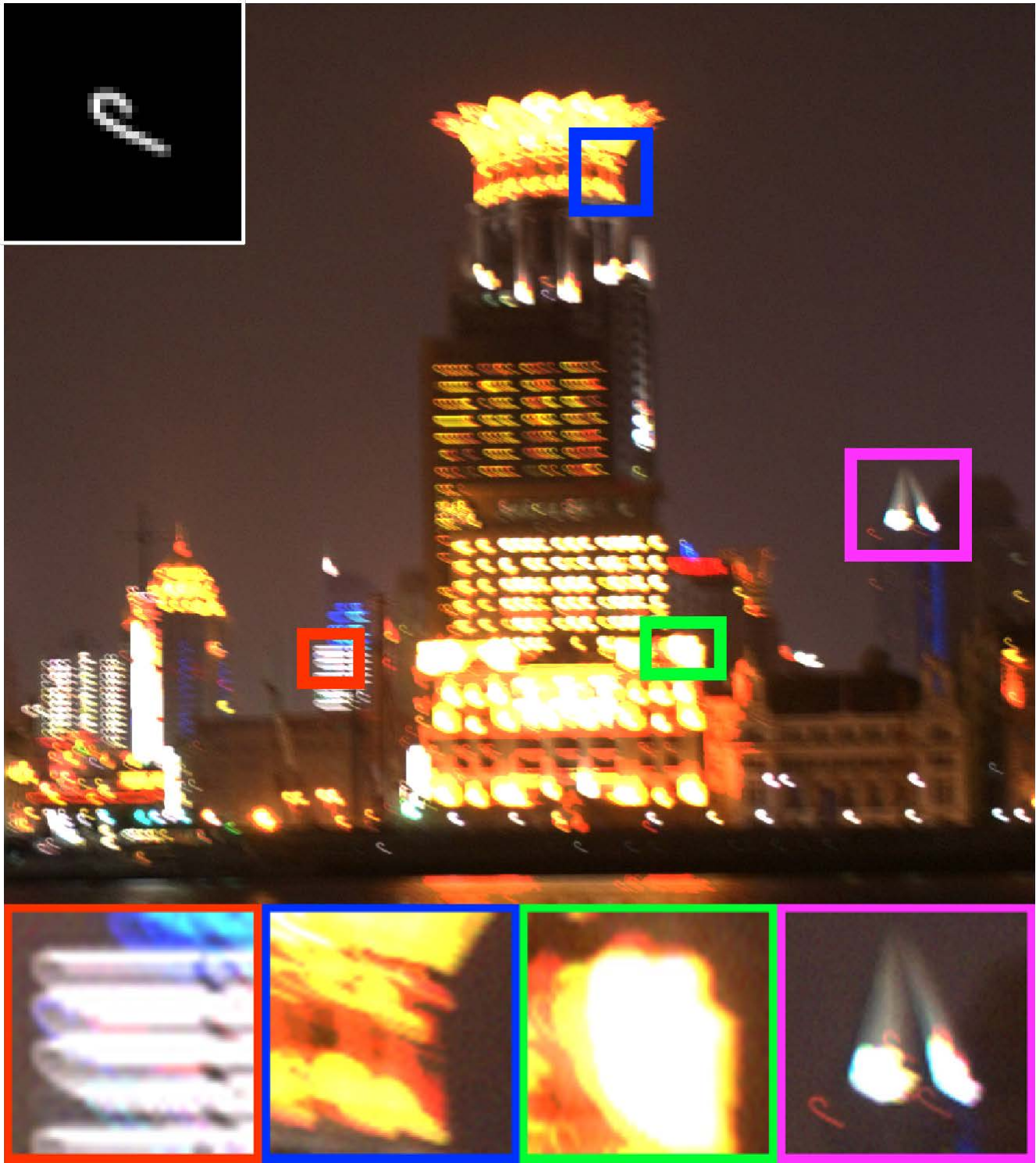}}
		\centerline{(a) Blurry image}
	\end{minipage}
	\begin{minipage}[b]{0.16\linewidth}
		\centering
		\centerline{
			\includegraphics[width =\linewidth]{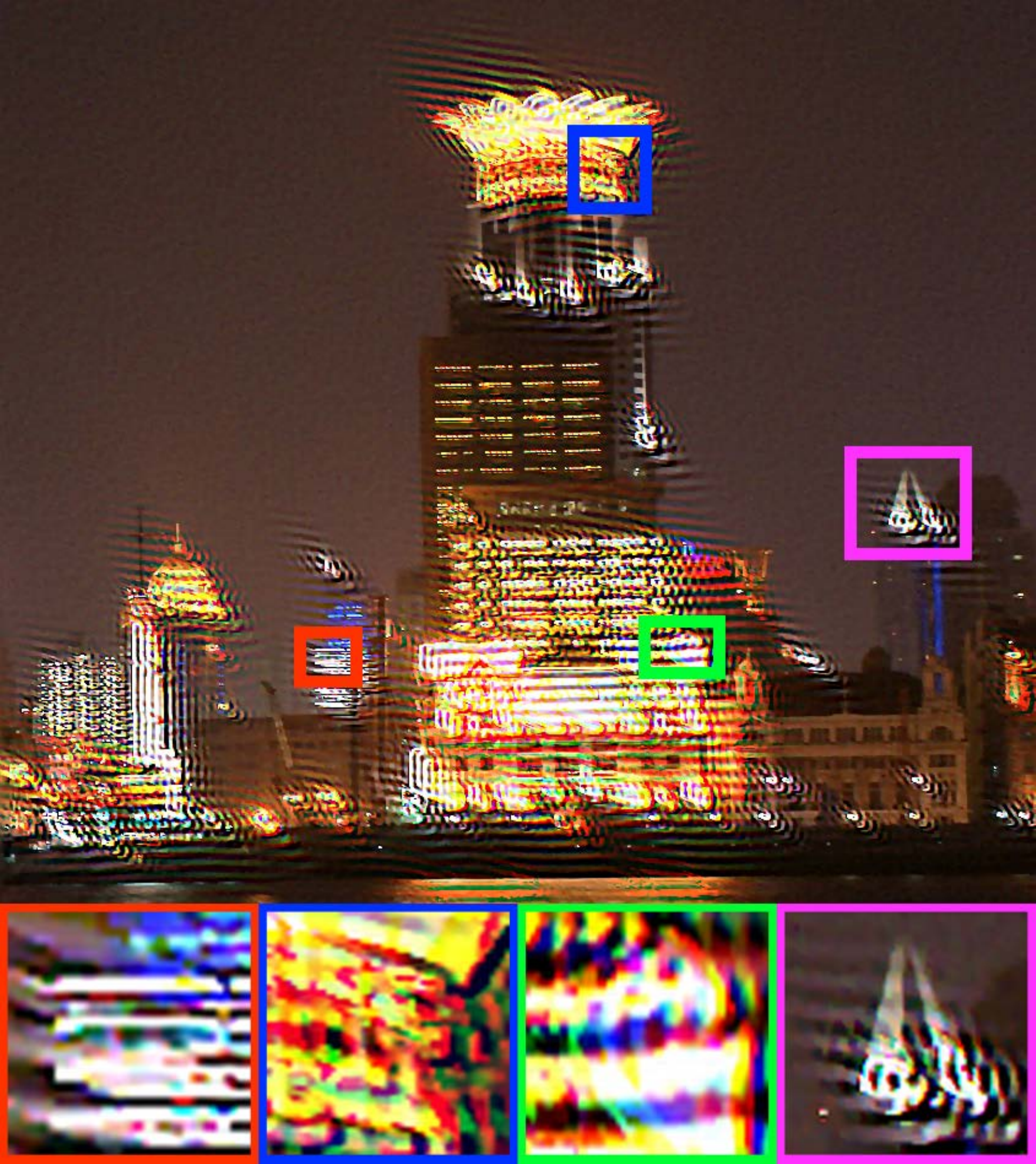}}
		\centerline{(b) Cho \etal~\cite{cho2011outlier}}
	\end{minipage}
	\begin{minipage}[b]{0.16\linewidth}
		\centering
		\centerline{
			\includegraphics[width =\linewidth]{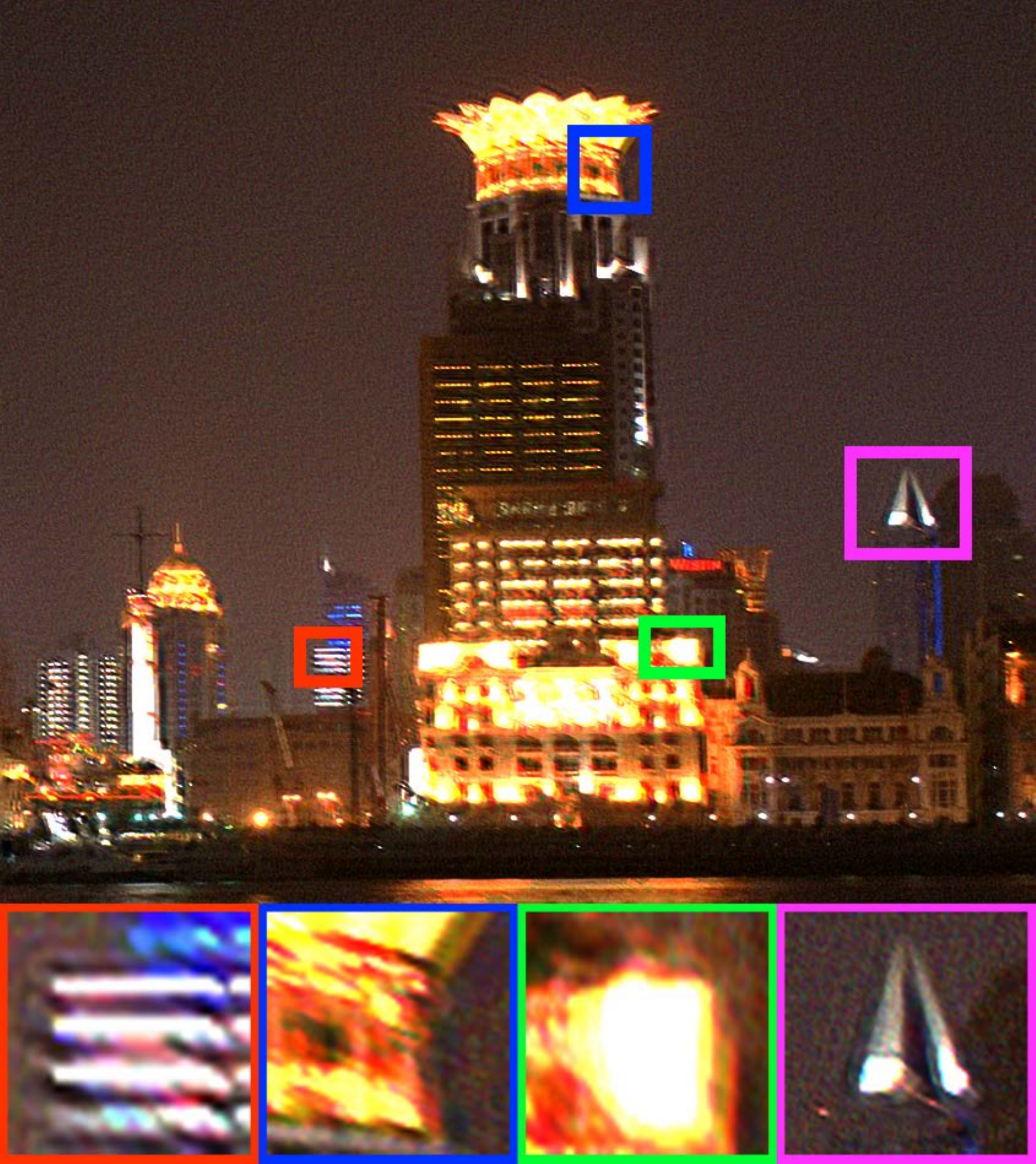}}
		\centerline{(c) Whyte \etal~\cite{Whyte14deblurring}}
	\end{minipage}
	\begin{minipage}[b]{0.16\linewidth}
		\centering
		\centerline{
			\includegraphics[width =\linewidth]{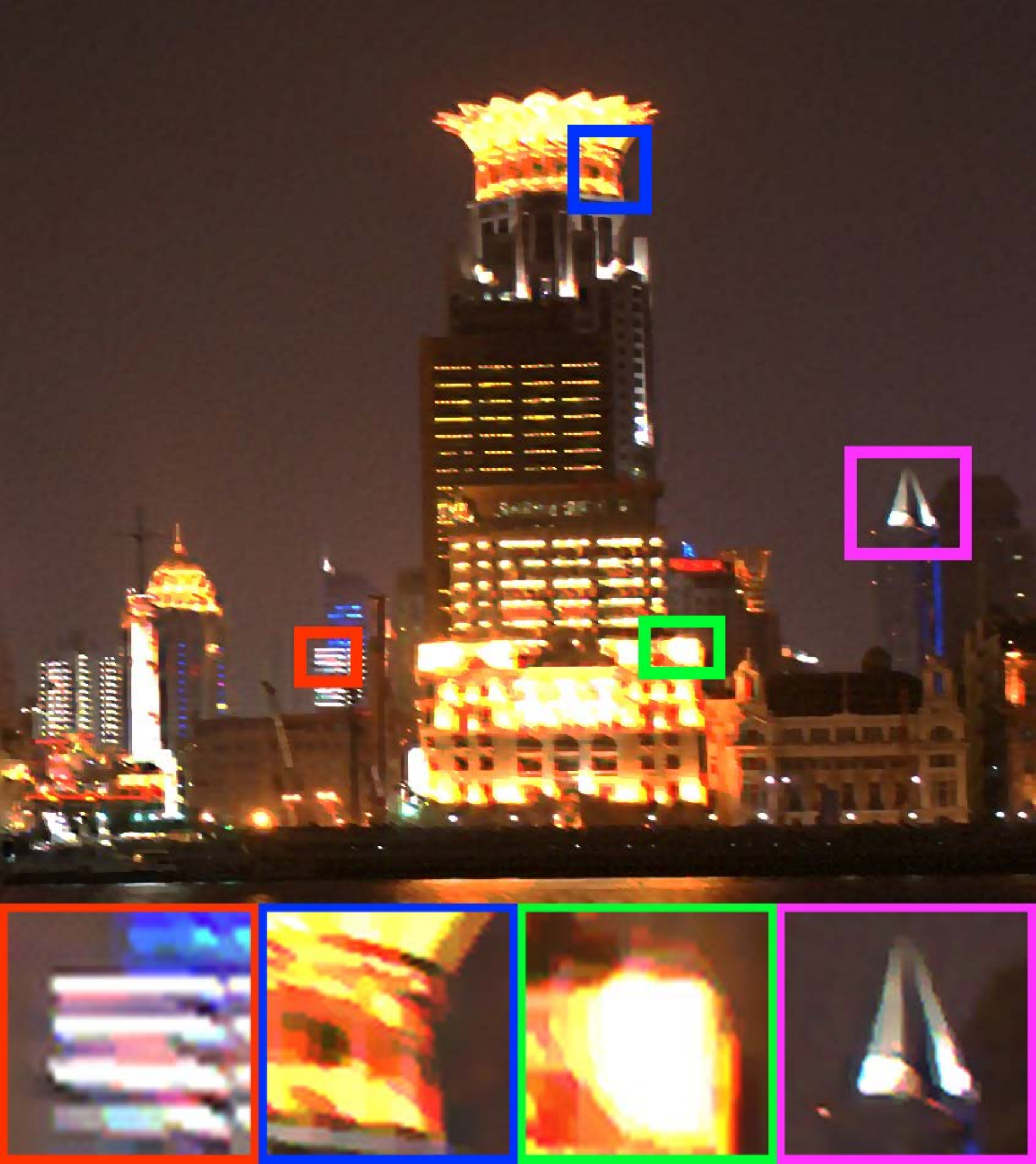}}
		\centerline{(d) Hu \etal~\cite{hu2014deblurring}}
	\end{minipage}
	\begin{minipage}[b]{0.16\linewidth}
		\centering
		\centerline{
			\includegraphics[width =\linewidth]{image/figset/hu.pdf}}
		\centerline{(e) Pan \etal~\cite{pan2016robust}}
	\end{minipage}
	\begin{minipage}[b]{0.16\linewidth}
		\centering
		\centerline{
			\includegraphics[width =\linewidth]{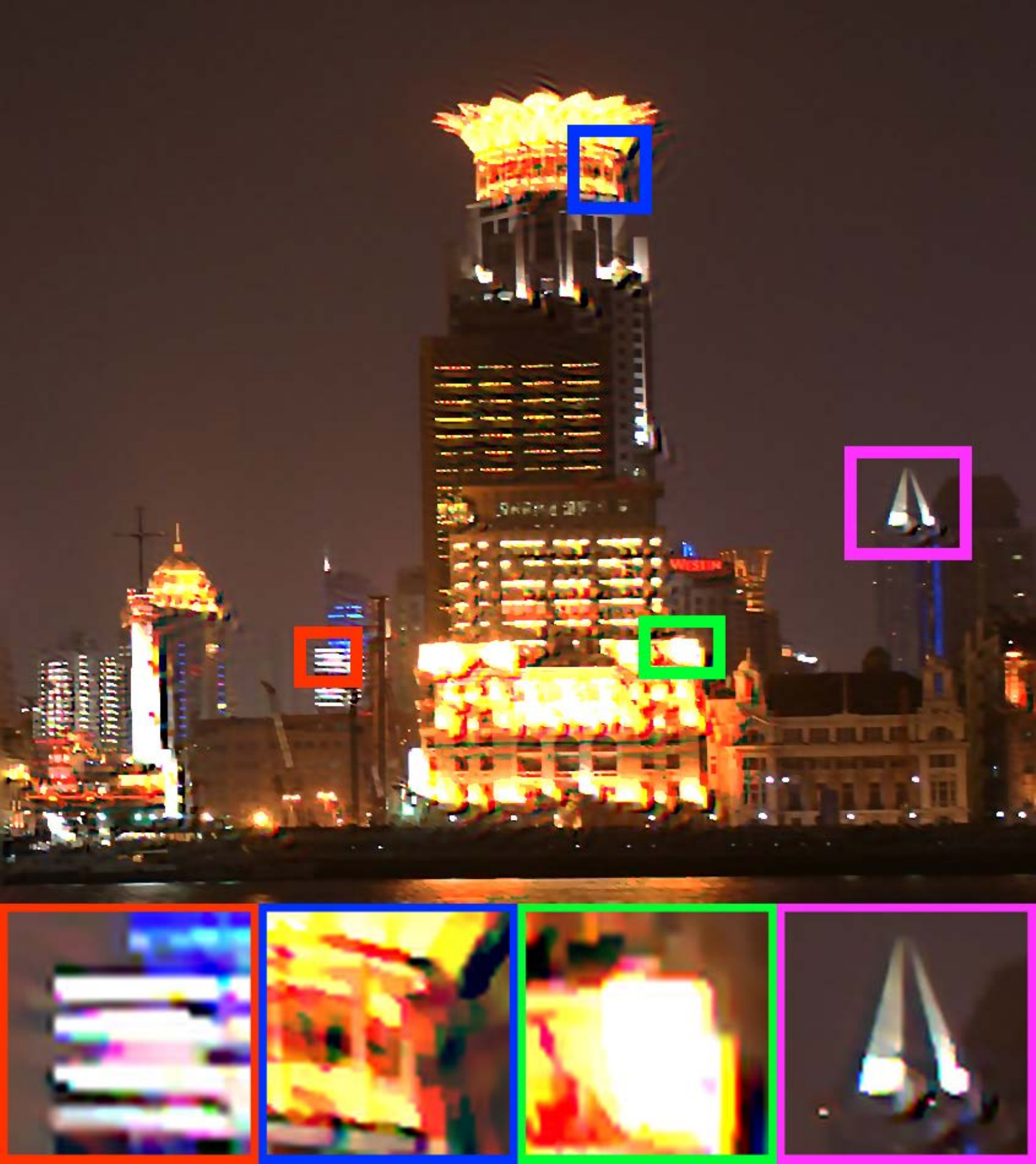}}
		\centerline{(f) Chen \etal~\cite{chen2020oid}}
	\end{minipage}\\
	\begin{minipage}[b]{0.16\linewidth}
		\centering
		\centerline{
			\includegraphics[width =\linewidth]{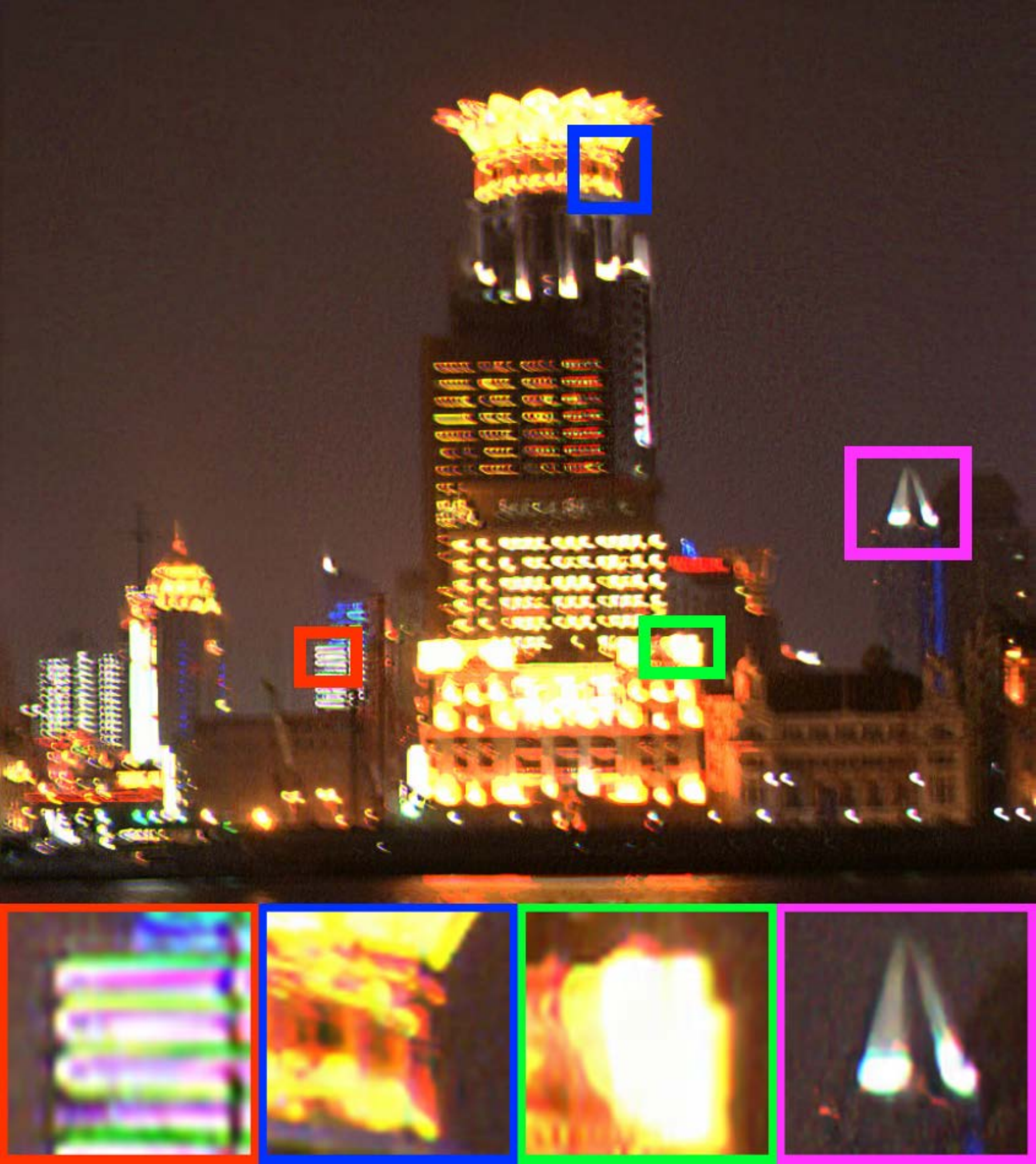}}
		\centerline{(g) SRN~\cite{tao2018scale}}
	\end{minipage}
	\begin{minipage}[b]{0.16\linewidth}
		\centering
		\centerline{
			\includegraphics[width =\linewidth]{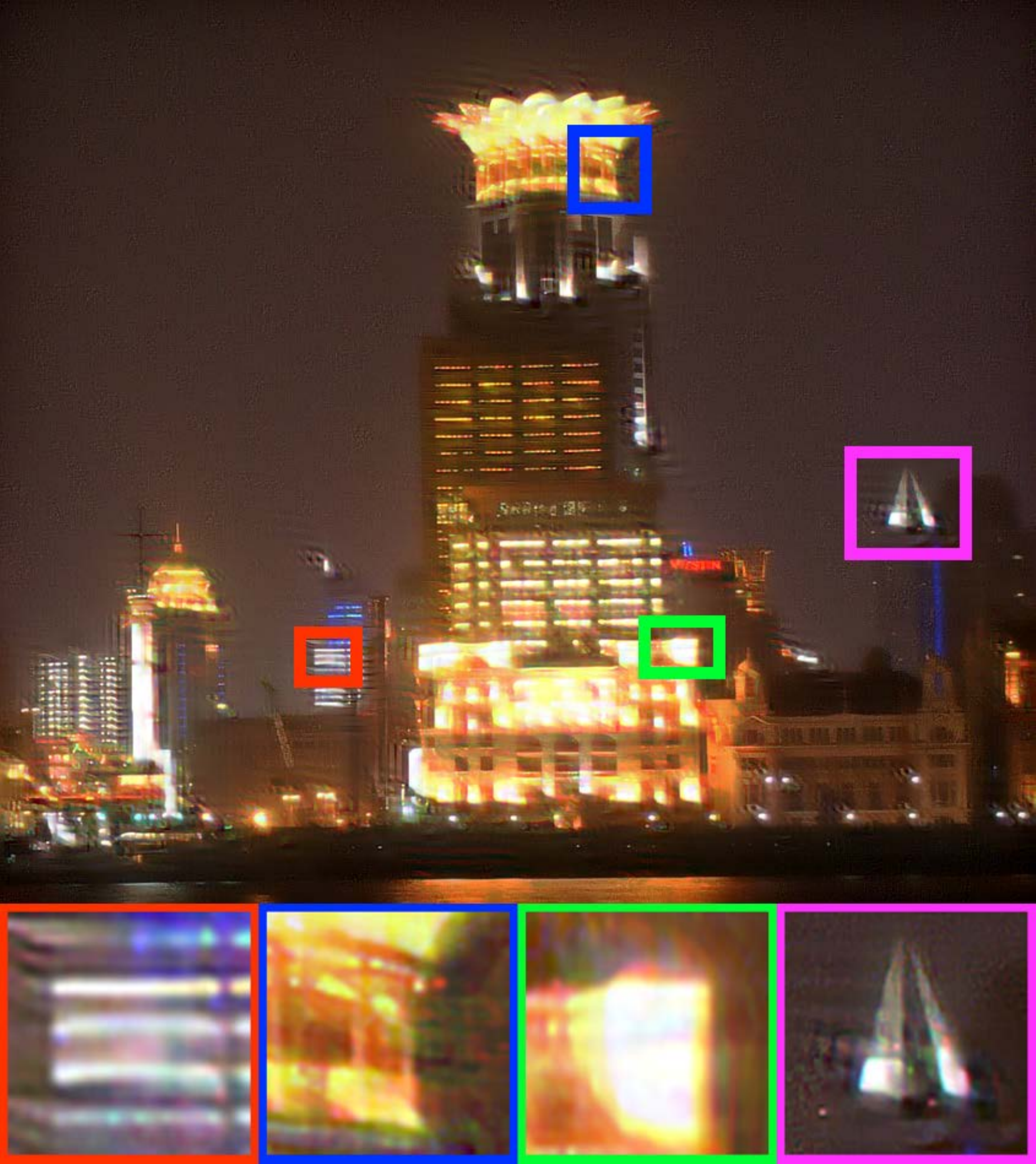}}
		\centerline{(h) FCNN~\cite{Zhang_2017_CVPR}}
	\end{minipage}
	\begin{minipage}[b]{0.16\linewidth}
		\centering
		\centerline{
			\includegraphics[width =\linewidth]{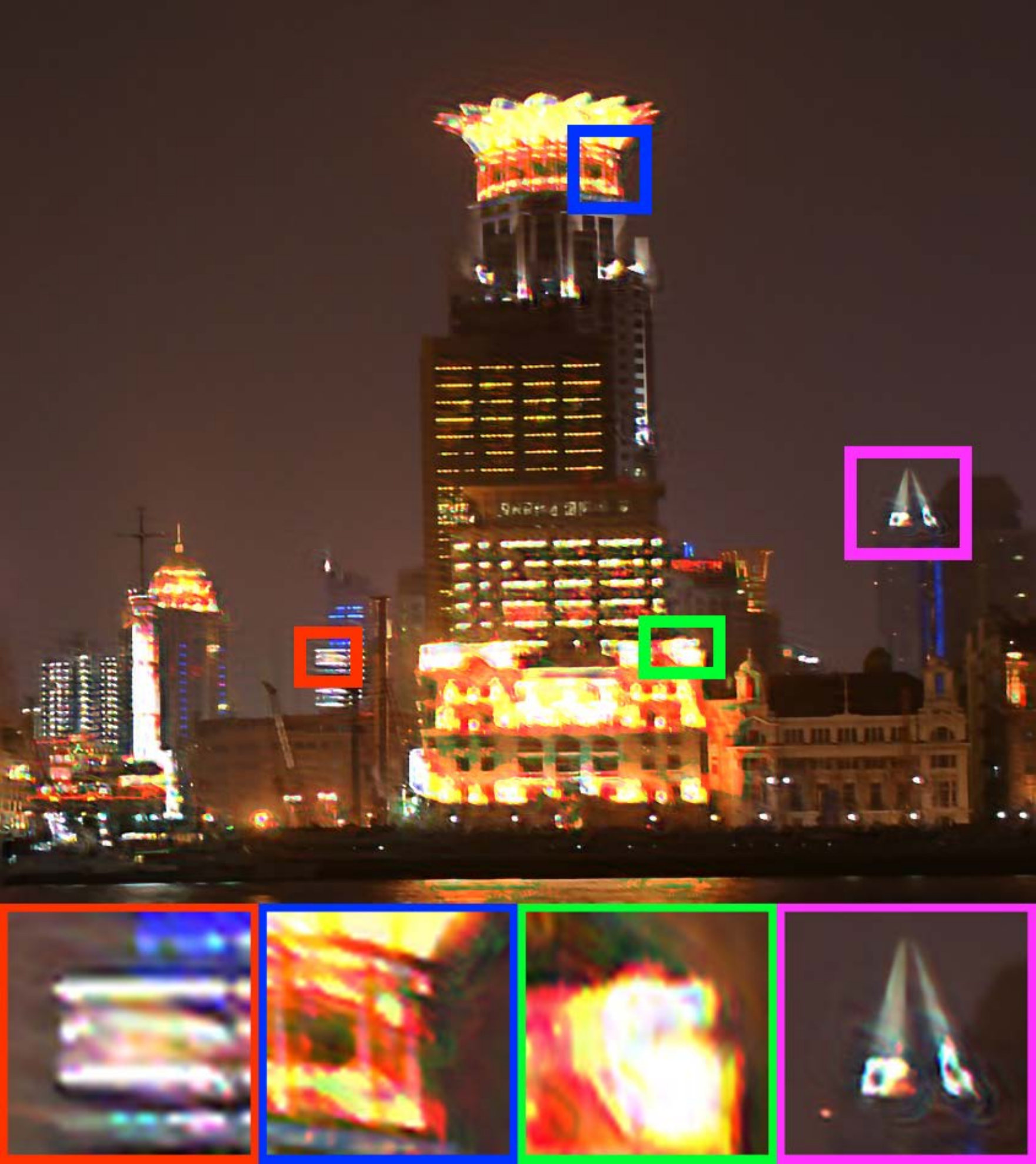}}
		\centerline{(i) NBDN~\cite{chen2021learning}}
	\end{minipage}
	\begin{minipage}[b]{0.16\linewidth}
		\centering
		\centerline{
			\includegraphics[width =\linewidth]{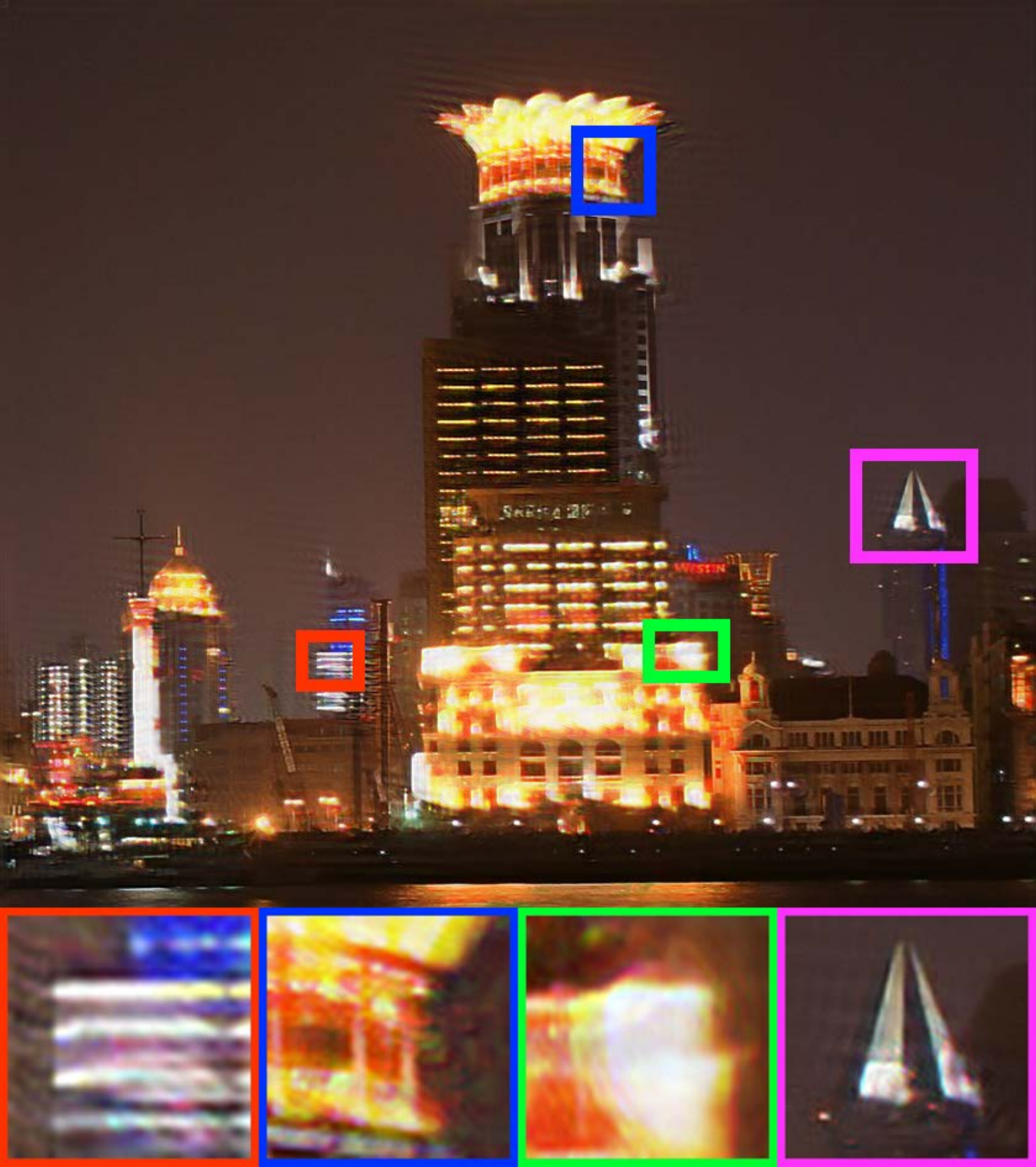}}
		\centerline{(j) DWDN~\cite{dong2020deep}}
	\end{minipage}
	\begin{minipage}[b]{0.16\linewidth}
		\centering
		\centerline{
			\includegraphics[width =\linewidth]{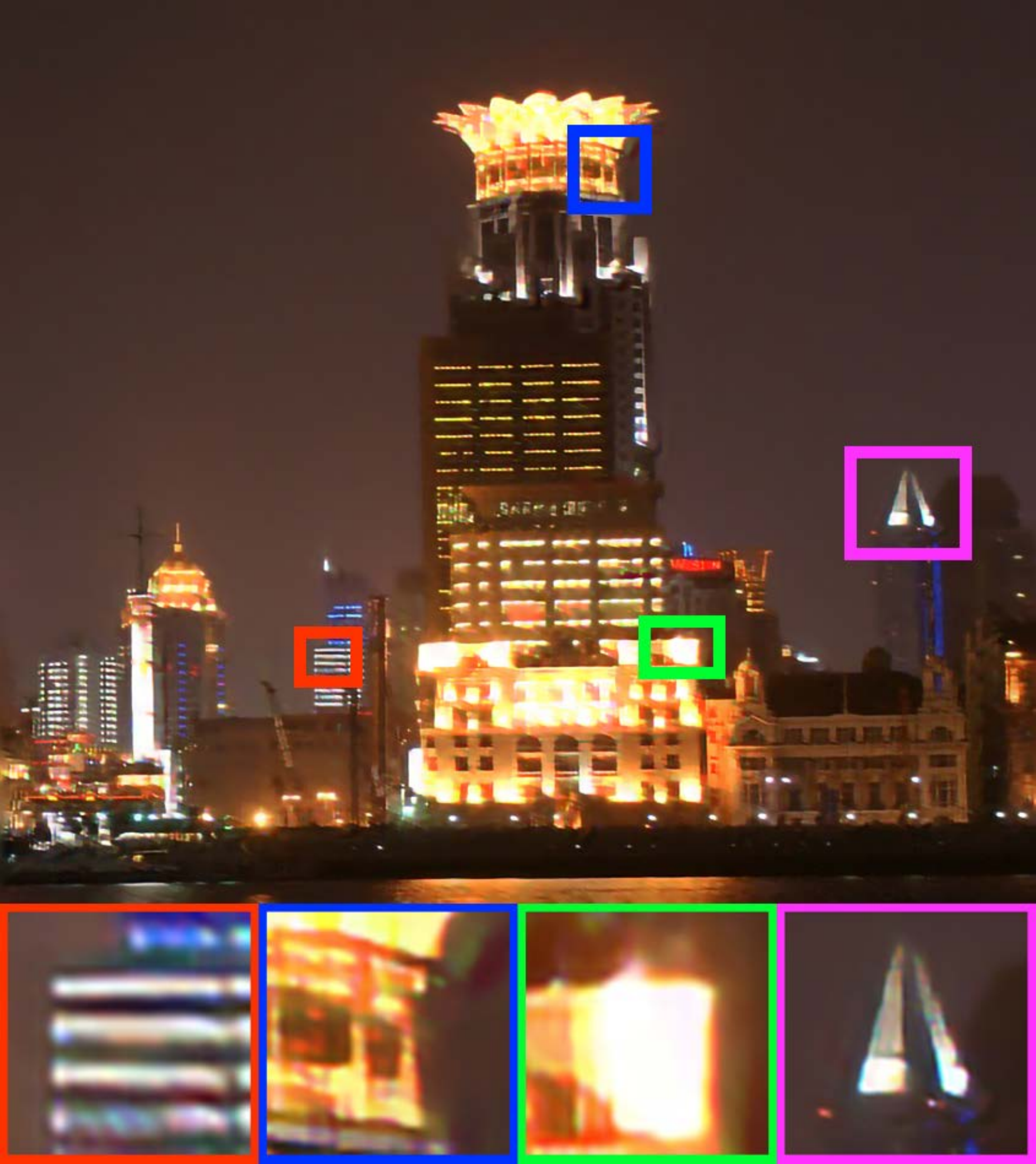}}
		\centerline{(k) Ours}
	\end{minipage}
	\begin{minipage}[b]{0.16\linewidth}
		\centering
		\centerline{
			\includegraphics[width =\linewidth]{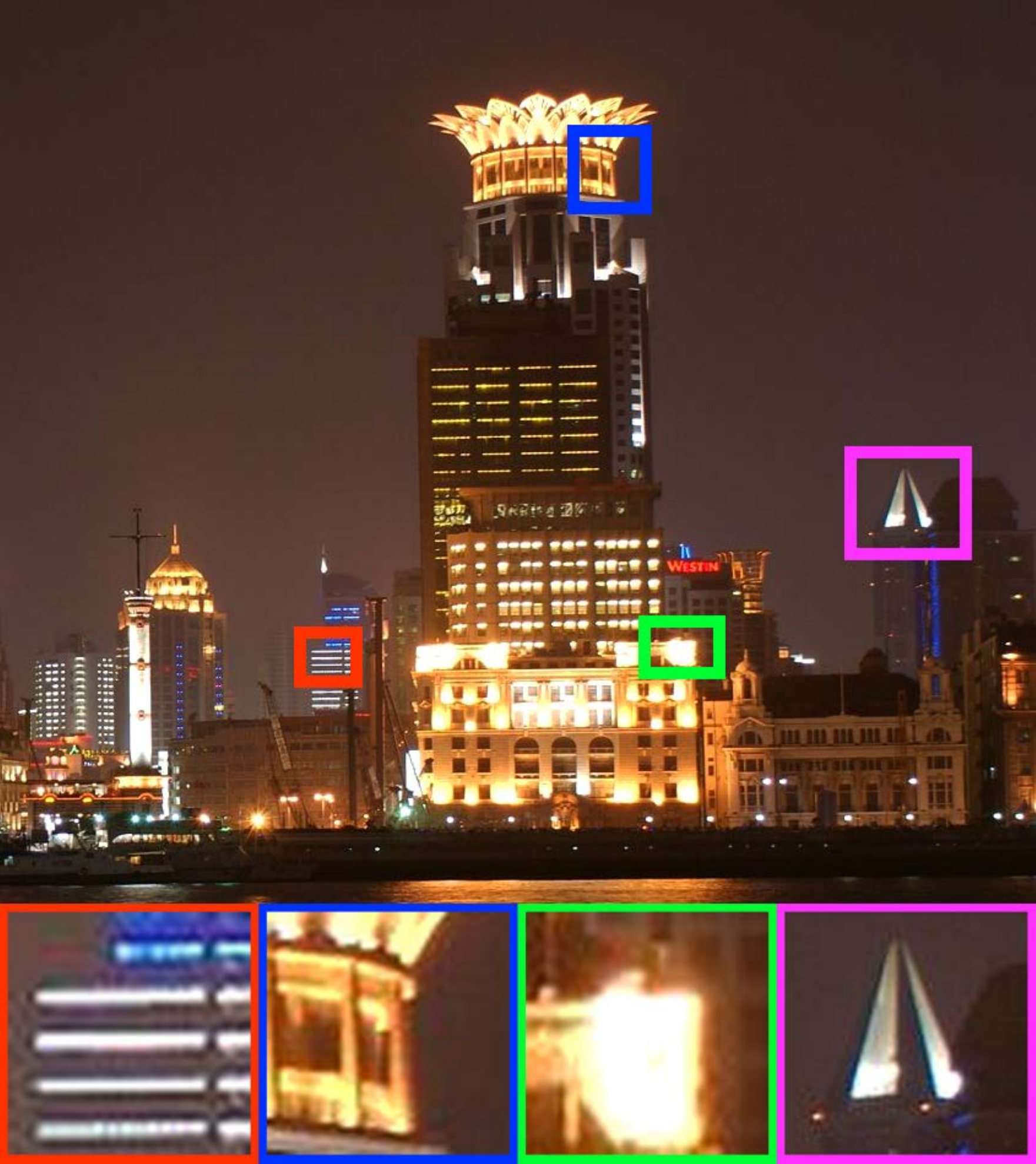}}
		\centerline{(l) GT}
	\end{minipage}
	\caption{Deblurring results of a saturated example from the proposed testing set.
Our method performs favorably compared with existing non-blind deblurring methods, which generates a result with finer details and fewer artifacts in the saturated regions. Please zoom-in for a better view.}
	\label{fig dataset}
	\vspace{-0.3 cm}
\end{figure*}

\section{Experiments}
In this section, we use both synthetic data and real-world saturated blurry images to evaluate the effectiveness of our method against state-of-the-art ones for non-blind deblurring.

\subsection{Datasets}
\noindent\textbf{Training datasets.}
We use the training dataset from~\cite{chen2021learning} which contains 500 night images from Flickr.
Then we randomly crop 10 patches of size 256 $\times$ 256 pixels from each of these images.
To generate saturated blurry images, we follow the strategy in~\cite{hu2014deblurring}.
The pixels in patches, which are larger than a threshold, are enlarged by $N$ times to obtain the saturated sharp patches\footnote{The dynamic range of all the images is [0, 1] in this paper. The threshold and $N$ are randomly sampled from 0.75-0.95 and 1.5-5.}.
In order to simulate motion blur, we use~\cite{chakrabarti2016neural} to generate 5 motion kernels with sizes ranging from 11 to 33 pixels for every patch.
Thus, a total of 25000 blurry and sharp image pairs are used in our training process.
We clip the obtained sharp and blurred patches after convolving the sharp patches by the corresponding blur kernel.
Merits of our training data synthesizing strategy can be found in Section~\ref{sec syndata}.

\noindent\textbf{Testing dataset.}
For the testing data, we use the 100 test images from~\cite{chen2021learning} without cropping to generate the blurry images in the same way as the training data. For every image, one motion kernel is randomly generated. The training data and test data do not overlap.
For the 100 blurry images in the testing set, we compare different methods with the ground truth kernels and the estimated kernels from~\cite{hu2014deblurring}.
Moreover, we also test our methods in the saturate dataset provided in~\cite{hu2014deblurring} and the unsaturate benchmark dataset from~\cite{levin2009understanding}.
We further use real-world examples to evaluate different arts.

\begin{table*}
\centering
\caption{Quantitative evaluations on non-saturated data from Levin \etal~\cite{levin2009understanding} and extra saturated data from Hu \etal~\cite{hu2014deblurring}}
    \centering
    \scalebox{0.92}{
    \begin{tabular}{cccccccccccc}
    \toprule
    & Cho~\cite{cho2011outlier}
    & Hu ~\cite{hu2014deblurring}
    & Whyte~\cite{Whyte14deblurring}
    & Pan~\cite{pan2016robust}
    & Chen~\cite{chen2020oid}
    & SRN~\cite{tao2018scale}
    & FCNN~\cite{Zhang_2017_CVPR}
    & IRCNN~\cite{kaiZhang_2017_CVPR}
    & RGDN~\cite{gong2018learning}
    & NBDN~\cite{chen2021learning}
    & Ours\\
    \midrule
    \multicolumn{12}{c}{Results on the non-saturated data from Levin \etal~\cite{levin2009understanding}}\\
    \midrule
    PSNR & 32.59 &31.27 &26.54 & 32.03 &33.03 &31.18 &33.22 &33.14 &33.86
    &32.91 &\textbf{34.12}\\
   SSIM &0.9135 &0.8930 &0.8301 & 0.9263 &0.9054 &0.8972 &0.9267 &0.9261 &\textbf{0.9335} 	
   &0.9000 &0.9329\\
    \midrule
    \multicolumn{12}{c}{Results on the extra saturated data from Hu \etal~\cite{hu2014deblurring}}\\
    \midrule
    PSNR & 20.94 &22.79 &21.43 & 22.47 &21.80 &22.21 &23.57 &19.68 &21.62
    &24.60 &\textbf{25.11}\\
    SSIM & 0.7022 &0.7764 &0.7414 &0.7583 &0.7014 &0.7526 &0.8004 &0.6147 &0.6592
    &0.8274 &\textbf{0.8439}\\
    \bottomrule
    \end{tabular}}
    \label{tab moredata}
\end{table*}

\begin{figure*}
\centering
	\begin{minipage}[b]{0.161\linewidth}
		\centering
		\centerline{
			\includegraphics[width =\linewidth]{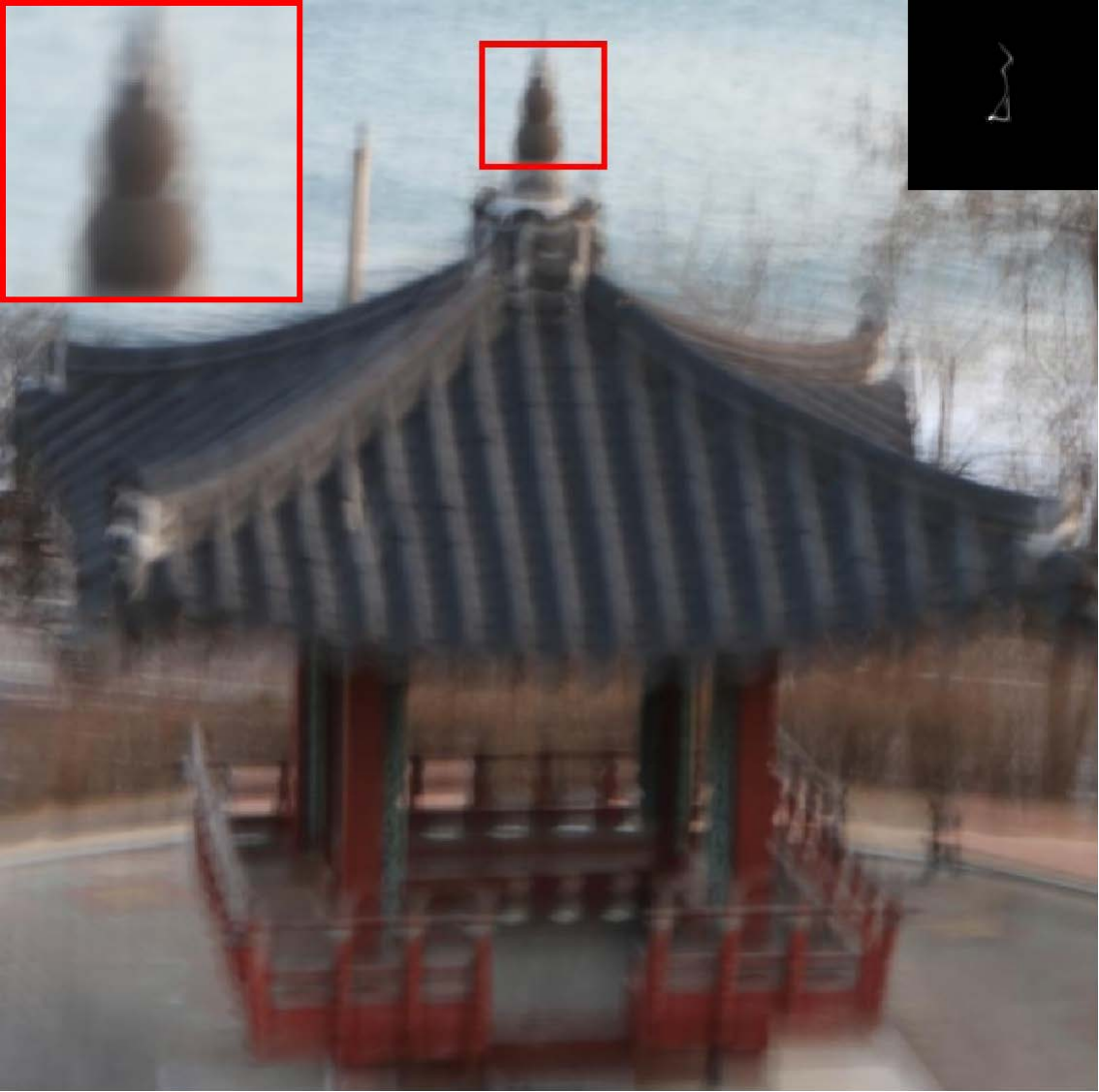}}
		\centerline{(a) Blurry image}
	\end{minipage}
	\begin{minipage}[b]{0.161\linewidth}
		\centering
		\centerline{
			\includegraphics[width =\linewidth]{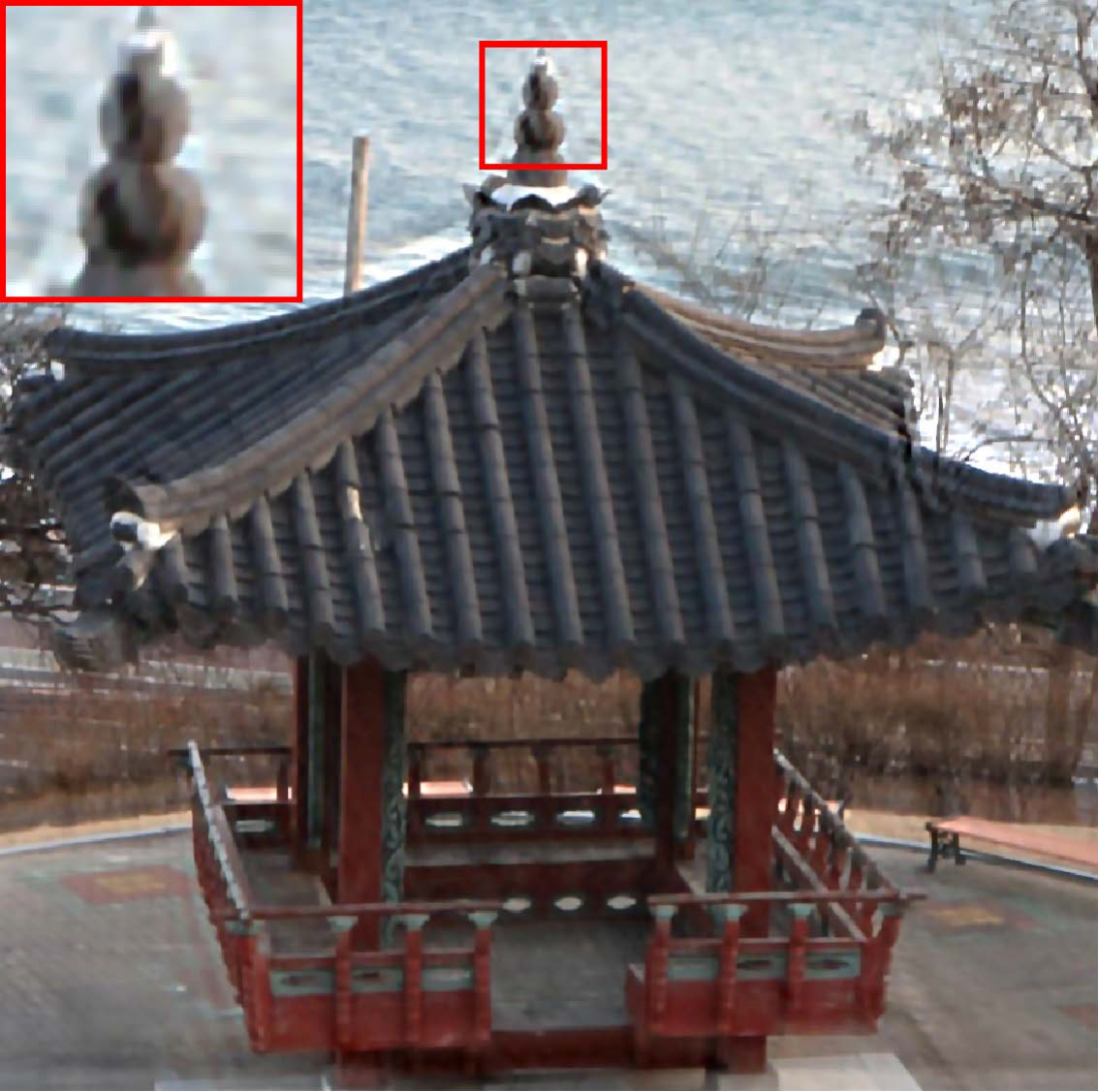}}
		\centerline{(b) Cho \etal~\cite{cho2011outlier}}
	\end{minipage}
	\begin{minipage}[b]{0.161\linewidth}
		\centering
		\centerline{
			\includegraphics[width =\linewidth]{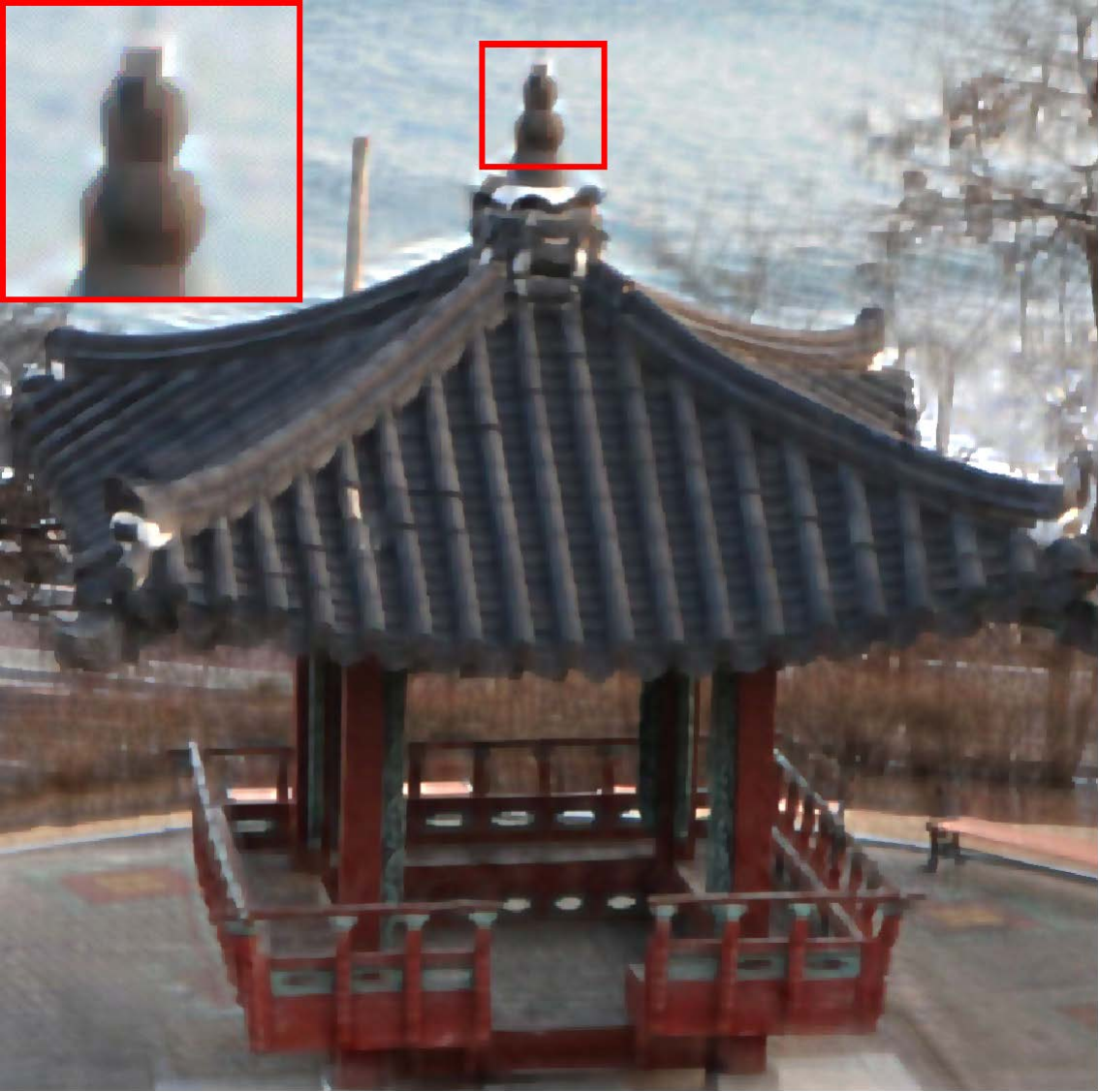}}
		\centerline{(c) Hu \etal~\cite{hu2014deblurring}}
	\end{minipage}
	\begin{minipage}[b]{0.161\linewidth}
		\centering
		\centerline{
			\includegraphics[width =\linewidth]{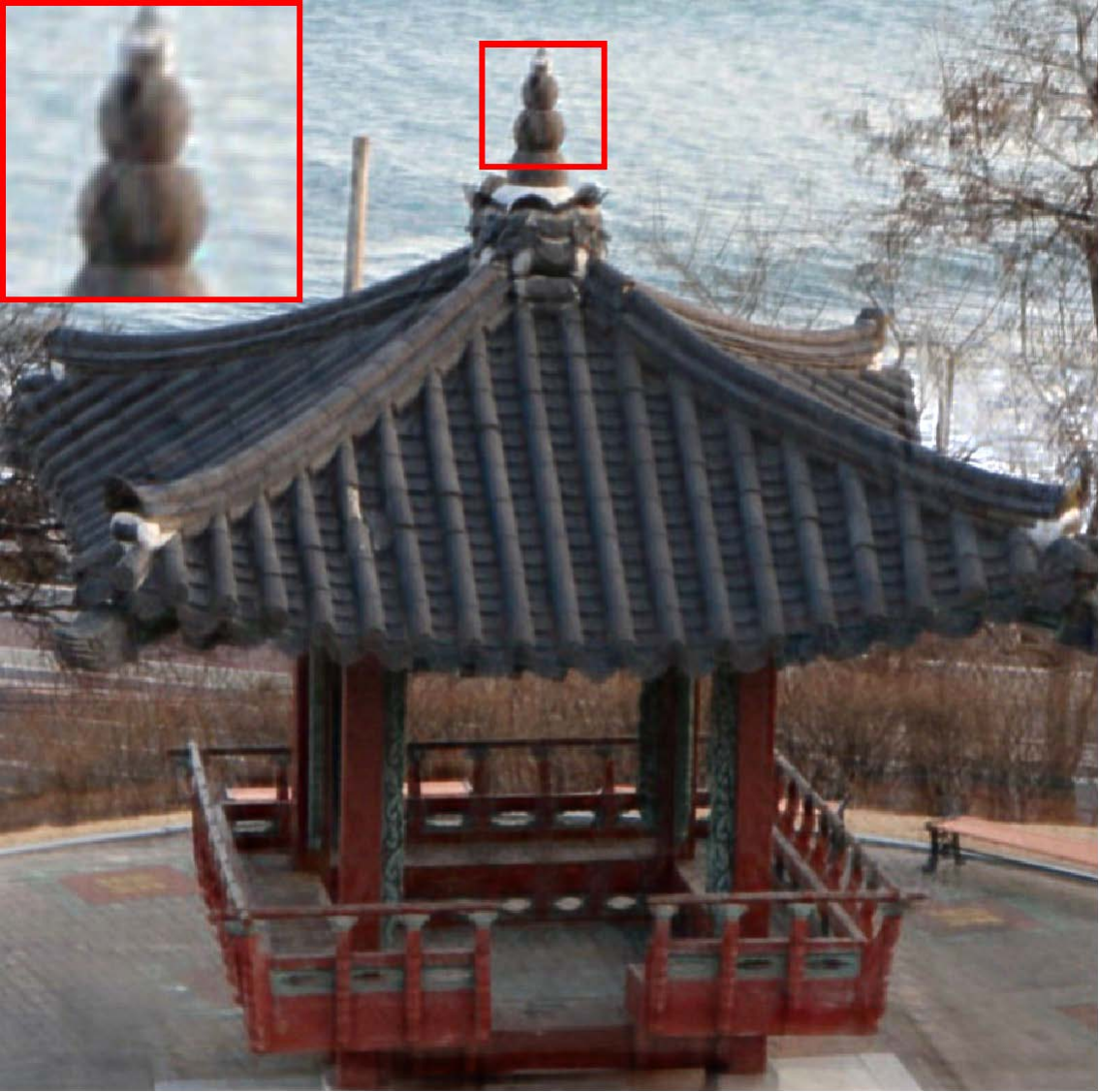}}
		\centerline{(d) Whyte \etal~\cite{Whyte14deblurring}}
	\end{minipage}
	\begin{minipage}[b]{0.161\linewidth}
		\centering
		\centerline{
			\includegraphics[width =\linewidth]{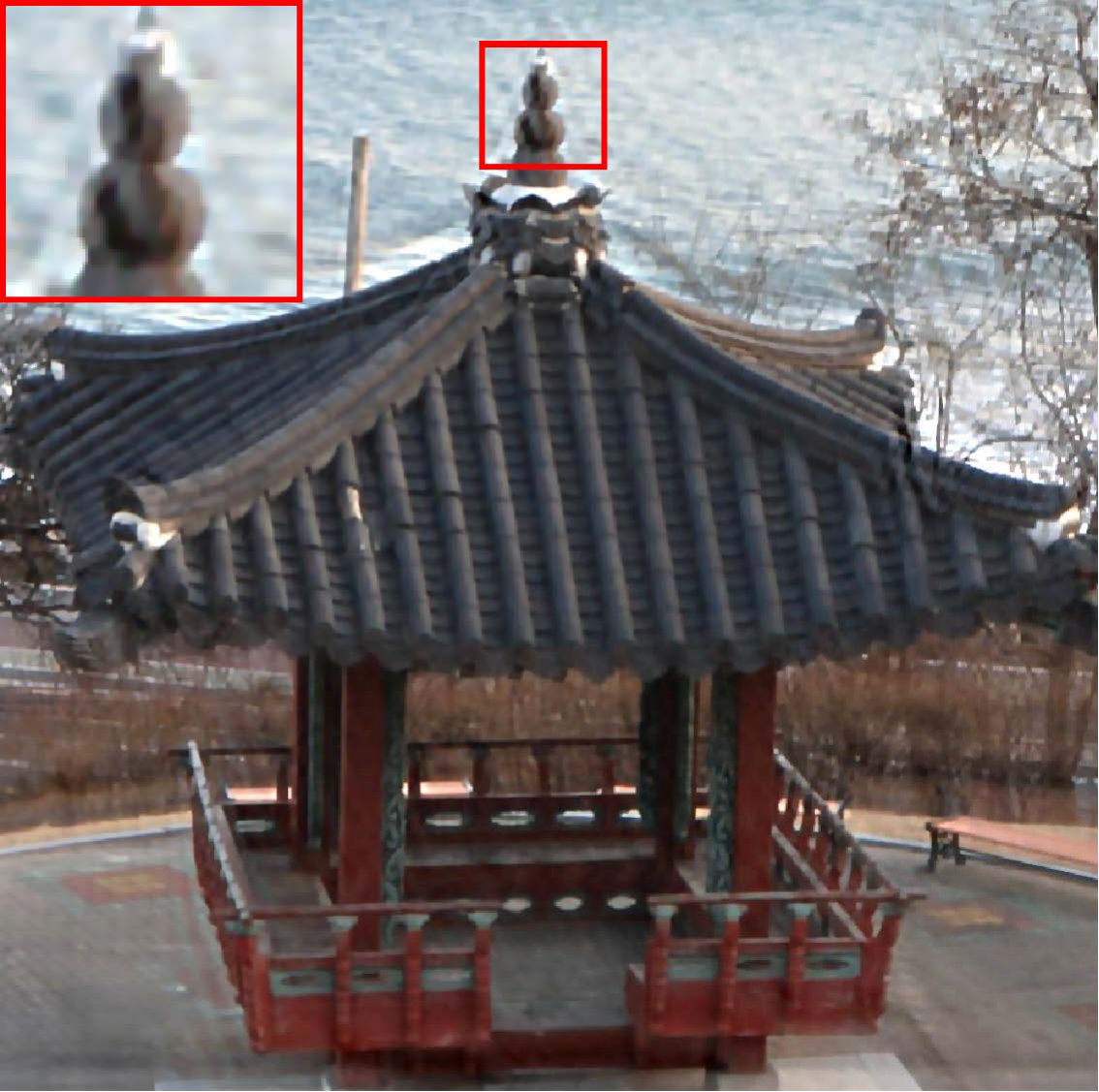}}
		\centerline{(e) Pan \etal~\cite{pan2016robust}}
	\end{minipage}
	\begin{minipage}[b]{0.161\linewidth}
		\centering
		\centerline{
			\includegraphics[width =\linewidth]{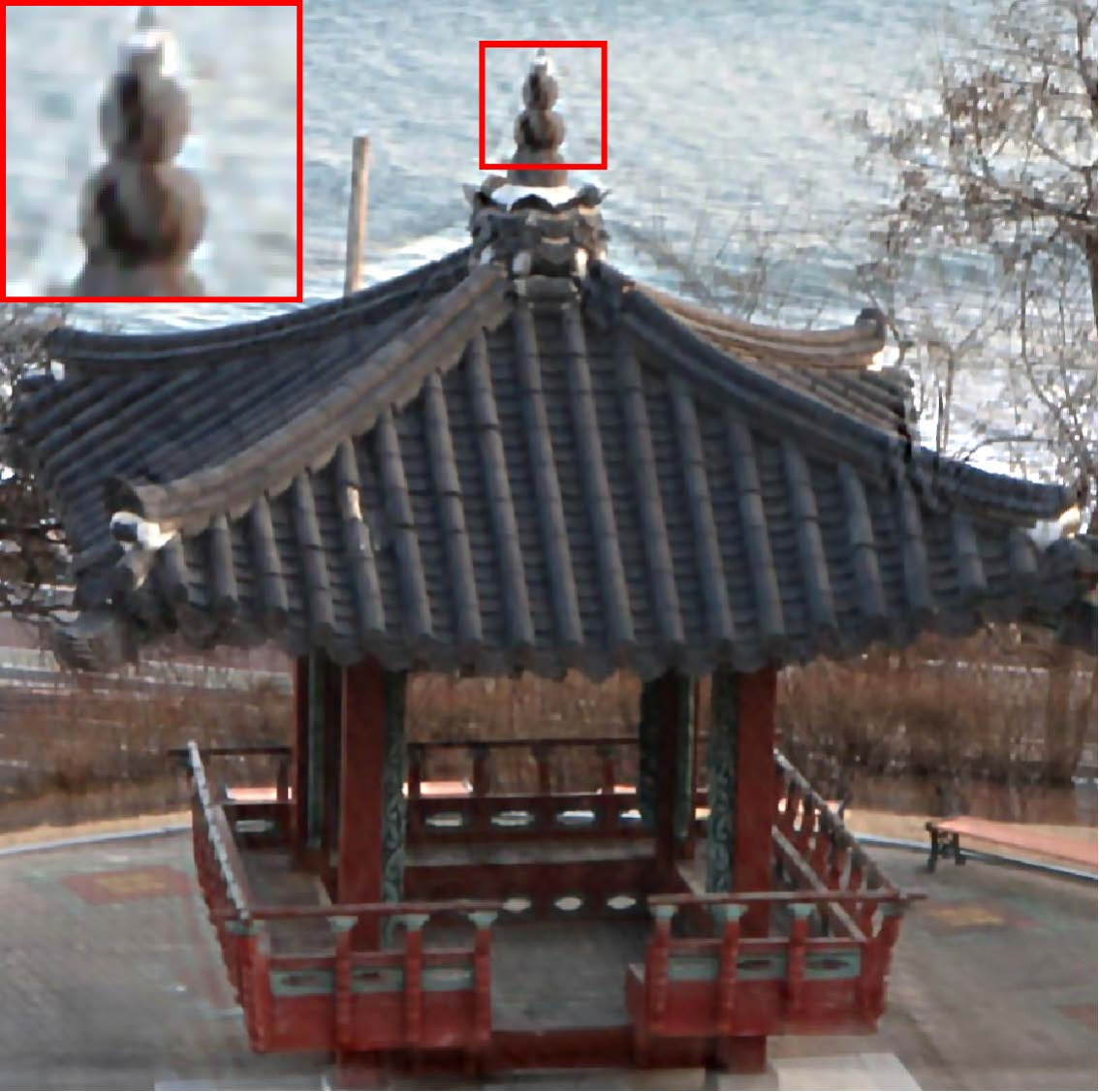}}
		\centerline{(f) Chen \etal~\cite{chen2020oid}}
	\end{minipage}\\
	\begin{minipage}[b]{0.161\linewidth}
		\centering
		\centerline{
			\includegraphics[width =\linewidth]{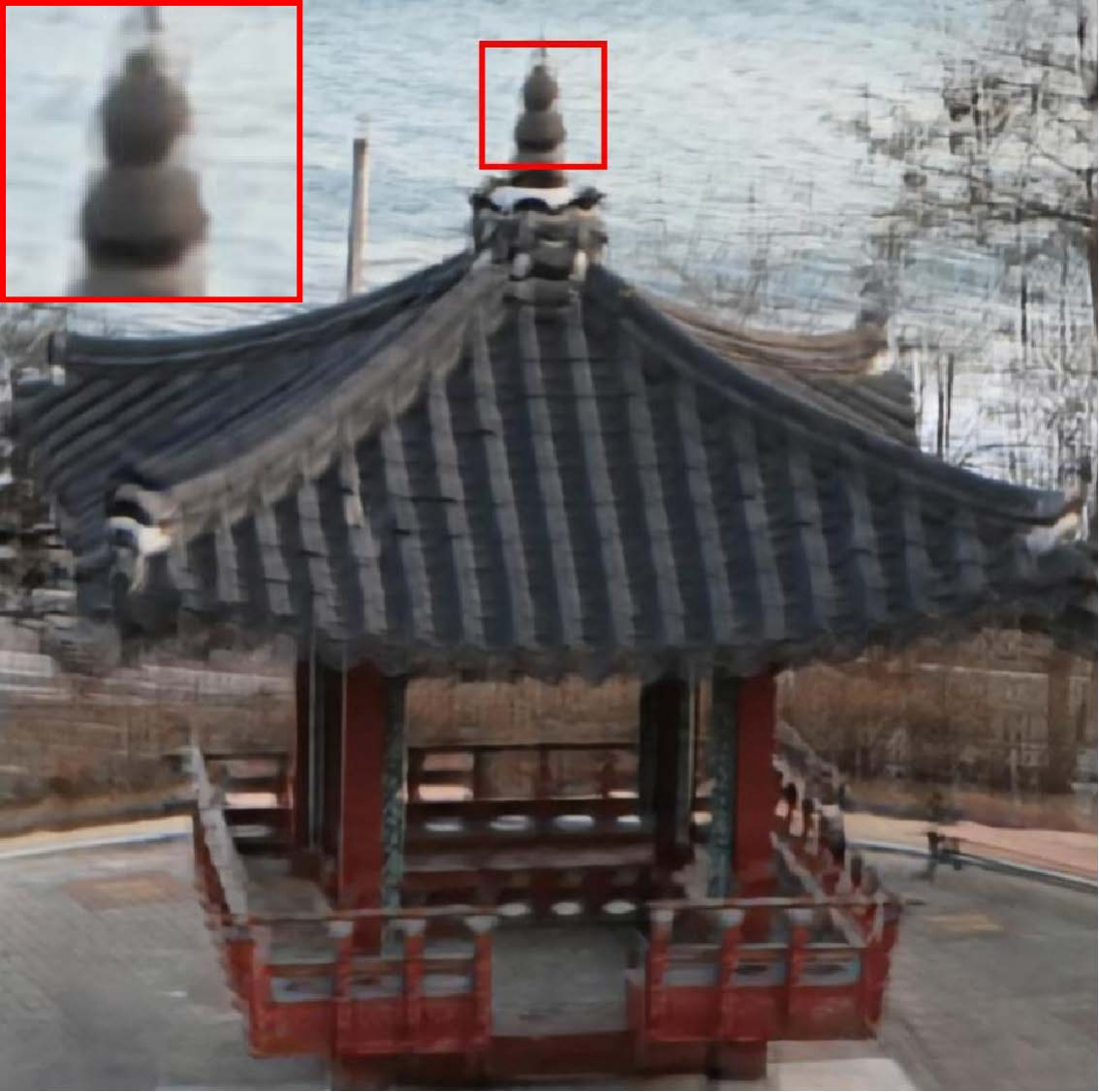}}
		\centerline{(g) Stripformer~\cite{tsai2022stripformer}}
	\end{minipage}
	\begin{minipage}[b]{0.161\linewidth}
		\centering
		\centerline{
			\includegraphics[width =\linewidth]{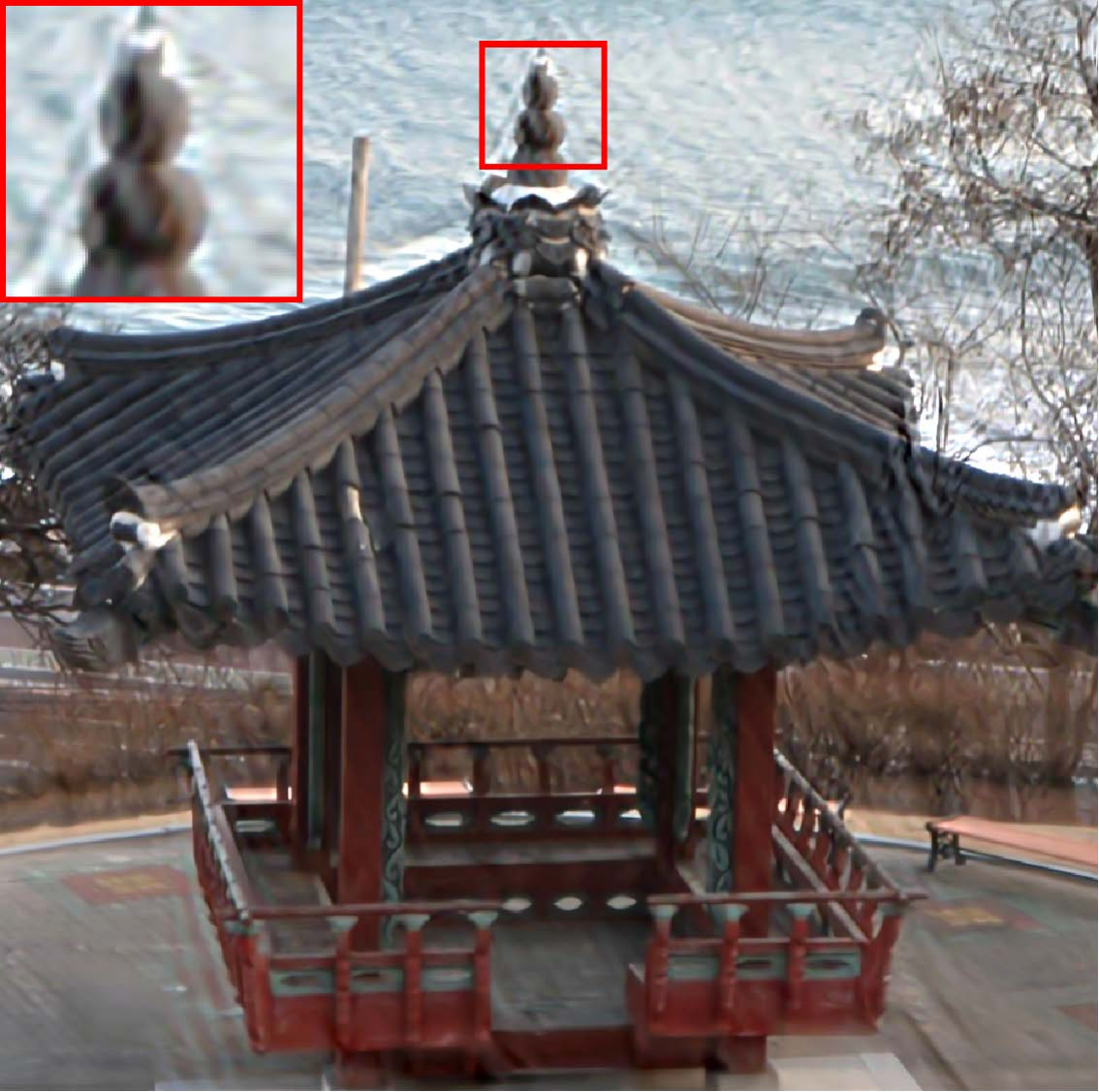}}
		\centerline{(h) IRCNN~\cite{kaiZhang_2017_CVPR}}
	\end{minipage}
	\begin{minipage}[b]{0.161\linewidth}
		\centering
		\centerline{
			\includegraphics[width =\linewidth]{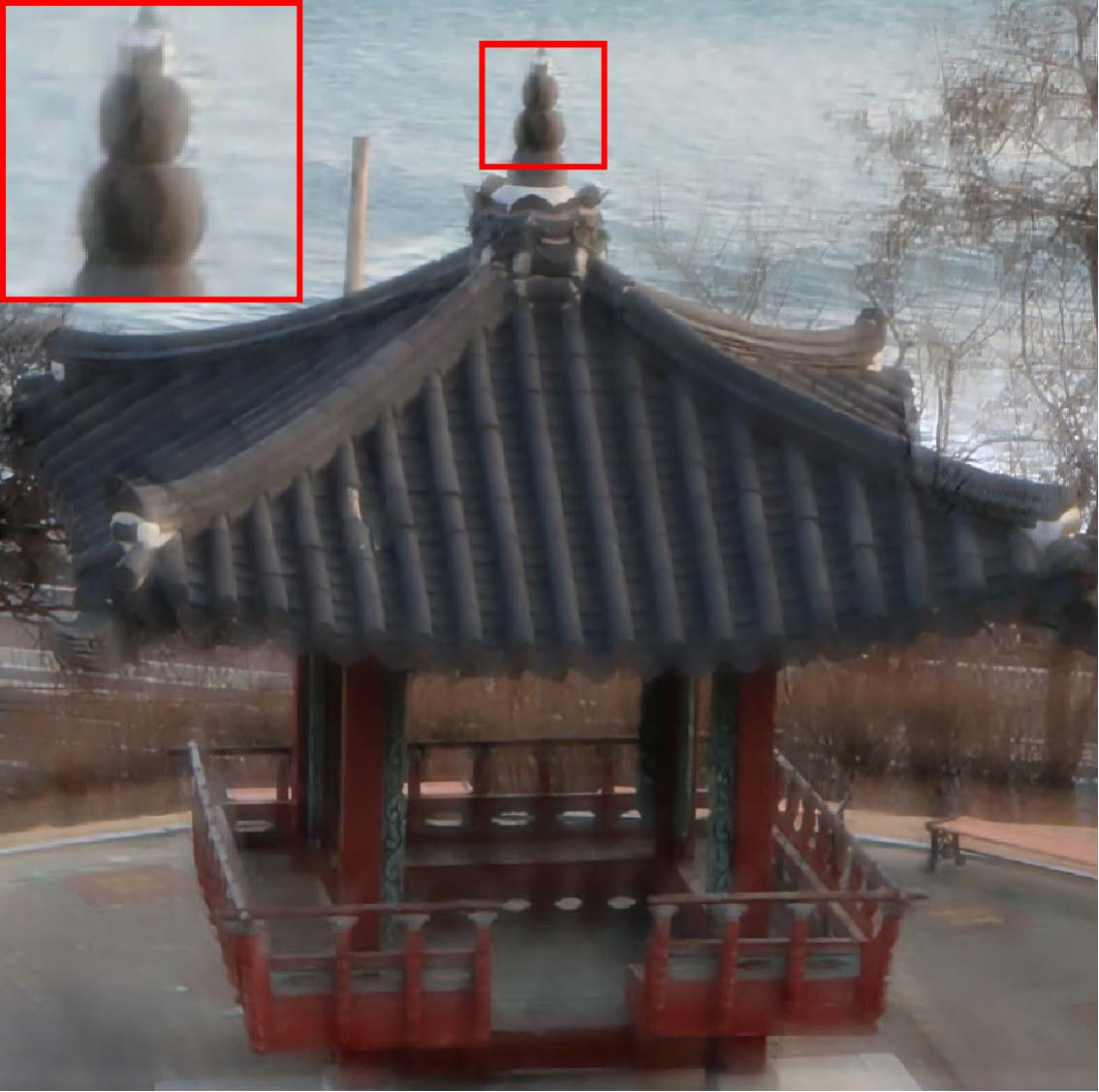}}
		\centerline{(i) FCNN~\cite{Zhang_2017_CVPR}}
	\end{minipage}
	\begin{minipage}[b]{0.161\linewidth}
		\centering
		\centerline{
			\includegraphics[width =\linewidth]{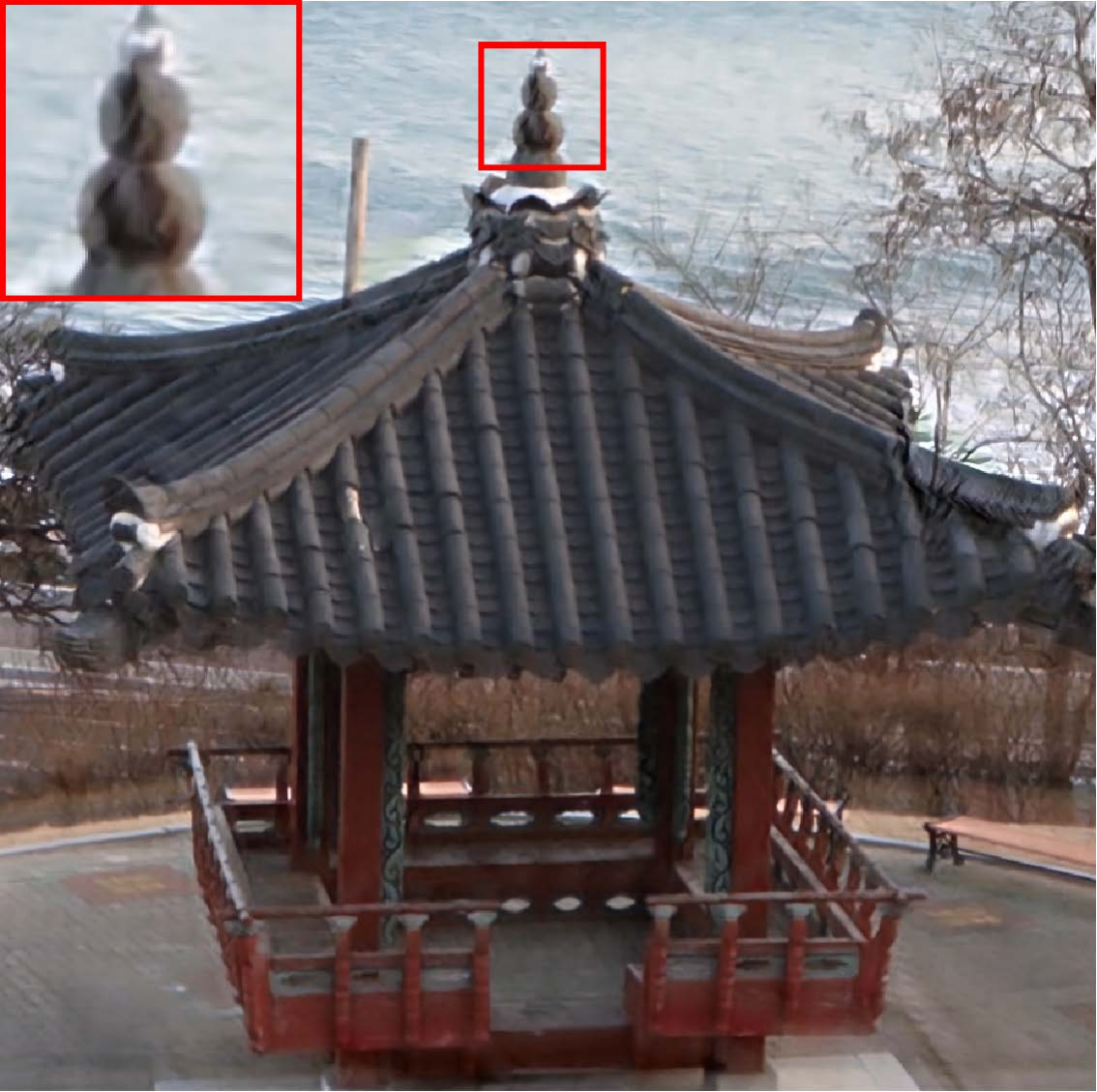}}
		\centerline{(j) RGDN~\cite{gong2018learning}}
	\end{minipage}
	\begin{minipage}[b]{0.161\linewidth}
		\centering
		\centerline{
			\includegraphics[width =\linewidth]{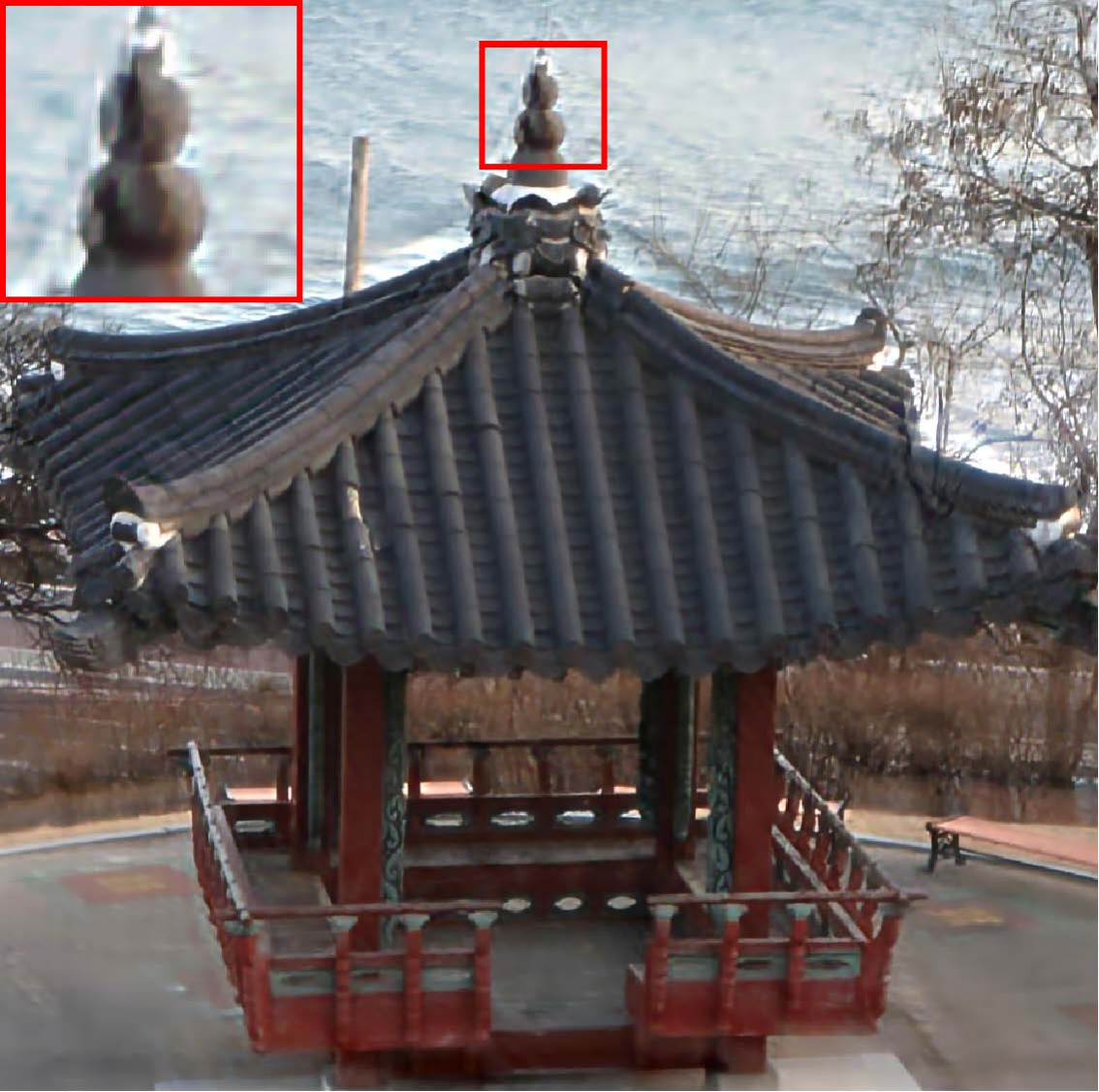}}
		\centerline{(k) NBDN~\cite{chen2021learning}}
	\end{minipage}
	\begin{minipage}[b]{0.161\linewidth}
		\centering
		\centerline{
			\includegraphics[width =\linewidth]{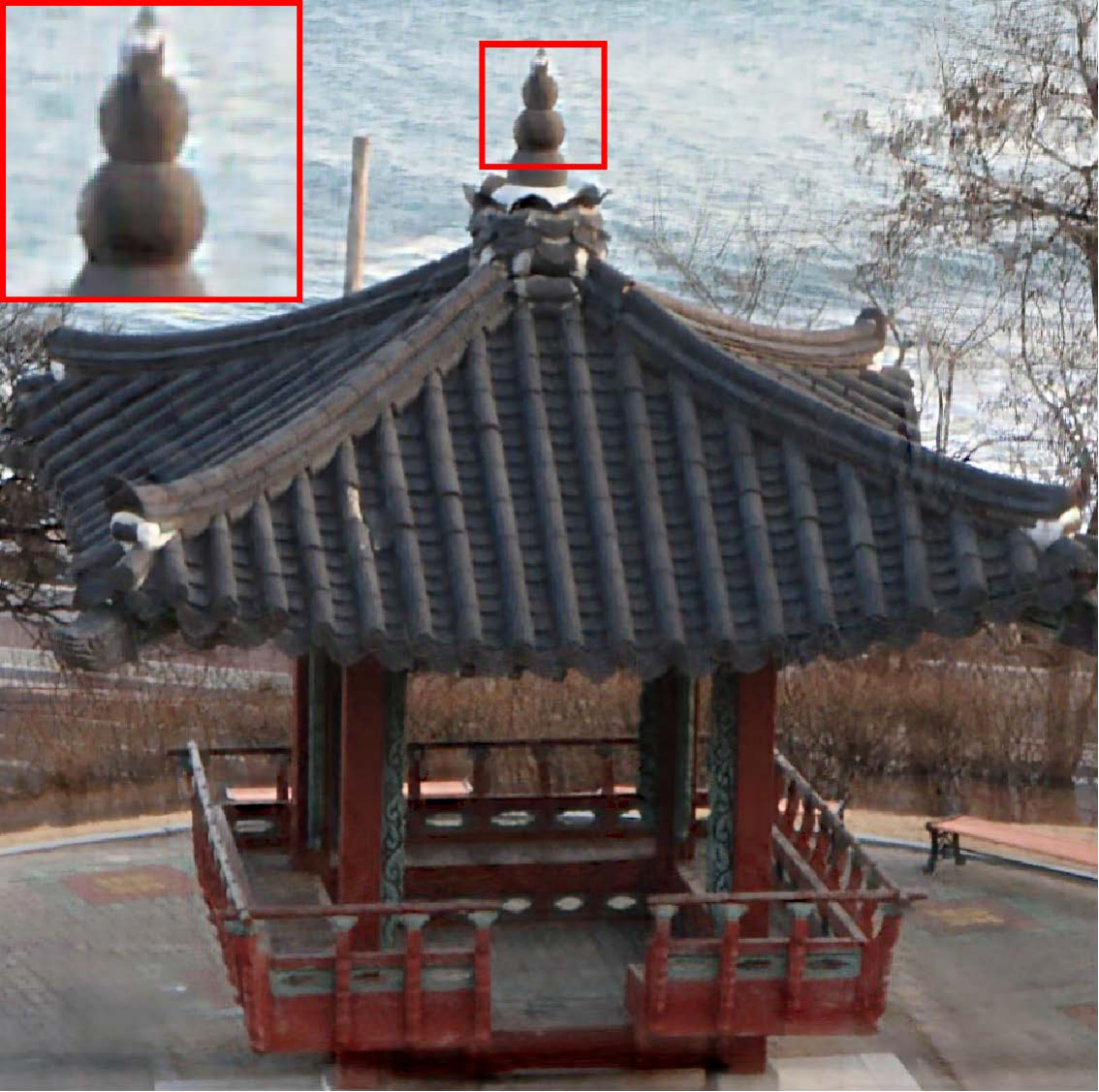}}
		\centerline{(l) Ours}
	\end{minipage}
\caption{Deblurring results of an unsaturated example. The kernel in (a) is from~\cite{pan2016blind}.
Although specially designed for saturated images, our method can still perform favorably against existing methods on the unsaturated image, which generates a result with finer details and fewer artifacts as shown in the boxes. Please zoom-in for a better view.}
	\label{fig unsat}
\end{figure*}

\subsection{Comparisons with State-of-the-Arts}
In this section, we compare our method with the optimization-based methods~\cite{cho2011outlier,Whyte14deblurring,hu2014deblurring,pan2016robust,chen2020oid} that are specially designed for saturated images and some recent learning-based arts, including FCNN~\cite{Zhang_2017_CVPR}, IRCNN~\cite{kaiZhang_2017_CVPR}, RGDN~\cite{gong2018learning}, DWDN~\cite{dong2020deep}, NBDN~\cite{chen2021learning}, SRN~\cite{tao2018scale}, Uformer~\cite{wang2022uformer}, and Stripformer~\cite{tsai2022stripformer}.
To ensure a fair comparison, we use the original implementation of these arts and finetune~\cite{Zhang_2017_CVPR,tao2018scale,dong2020deep,chen2021learning,wang2022uformer,tsai2022stripformer} in our dataset.
We also tune the hyper-parameters involved in the compared optimization-based arts for relatively better performances.

\noindent\textbf{Saturated image from the proposed testing set.}
We first evaluate different methods using the proposed testing set in terms of average PSNR and SSIM, and the results are demonstrated in Table~\ref{tab 1}.
For both ground truth and estimated kernels, our method performs the best among the models evaluated.

Fig.~\ref{fig dataset} shows visual results from different methods.
The deep learning-based approach~\cite{tao2018scale} performs less effectively than other methods due to the lack of blur kernel information and ignoring of the imaging process as shown in Fig.~\ref{fig dataset} (g).
The optimization-based methods~\cite{cho2011outlier,hu2014deblurring,pan2016robust,chen2020oid} contain artifacts in the saturated regions because some saturated pixels are not properly handled in their models (Fig.~\ref{fig dataset} (b), (d), (e) and (f)).
Moreover, the details are also not recovered well in their results.
This is mainly because of the ineffectiveness of the sparse image prior~\cite{Levin07image} that they adopted.
Artifacts can also be found in the result from~\cite{chen2021learning} (Fig.~\ref{fig dataset} (i)) because they suggest excluding saturated pixels in their deblurring process. This setting is ineffective because it is difficult to precisely separate saturated and unsaturated pixels in this example.
The algorithm based on Eq.~\eqref{eq blur_clip} (\ie Whyte \etal~\cite{Whyte14deblurring}) can remove the blur in the saturated regions to a certain extent as shown in Fig.~\ref{fig dataset}~(c).
However, due to the lack of prior information, their result still contains many ringings around the saturated regions.
The learning-based model~\cite{Zhang_2017_CVPR} shows their advantages in the regions without saturation due to the learned prior as shown in Fig.~\ref{fig dataset}~(h).
But they are based on the blur model that does not explicitly consider saturation (\eg Eq.~\eqref{eq blur}). It is not surprising that their result contains many artifacts around the saturated pixels. The same result can be observed from~\cite{dong2020deep} (Fig.~(j)).
In comparison, the saturated pixels have less influence on our method because of the robust blur model Eq.~\eqref{eq our}.
By taking advantage of the learned image prior, our method can restore a clear image with sharper edges and fewer artifacts around the saturated regions (Fig.~\ref{fig dataset} (k)).

\begin{figure*}
\centering
	\begin{minipage}[b]{0.16\linewidth}
		\centering
		\centerline{
			\includegraphics[width =\linewidth]{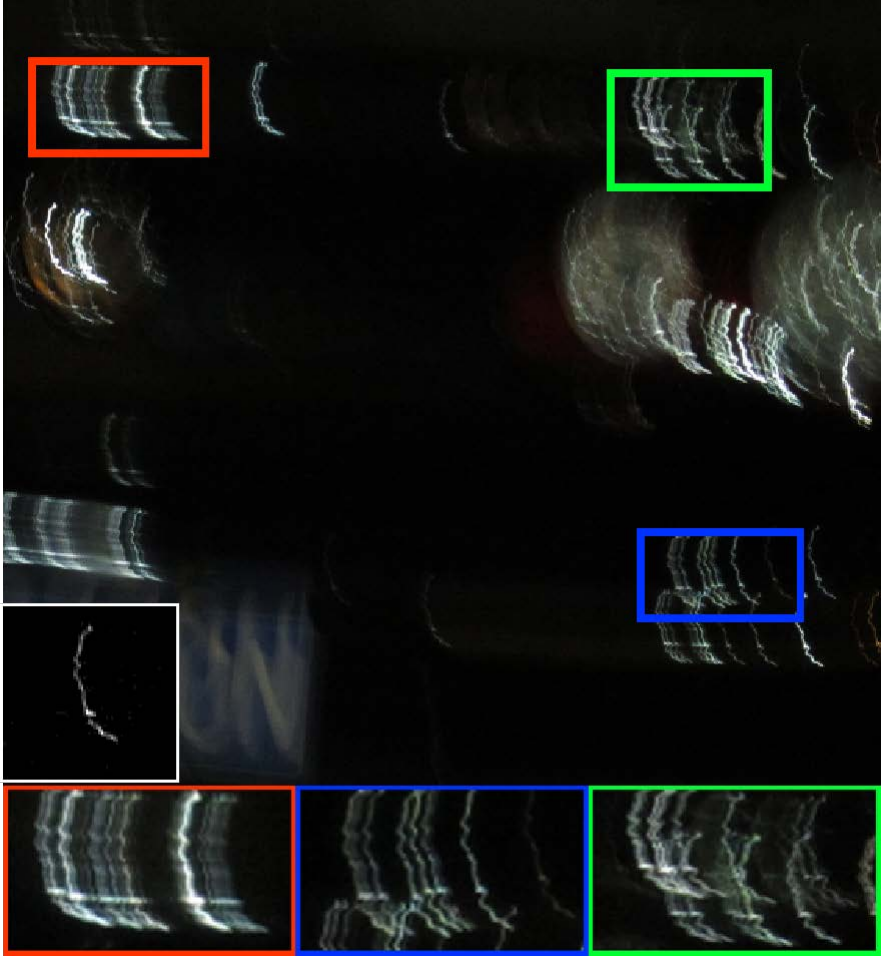}}
		\centerline{(a) Blurry image}
	\end{minipage}
	\begin{minipage}[b]{0.16\linewidth}
		\centering
		\centerline{
			\includegraphics[width =\linewidth]{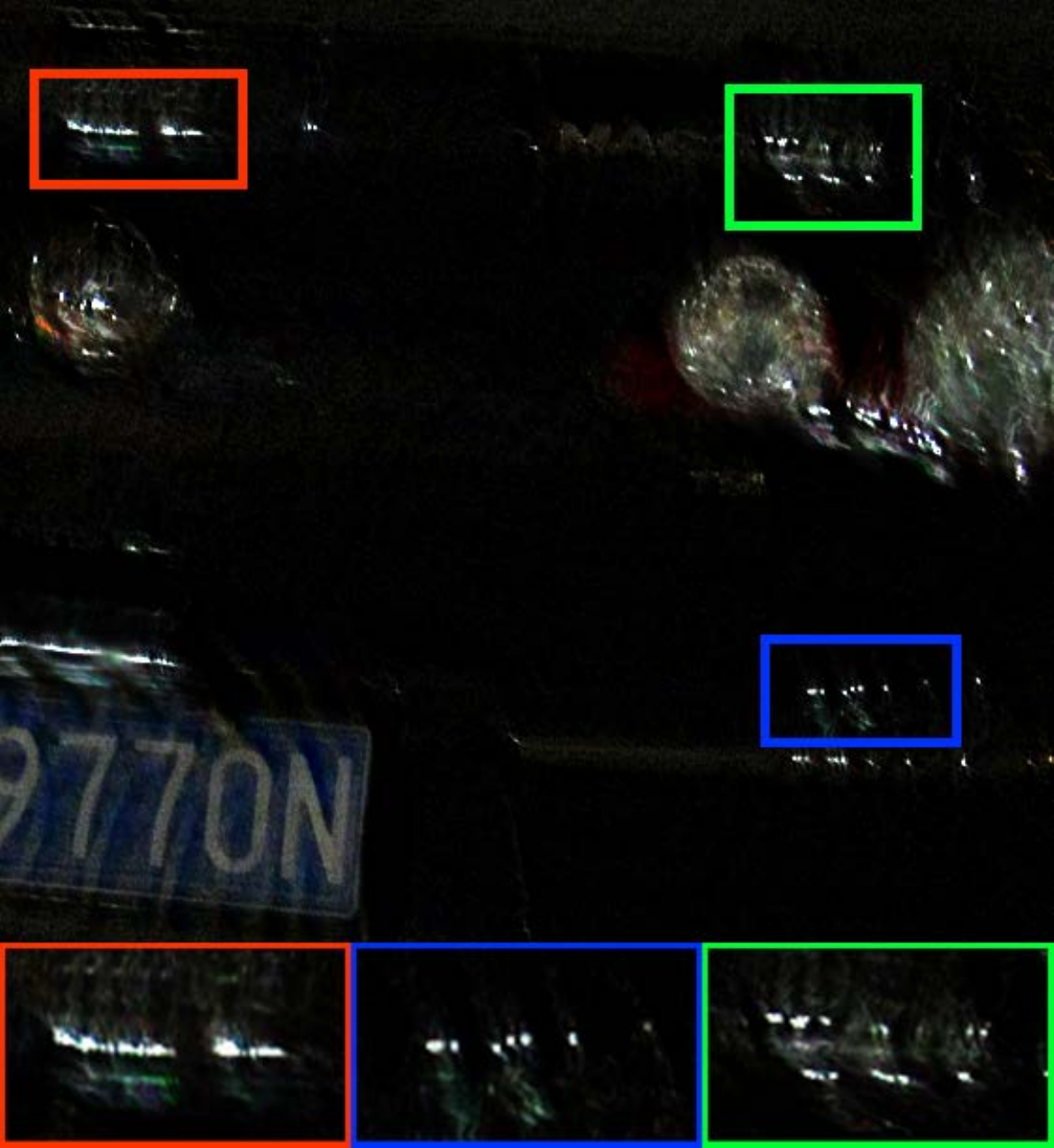}}
		\centerline{(b) Whyte \etal~\cite{Whyte14deblurring}}
	\end{minipage}
	\begin{minipage}[b]{0.16\linewidth}
		\centering
		\centerline{
			\includegraphics[width =\linewidth]{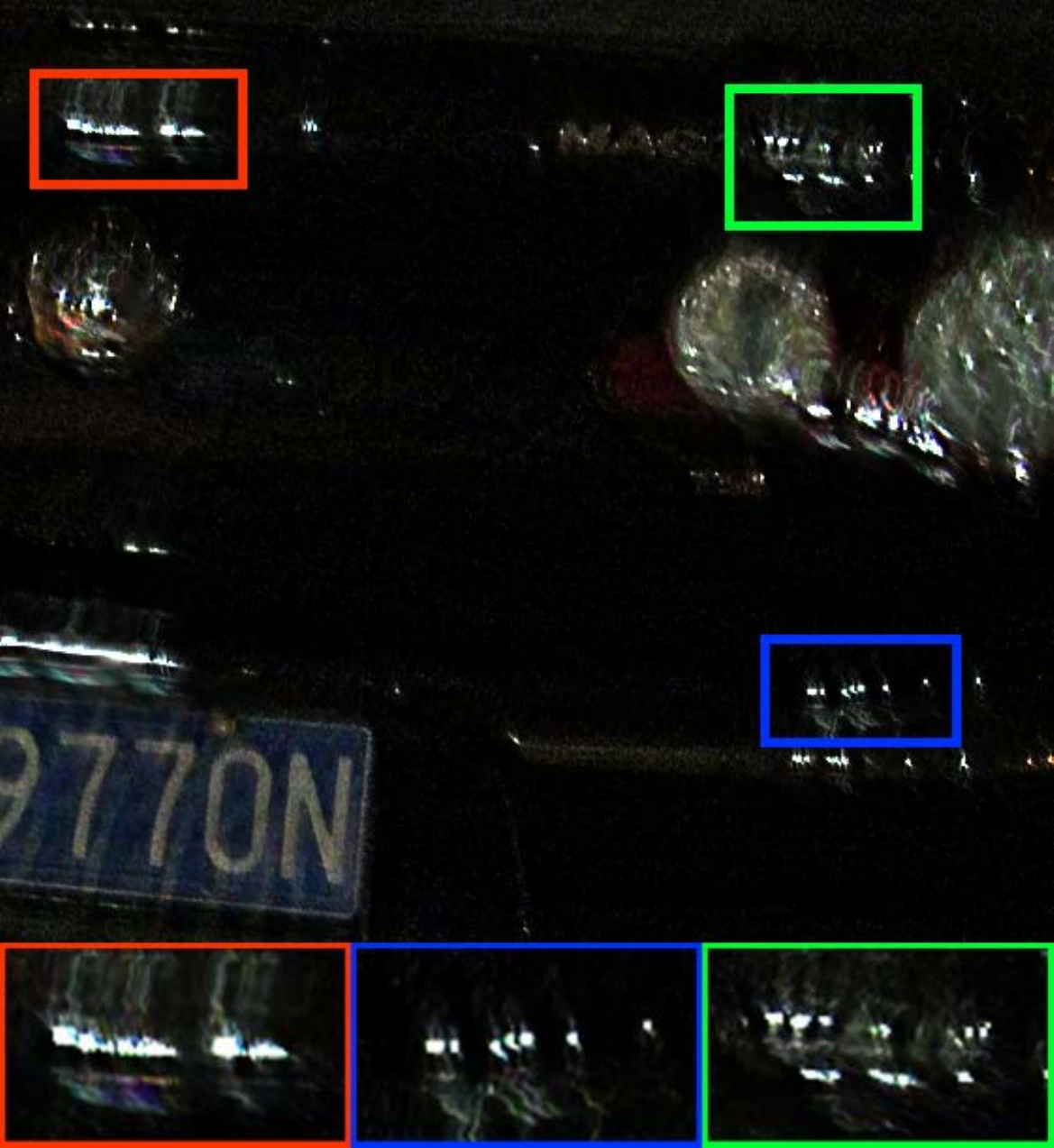}}
		\centerline{(c) Hu \etal~\cite{hu2014deblurring}}
	\end{minipage}
	\begin{minipage}[b]{0.16\linewidth}
		\centering
		\centerline{
			\includegraphics[width =\linewidth]{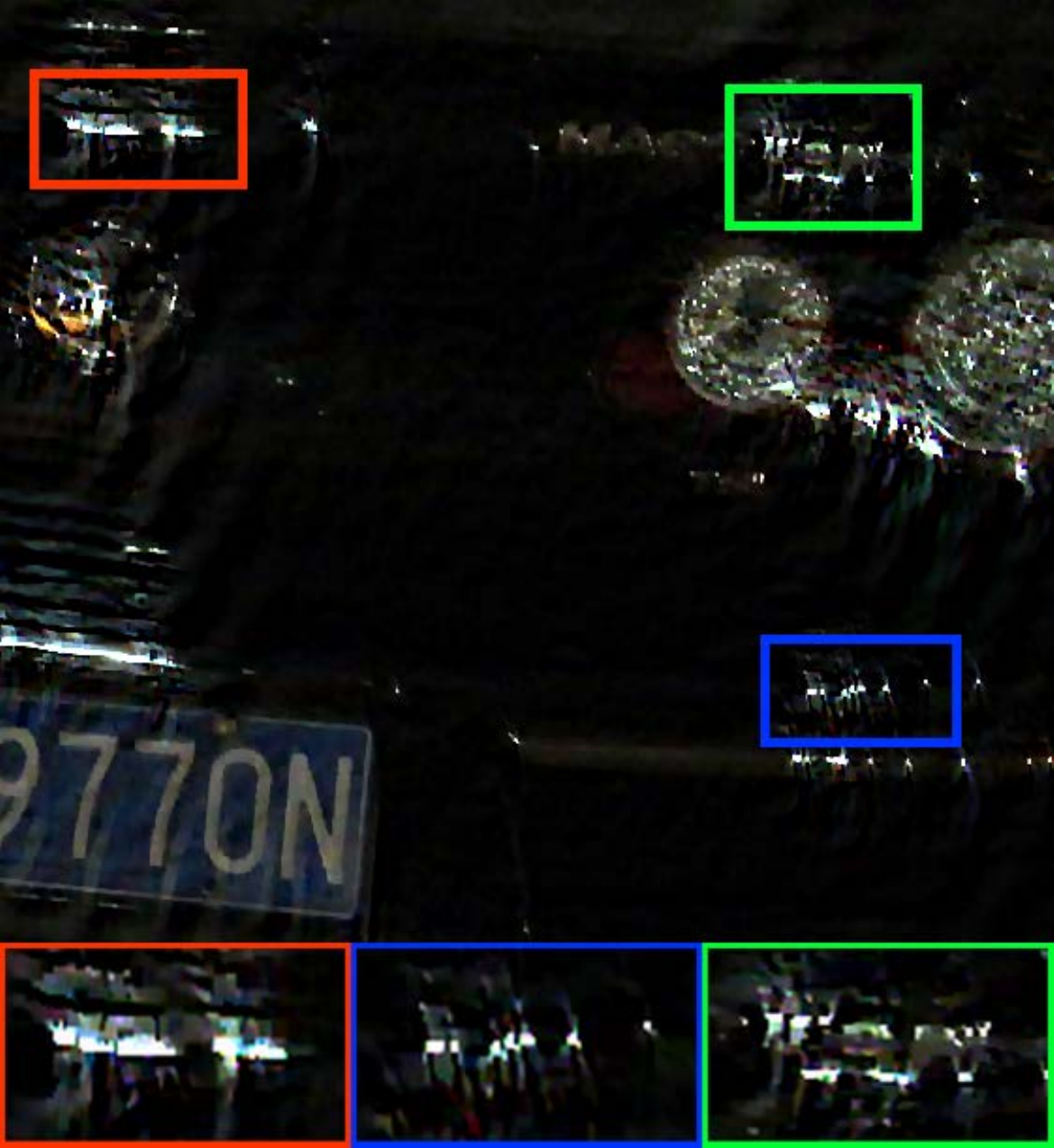}}
		\centerline{(d) Chen \etal~\cite{chen2020oid}}
	\end{minipage}
	\begin{minipage}[b]{0.16\linewidth}
		\centering
		\centerline{
			\includegraphics[width =\linewidth]{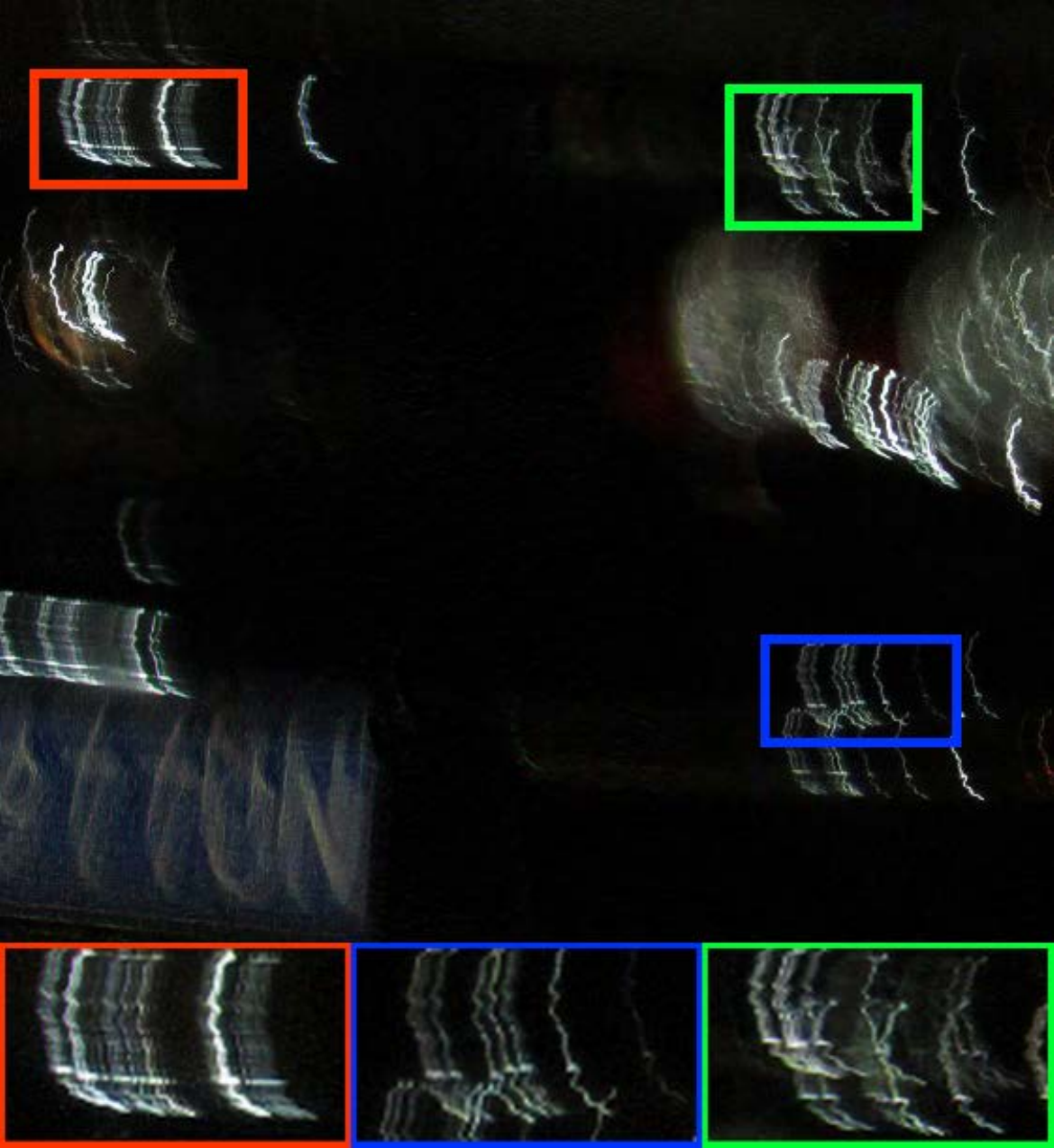}}
		\centerline{(e) SRN~\cite{tao2018scale}}
	\end{minipage}
	\begin{minipage}[b]{0.16\linewidth}
		\centering
		\centerline{
			\includegraphics[width =\linewidth]{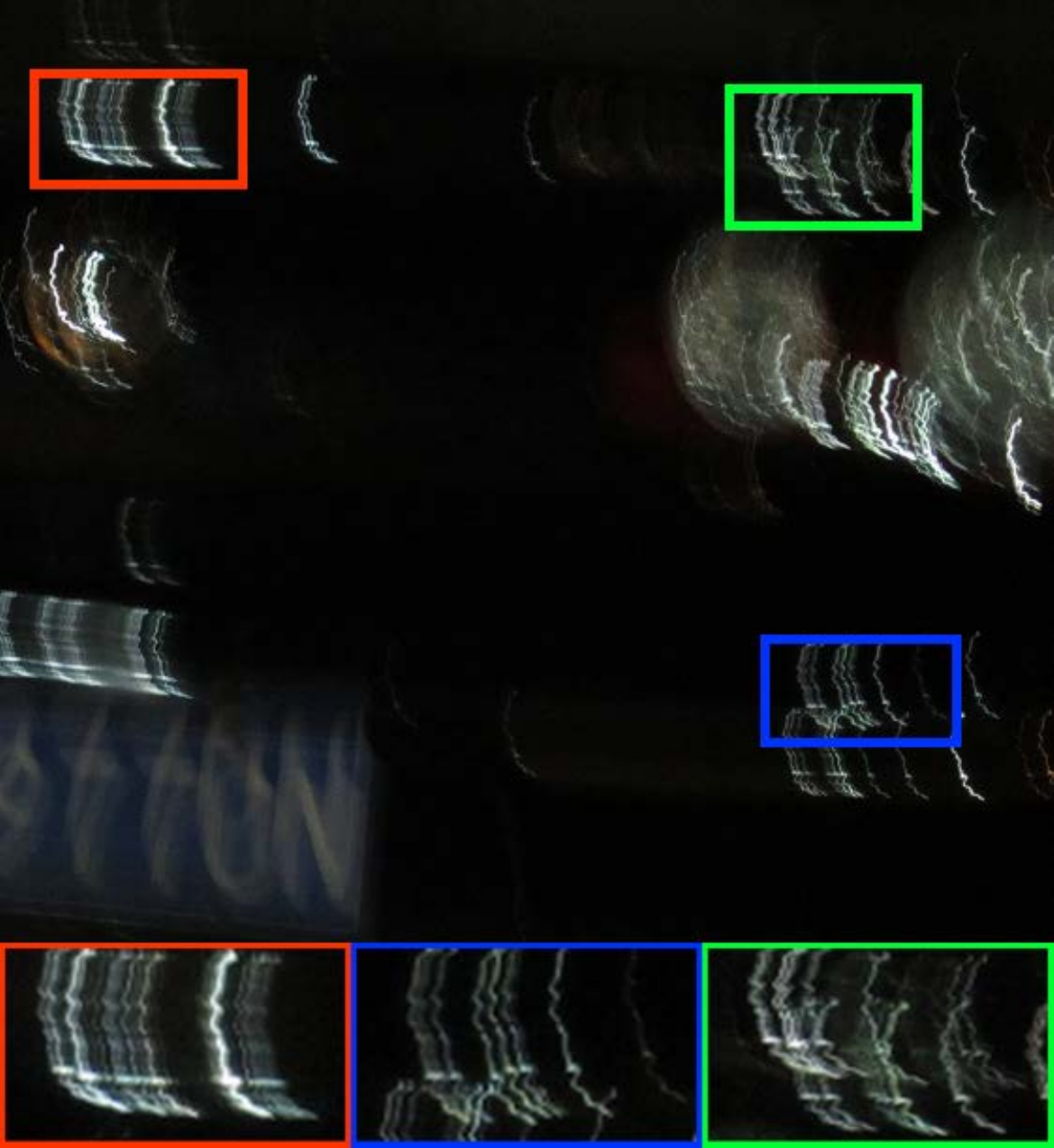}}
		\centerline{(e) Uformer~\cite{wang2022uformer}}
	\end{minipage}\\
	\begin{minipage}[b]{0.16\linewidth}
		\centering
		\centerline{
			\includegraphics[width =\linewidth]{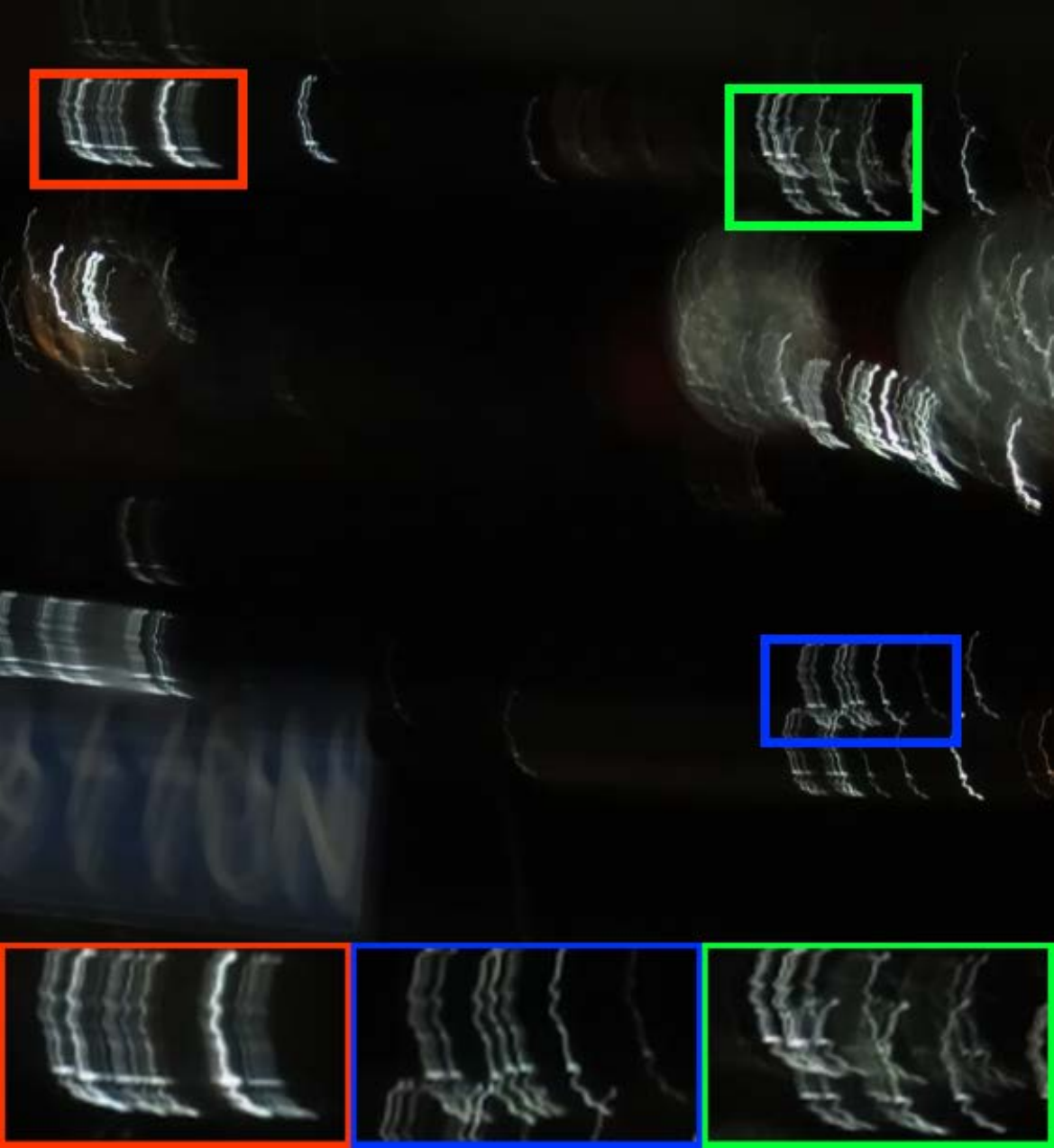}}
		\centerline{(e) Strpformer~\cite{tsai2022stripformer}}
	\end{minipage}
	\begin{minipage}[b]{0.16\linewidth}
		\centering
		\centerline{
			\includegraphics[width =\linewidth]{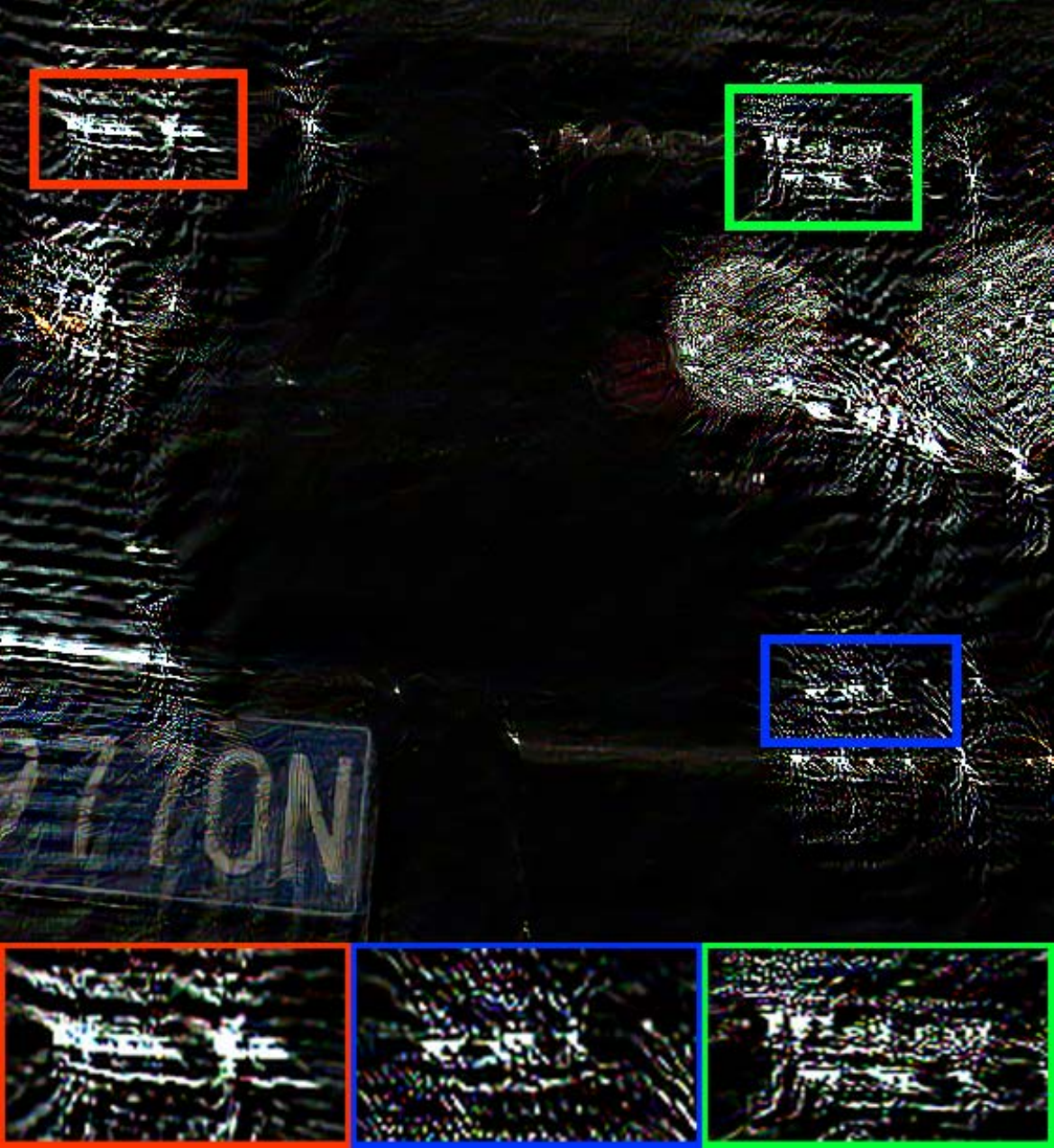}}
		\centerline{(f) IRCNN~\cite{kaiZhang_2017_CVPR}}
	\end{minipage}
	\begin{minipage}[b]{0.16\linewidth}
		\centering
		\centerline{
			\includegraphics[width =\linewidth]{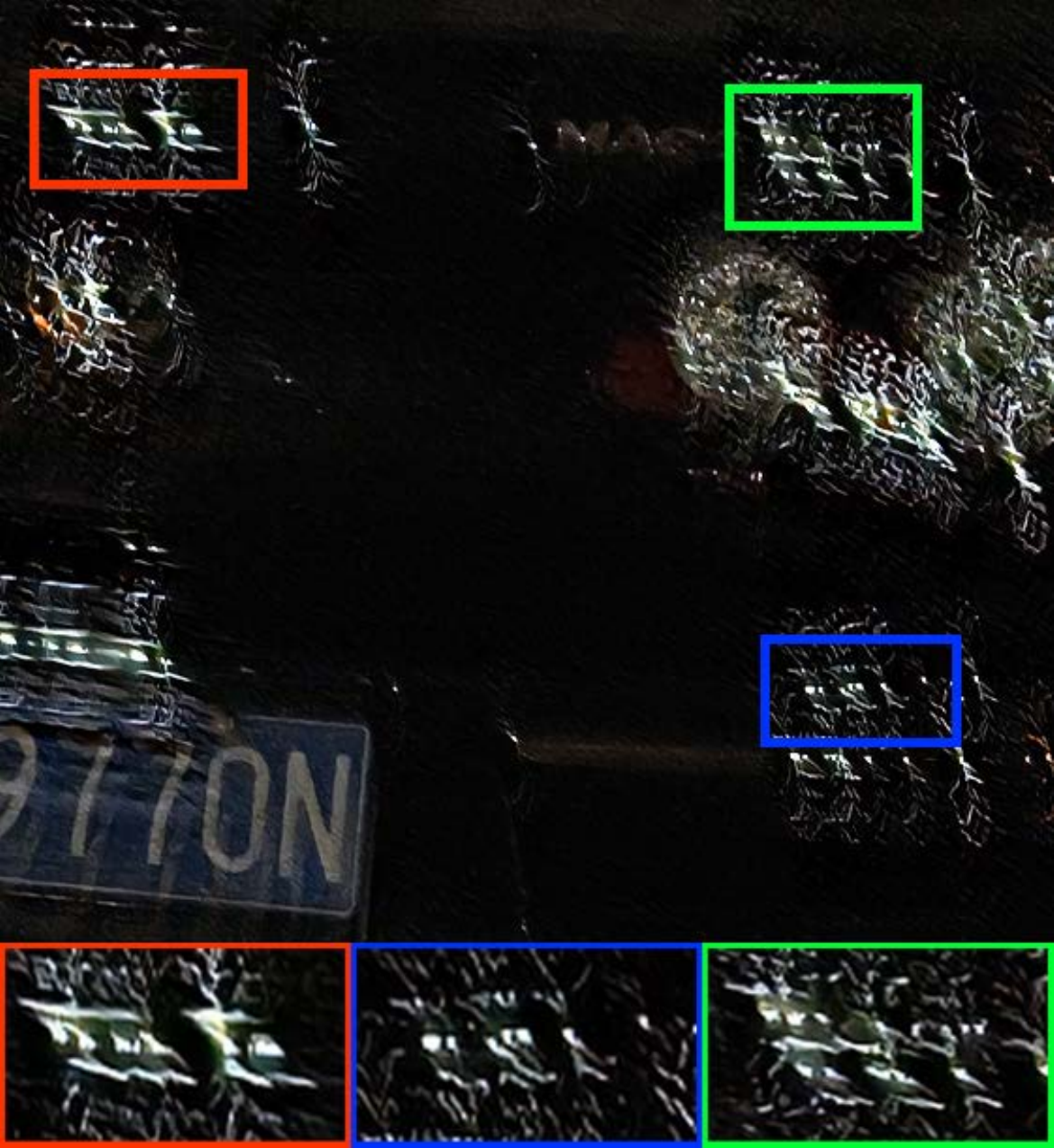}}
		\centerline{(g) RGDN~\cite{gong2018learning}}
	\end{minipage}
	\begin{minipage}[b]{0.16\linewidth}
		\centering
		\centerline{
			\includegraphics[width =\linewidth]{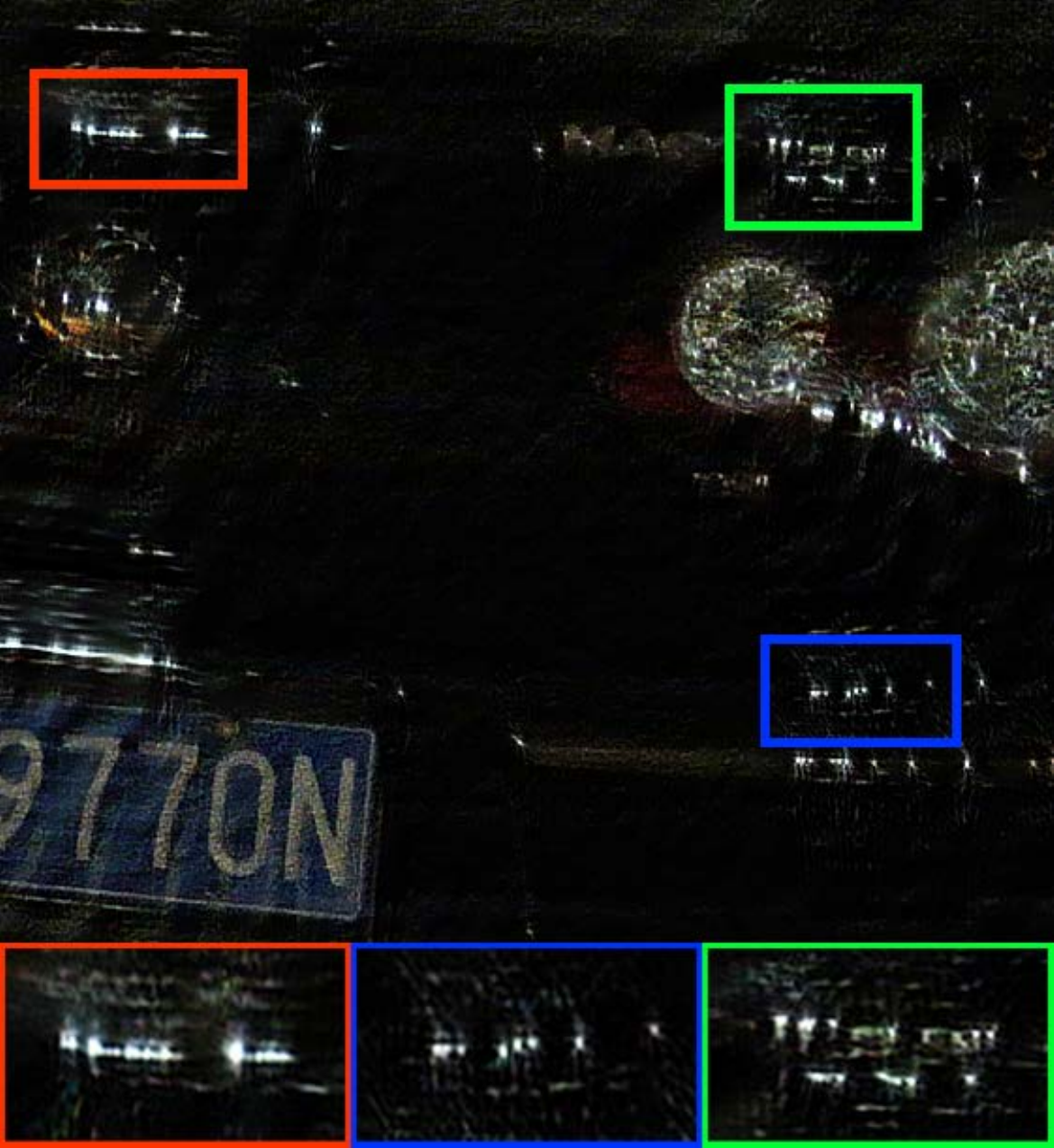}}
		\centerline{(h) DWDN~\cite{dong2020deep}}
	\end{minipage}
	\begin{minipage}[b]{0.16\linewidth}
		\centering
		\centerline{
			\includegraphics[width =\linewidth]{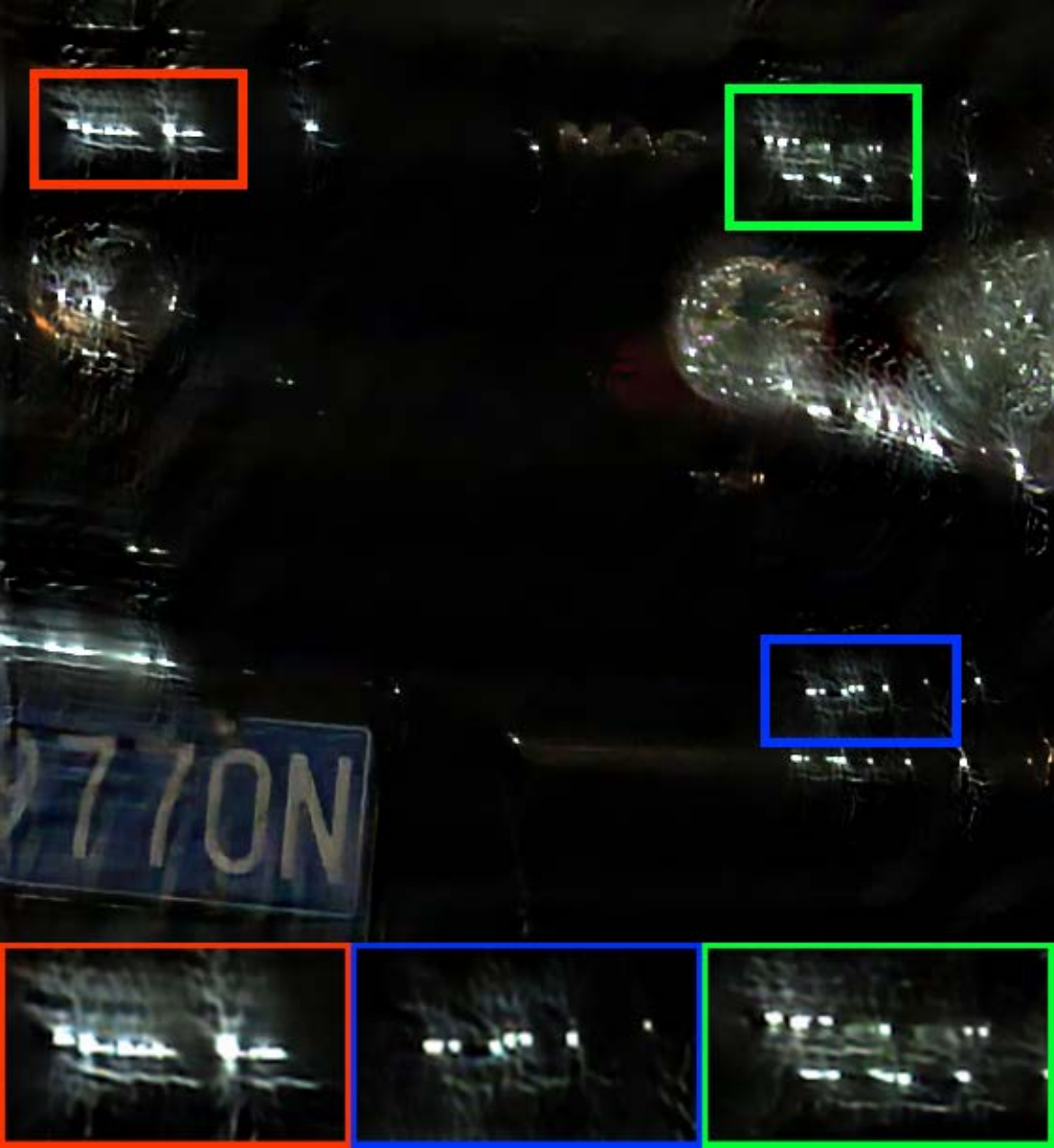}}
		\centerline{(i) NBDN~\cite{chen2021learning}}
	\end{minipage}
	\begin{minipage}[b]{0.16\linewidth}
		\centering
		\centerline{
			\includegraphics[width =\linewidth]{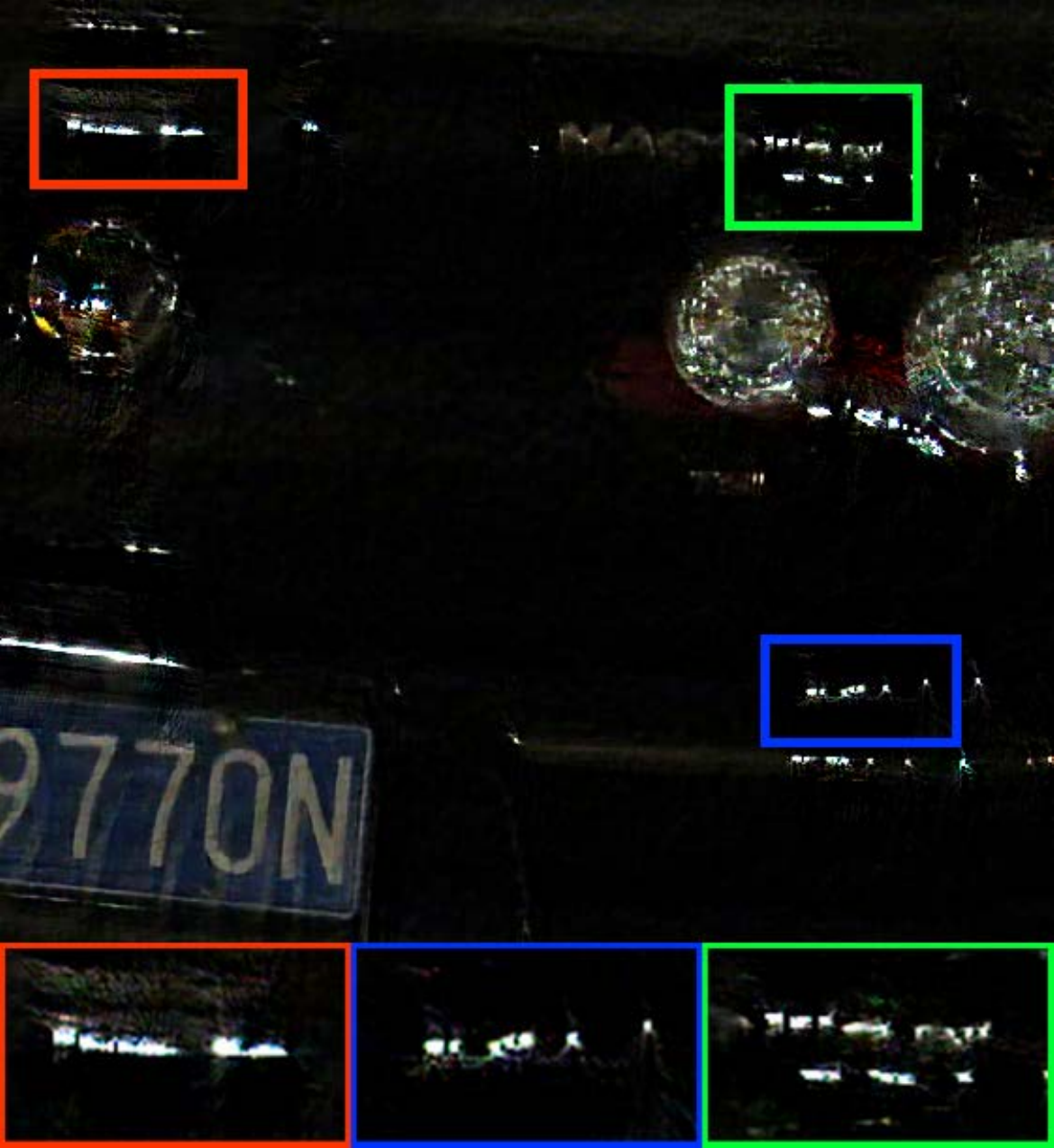}}
		\centerline{(j) Ours}
	\end{minipage}
\caption{Deblurring results of a real-world example with numerous saturated pixels. The kernel in the white box is from~\cite{dong2017blind}. Our method performs favorably compared with existing non-blind deblurring methods, which generates a result with fewer color artifacts in the boxes. Please zoom-in for a better view.}
	\label{fig real1}
\end{figure*}

\begin{figure*}
\scriptsize
\centering
	\begin{minipage}[b]{0.107\linewidth}
		\centering
		\centerline{
			\includegraphics[width =\linewidth]{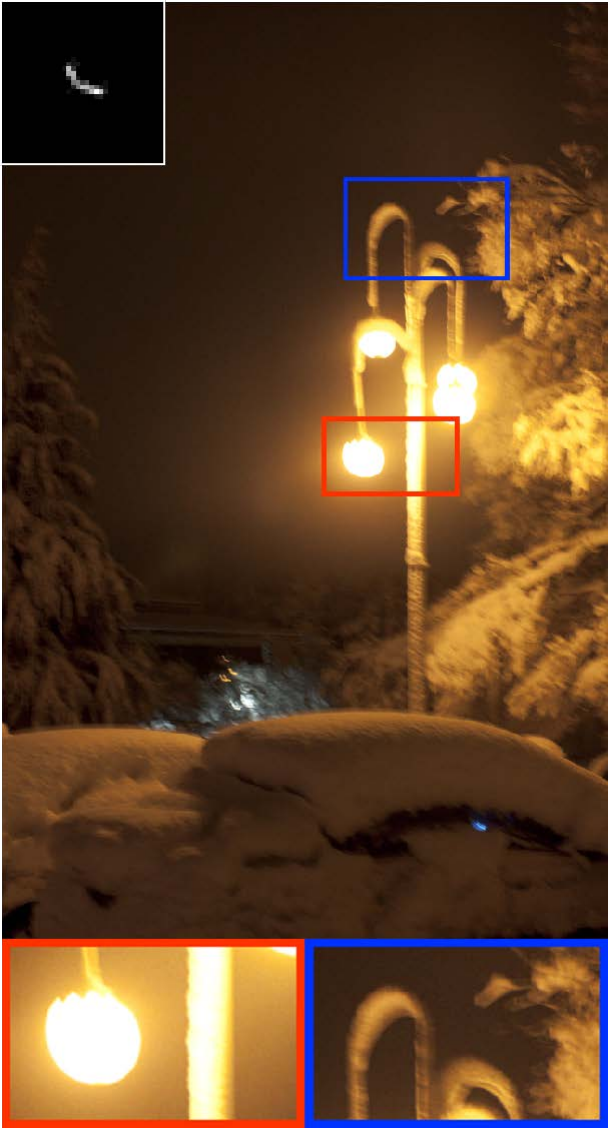}}
		\centerline{(a) Blurry image}
	\end{minipage}
	\begin{minipage}[b]{0.107\linewidth}
		\centering
		\centerline{
			\includegraphics[width =\linewidth]{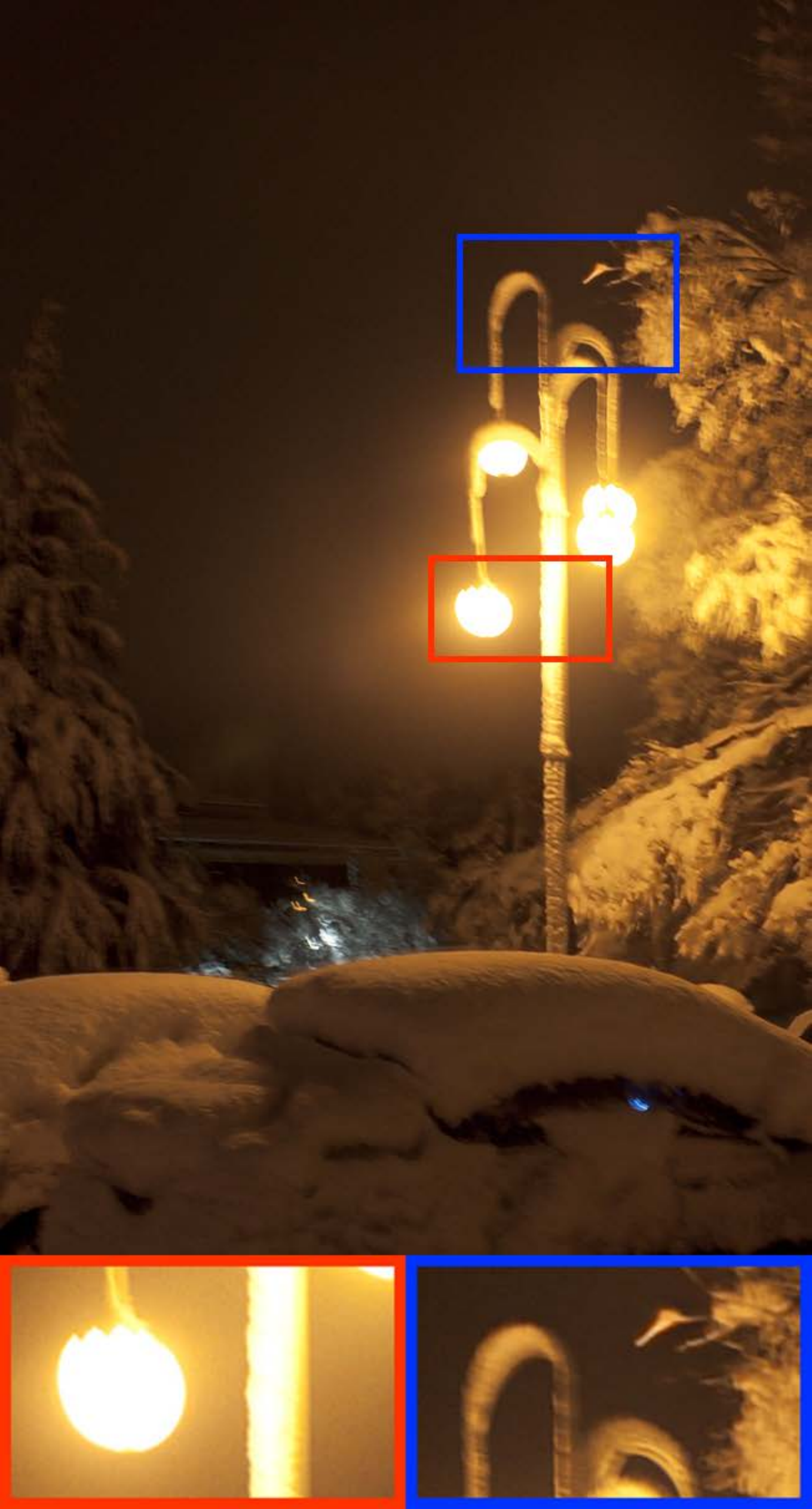}}
		\centerline{(b) Uformer~\cite{wang2022uformer}}
	\end{minipage}
	\begin{minipage}[b]{0.107\linewidth}
		\centering
		\centerline{
			\includegraphics[width =\linewidth]{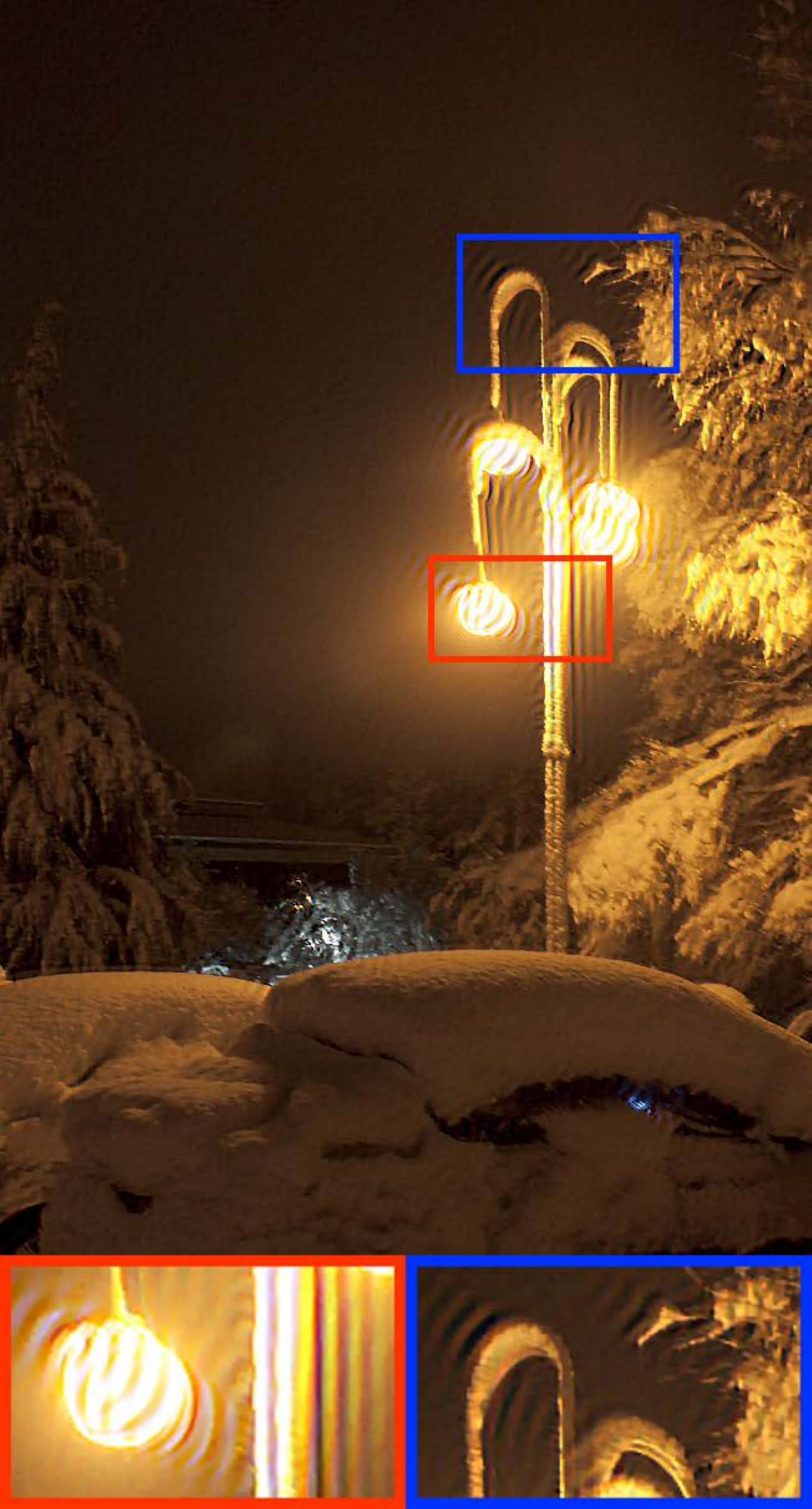}}
		\centerline{(c) Cho \etal~\cite{cho2011outlier}}
	\end{minipage}
	\begin{minipage}[b]{0.107\linewidth}
		\centering
		\centerline{
			\includegraphics[width =\linewidth]{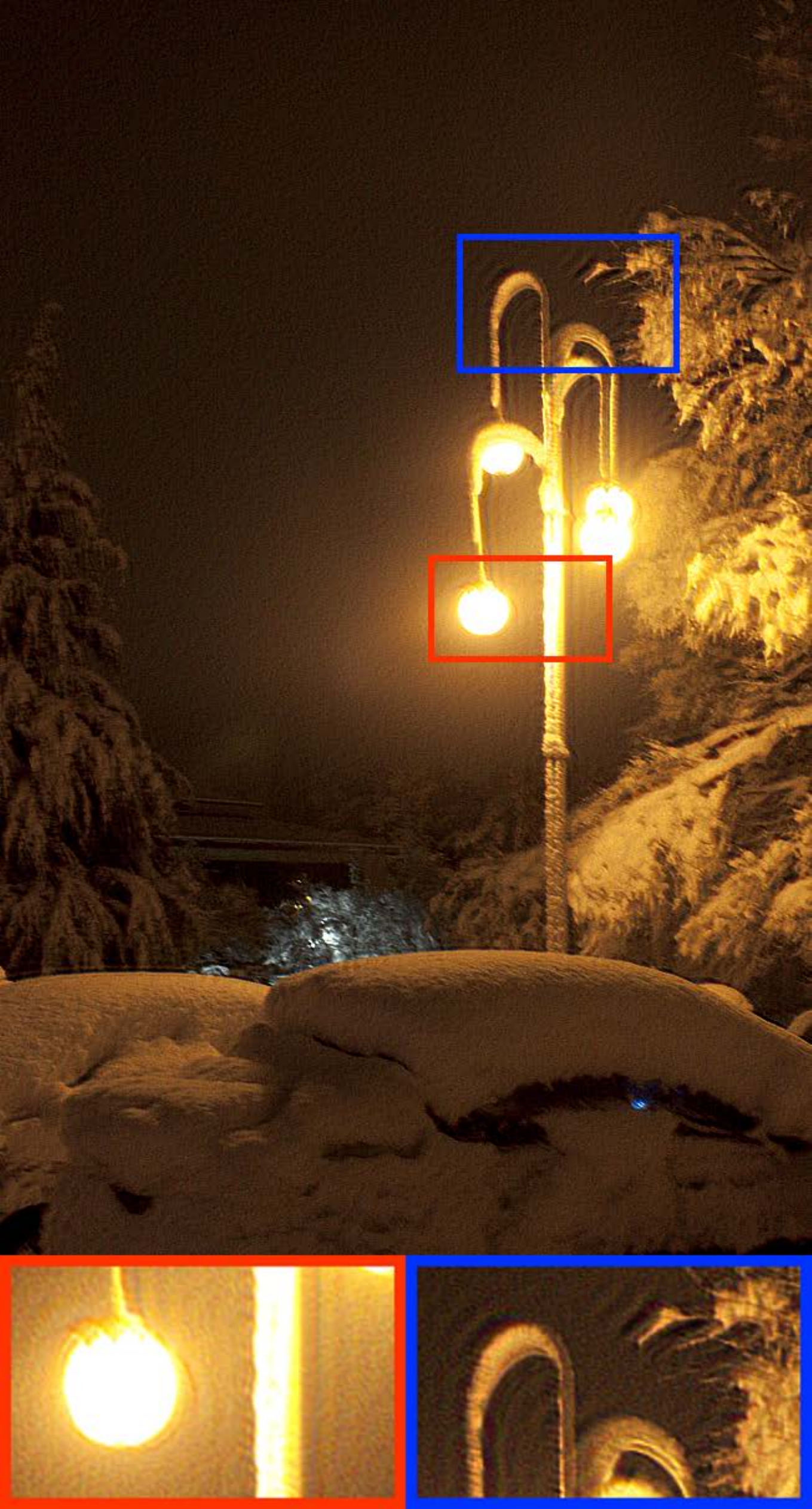}}
		\centerline{(d) Whyte \etal~\cite{Whyte14deblurring}}
	\end{minipage}
	\begin{minipage}[b]{0.107\linewidth}
		\centering
		\centerline{
			\includegraphics[width =\linewidth]{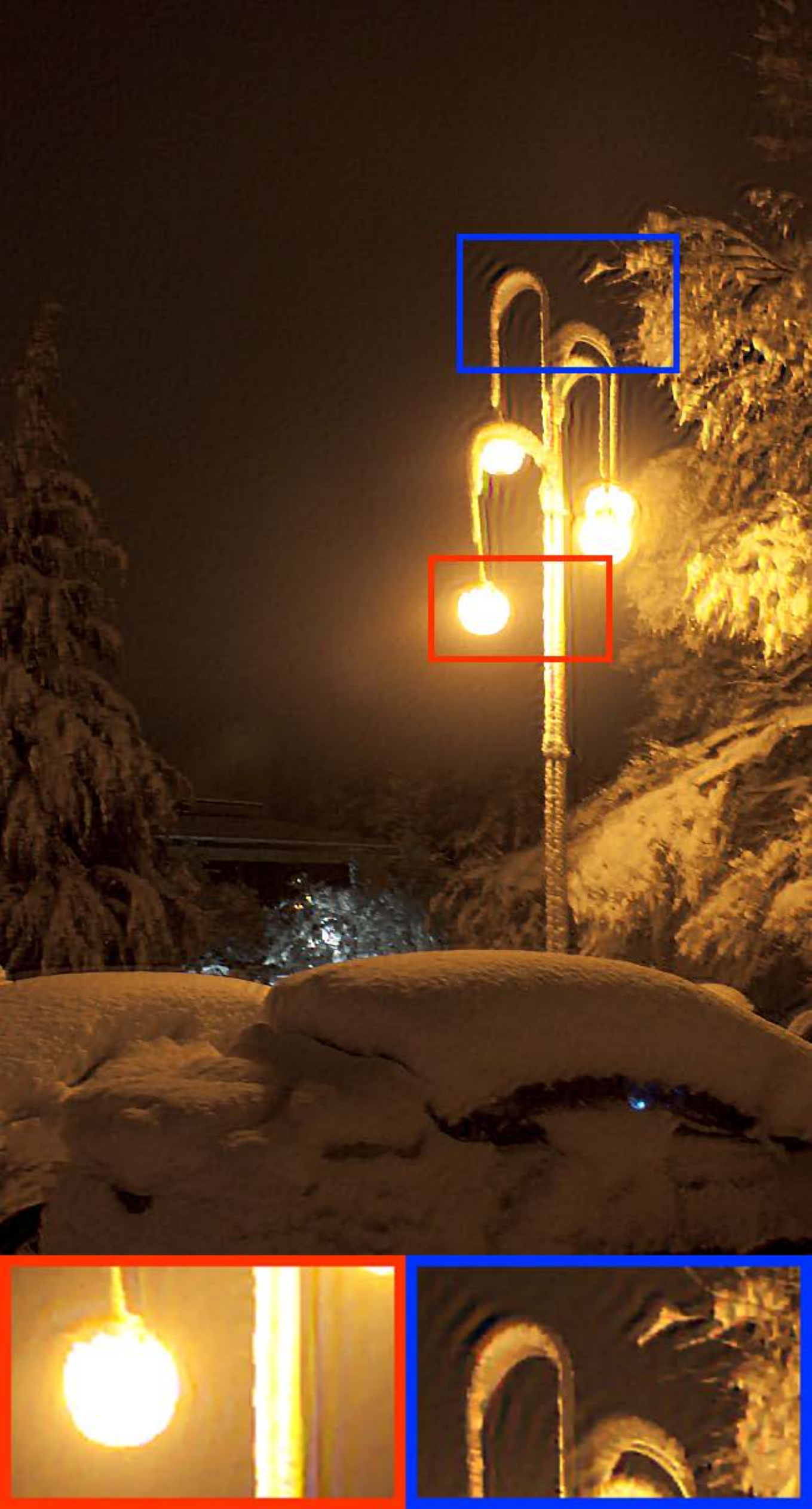}}
		\centerline{(e) Pan \etal~\cite{pan2016robust}}
	\end{minipage}
	\begin{minipage}[b]{0.107\linewidth}
		\centering
		\centerline{
			\includegraphics[width =\linewidth]{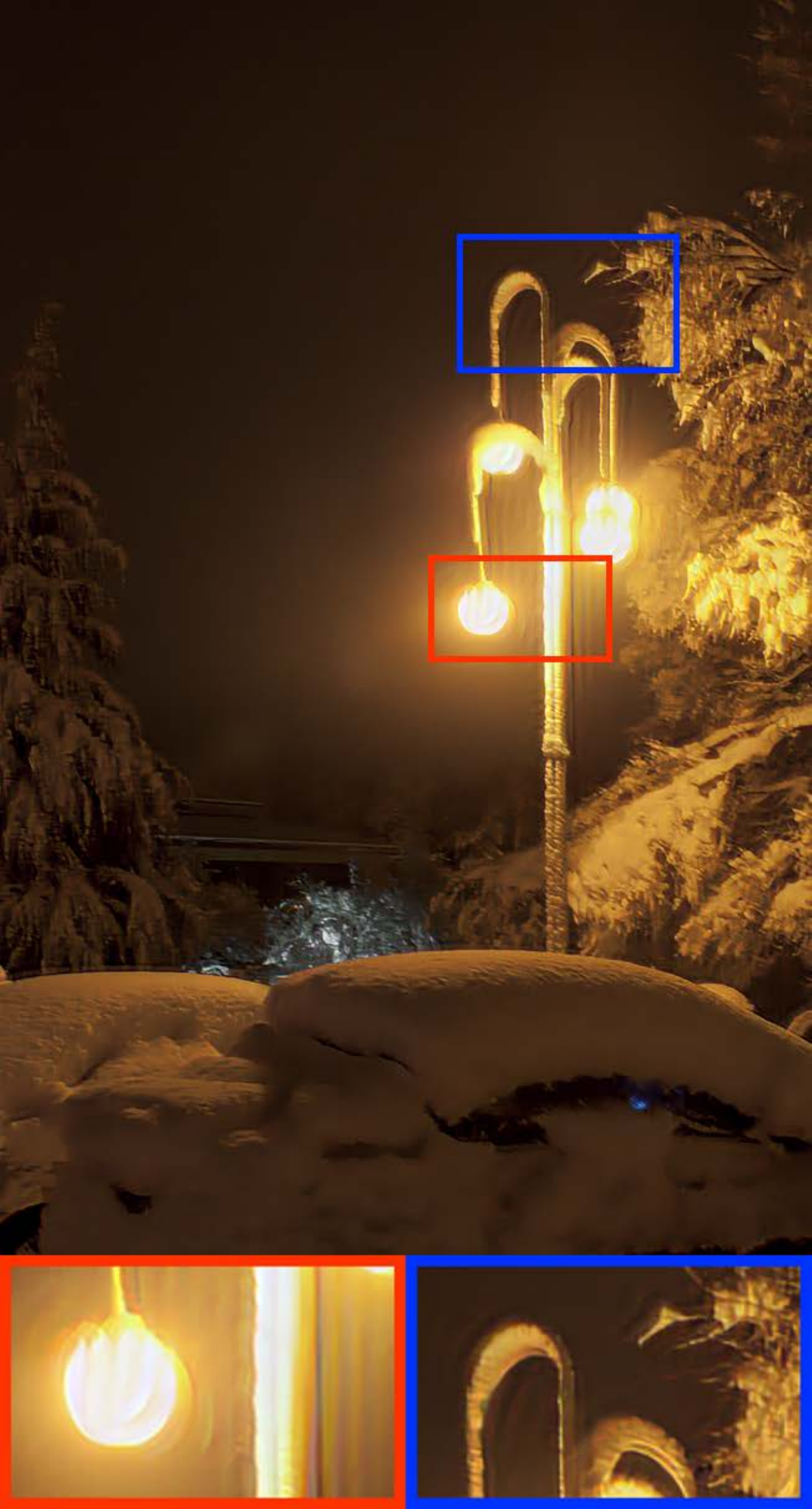}}
		\centerline{(f) FCNN~\cite{Zhang_2017_CVPR}}
	\end{minipage}
	\begin{minipage}[b]{0.107\linewidth}
		\centering
		\centerline{
			\includegraphics[width =\linewidth]{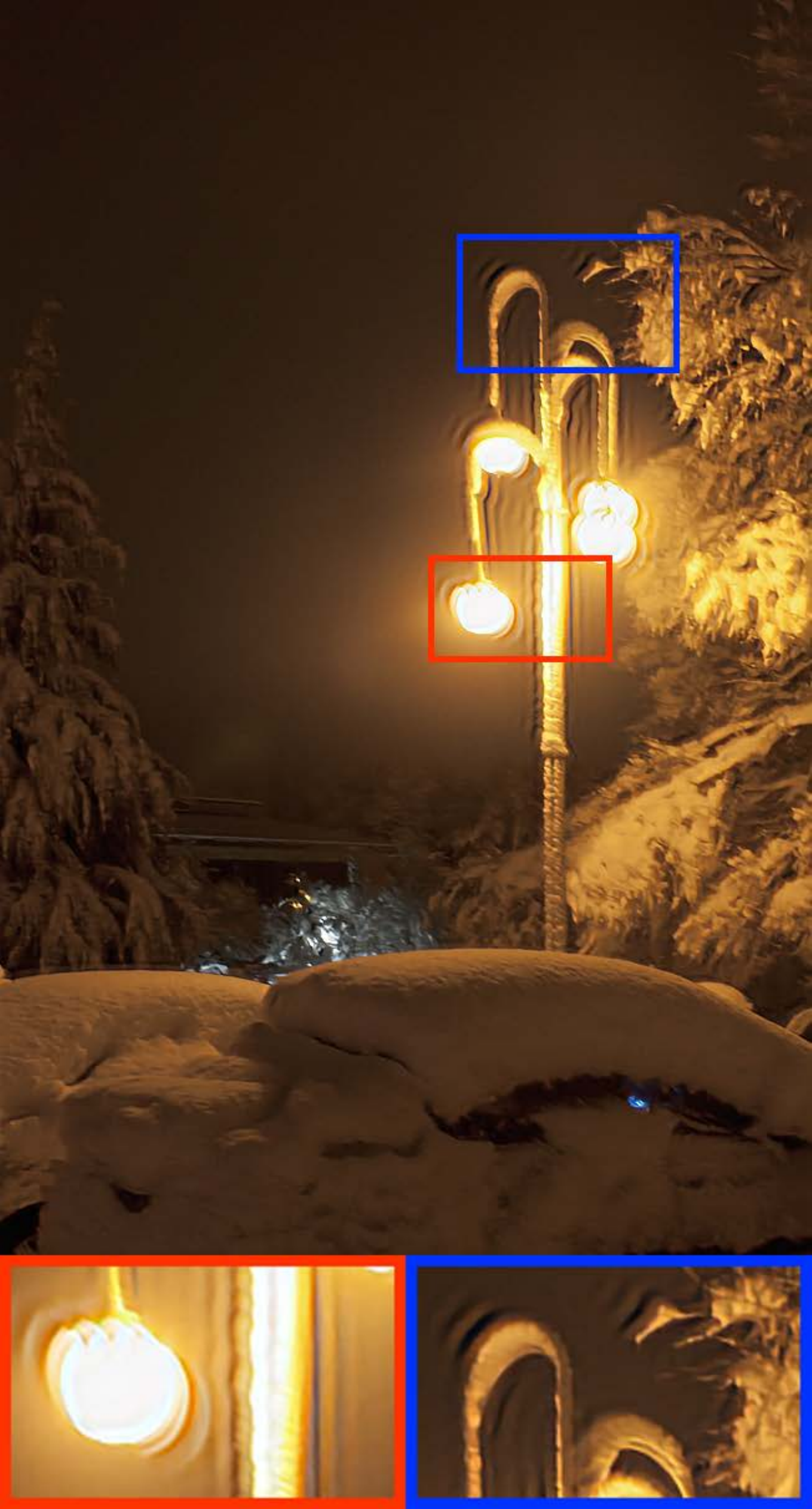}}
		\centerline{(g) RGDN~\cite{gong2018learning}}
	\end{minipage}
	\begin{minipage}[b]{0.107\linewidth}
		\centering
		\centerline{
			\includegraphics[width =\linewidth]{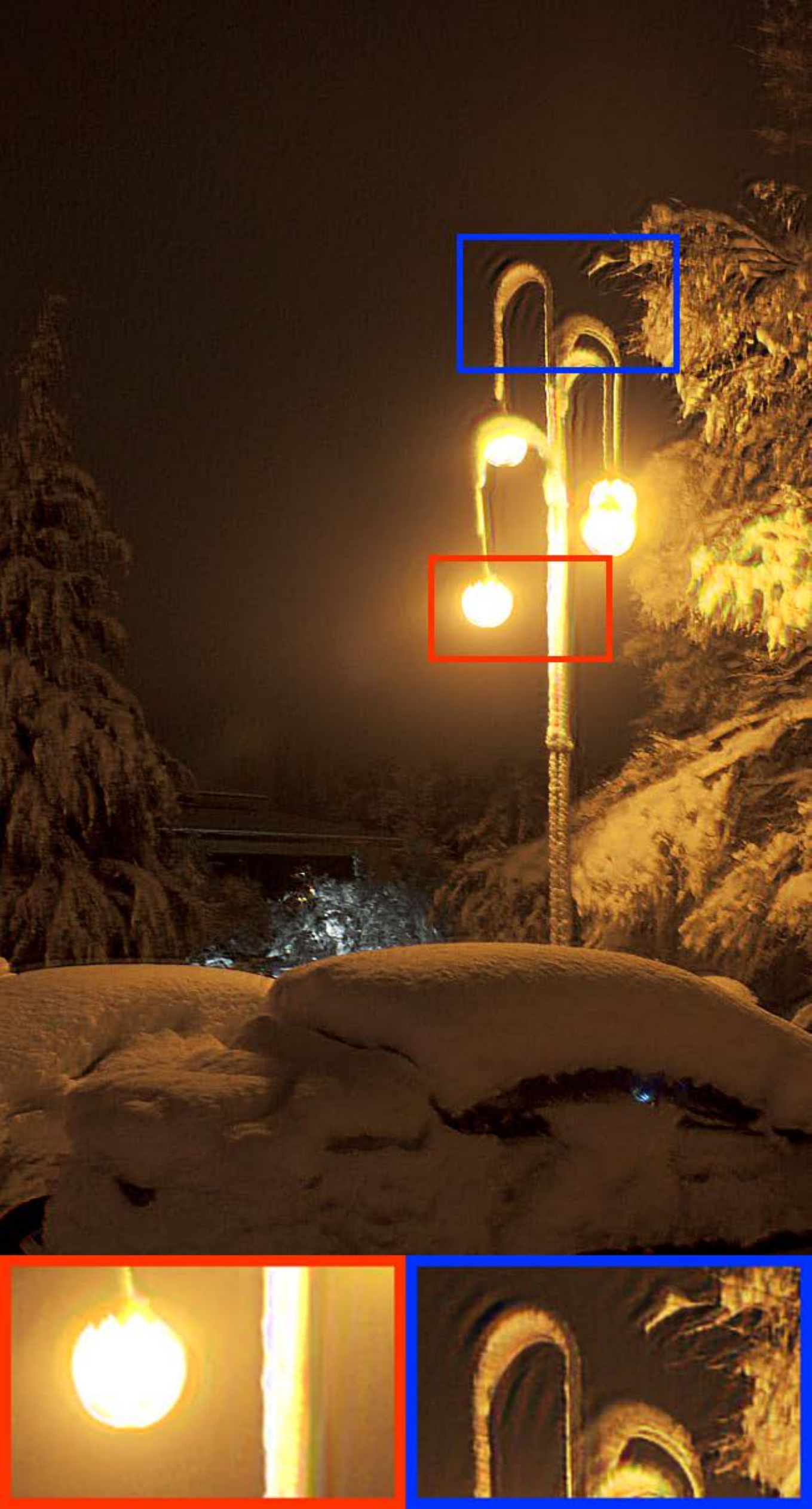}}
		\centerline{(h) NBDN~\cite{chen2021learning}}
	\end{minipage}
	\begin{minipage}[b]{0.107\linewidth}
		\centering
		\centerline{
			\includegraphics[width =\linewidth]{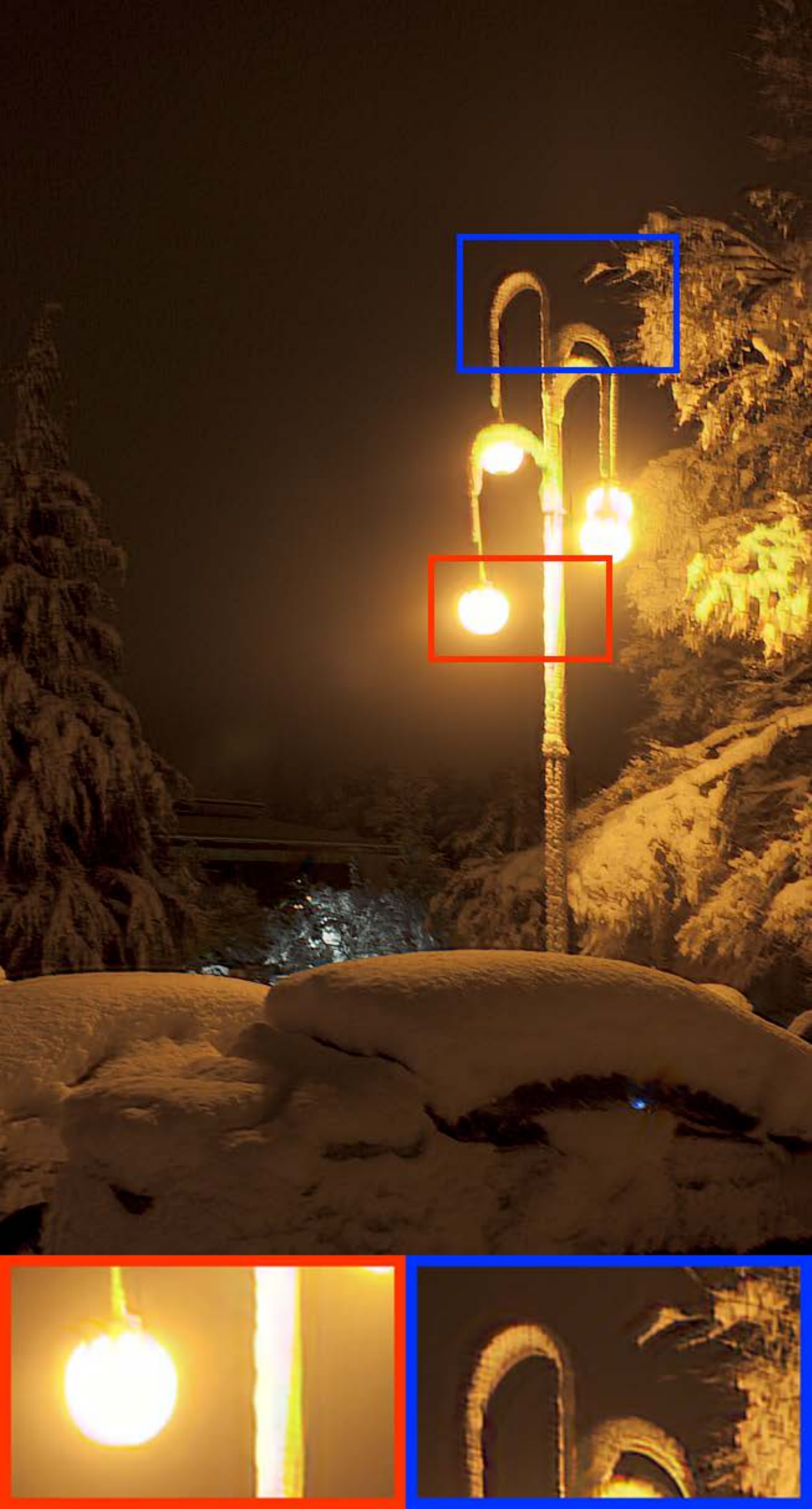}}
		\centerline{(i) Ours}
	\end{minipage}
\caption{Deblurring results of a real-world example with numerous saturated pixels. The kernel in the white box is from~\cite{cho2011outlier}. Our method performs favorably compared with existing non-blind deblurring methods, which generates a result with fewer color artifacts in the boxes. Please zoom-in for a better view.}
	\label{fig real2}
\end{figure*}

\noindent\textbf{Unsaturated images.}
Although our method is specially designed for saturated image deblurring, we show it can also be used to deal with unsaturated data.
Blurry images from the benchmark dataset~\cite{levin2009understanding} are used for evaluation.
PSNR and SSIM values are shown in Table~\ref{tab moredata}. We can observe that although our method is designed for saturated images, it performs competitively against state-of-the-art methods on images without saturation.
Deblurring results of an unsaturated example from different methods are shown in Fig.~\ref{fig unsat}. Our method performs favorably against existing methods, which generates a result with finer details and fewer artifacts compared with others.

\noindent\textbf{Saturated images from Hu \etal~\cite{hu2014deblurring}.}
Besides the saturated data in the proposed testing set, we also use the benchmark dataset from Hu \etal~\cite{hu2014deblurring} to show the effectiveness of the proposed algorithm.
This dataset consists of a total of 154 saturated blurry images and 12 blur kernels.
Results shown in Table~\ref{tab moredata} demonstrate that with either ground truth kernels or estimated blur kernels, our method can generate favorable results against existing methods.
The results show that our method can generalize well to saturated images from another dataset.

\noindent\textbf{Real-world saturated images.}
We also use some challenging real-world examples to evaluate our method.
The comparisons are presented in Fig.~\ref{fig real1} and \ref{fig real2}.
We observe that the optimization-based methods~\cite{cho2011outlier,pan2016robust,hu2014deblurring,Whyte14deblurring,chen2020oid} have difficulties in simultaneously restoring details and removing artifacts due to the ineffectiveness or the absence of the adopted image prior.
In comparison, the learning-based methods~\cite{Zhang_2017_CVPR,kaiZhang_2017_CVPR,gong2018learning,dong2020deep} can generate results with more details.
However, saturated pixels are not specially considered in their models.
As a result, the recovered results contain severe artifacts in the saturated regions.
The deep learning-based deblurring methods~\cite{tao2018scale,wang2022uformer,tsai2022stripformer} can hardly obtain satisfying results as the imaging process is ignored.
Moreover, although the method from the robust learning-based method~\cite{chen2021learning} can generate a result with fewer artifacts, we observe there are still ringings around the saturated region. This is mainly because their adopted model is ineffective in handling all saturated pixels.
Different from those methods, the comparison results demonstrate that our method can prevent side-effects from the saturated pixels while obtaining clearer details at the same time.

\begin{table}[]
\centering
\caption{Model size and running time comparisons.}
\scalebox{1}{
\begin{tabular}{c|c|c}
\bottomrule
Methods & Total parameters (M) & Running time (s)\\
\midrule
Cho \etal~\cite{cho2011outlier} &- &5.25 (CPU) \\
Whyte \etal~\cite{Whyte14deblurring} &- &2.86 (CPU)\\
Hu \etal~\cite{hu2014deblurring} &- &5.61 (CPU) \\
Pan \etal~\cite{pan2016robust} &- &15.41 (CPU)\\
Chen \etal~\cite{chen2021blind} &- &5.45 (CPU) \\
FCNN~\cite{Zhang_2017_CVPR} & 0.45 &0.13\\
IRCNN~\cite{kaiZhang_2017_CVPR} & 0.15 &1.11\\
RGDN~\cite{gong2018learning}  & 1.26 &5.34\\
DWDN~\cite{dong2020deep} &7.05 &2.60 \\
NBDN~\cite{chen2021learning} &0.39 &0.25\\
Ours & 0.16 &0.70\\
\bottomrule
\end{tabular}}
\label{tab 2}
\end{table}

\begin{table*}[t]
\centering
\caption{Comparisons on the proposed testing set \wrt different blur models.
Ours w/o $M$ is implemented by fixing $M=\textbf{1}$;
Ours w/~\cite{Whyte14deblurring} is implemented by replacing our latent map $M$ with the approximation in~\cite{Whyte14deblurring};
Ours w/~\cite{chen2021blind} is implemented by computing $M$ using the strategy from~\cite{chen2021blind}.}
    \scalebox{0.97}{
    \begin{tabular}{cc|c|c|c|c}
    \midrule
    &\multicolumn{5}{c}{GT blur kernels}\\
   \bottomrule
   & Ours w/o $M$ & Ours w/ blur model in Whyte~\cite{Whyte14deblurring} & Ours w/ blur model in NBDN~\cite{chen2021learning} & Ours w/ blur model in Chen~\cite{chen2021blind} & Ours \\
    PSNR &23.34  & 24.07 &24.69 &24.13 & 25.66 \\
    SSIM &0.8285 & 0.8304 &0.8400 &0.8312 & 0.8595\\
    \midrule
    &\multicolumn{5}{c}{Estimated kernels from~\cite{hu2014deblurring}}\\
    \midrule
    & Ours w/o $M$ & Ours w/ blur model in Whyte~\cite{Whyte14deblurring} & Ours w/ blur model in NBDN~\cite{chen2021learning} & Ours w/ blur model in Chen~\cite{chen2021blind} & Ours \\
    \midrule
    PSNR & 23.18 &23.56 &24.23 &24.07 & 25.11 \\
    SSIM & 0.8053 &0.8167 &0.8199 &0.8115 & 0.8367 \\
    \midrule
    \end{tabular}}
    \label{tab 3}
\end{table*}

\subsection{Model Size and Running Time}
Table \ref{tab 2} summarizes total numbers of parameters from different algorithms~\cite{cho2011outlier,pan2016robust,hu2014deblurring,Whyte14deblurring,Zhang_2017_CVPR,kaiZhang_2017_CVPR,gong2018learning,dong2020deep,chen2021learning} and their corresponding running time on a 300 $\times$ 300 blurry image.
All methods are evaluated on the same PC with an Intel (R) Xeon (R) CPU and an Nvidia Tesla 1080 GPU.
The proposed method does not require many parameters, and it performs favorably against state-of-the-art models in the term of running time.

\section{Analysis and Discussion}
In this section, we first analyze the effectiveness of the proposed method.
Then, we analyze the accuracies of the outputs of the proposed MEN and PEN.
We further show the robustness of our method against noise and analyze the convergence property of our method.
Finally, we discuss the merit of our training sample synthesizing step against that from~\cite{chen2021learning}.

\begin{figure*}
\small
\centering
	\begin{minipage}[b]{0.195\linewidth}
		\centering
		\centerline{
			\includegraphics[width =\linewidth]{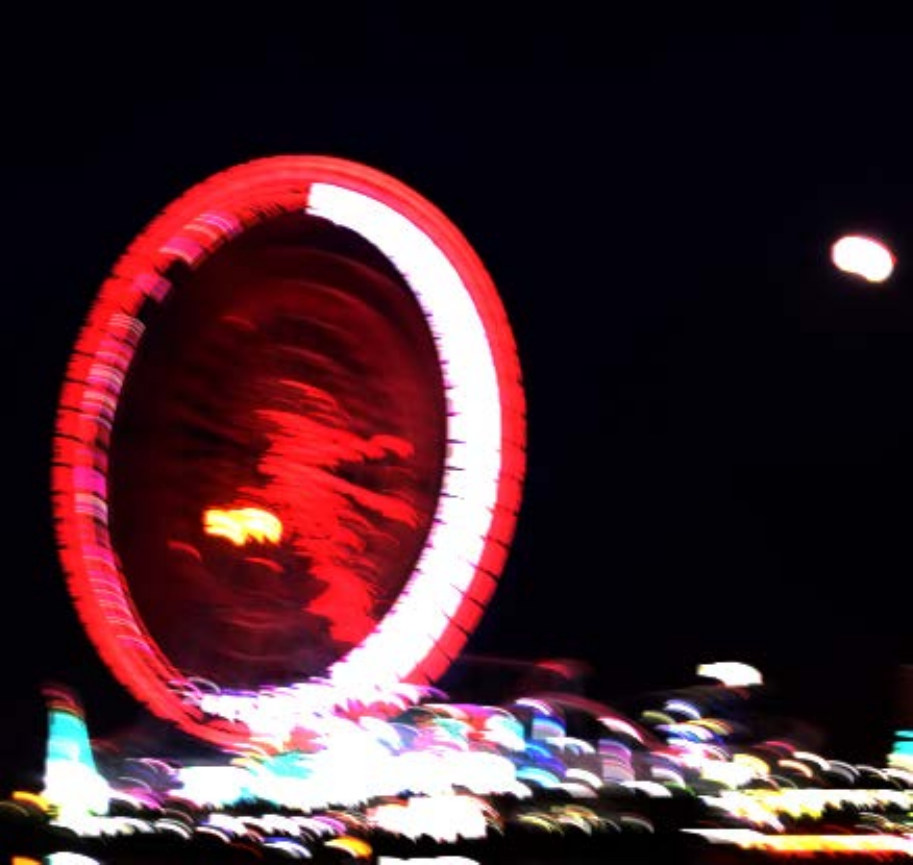}}
		\centerline{(a) Input}
	\end{minipage}
	\begin{minipage}[b]{0.195\linewidth}
		\centering
		\centerline{
			\includegraphics[width =\linewidth]{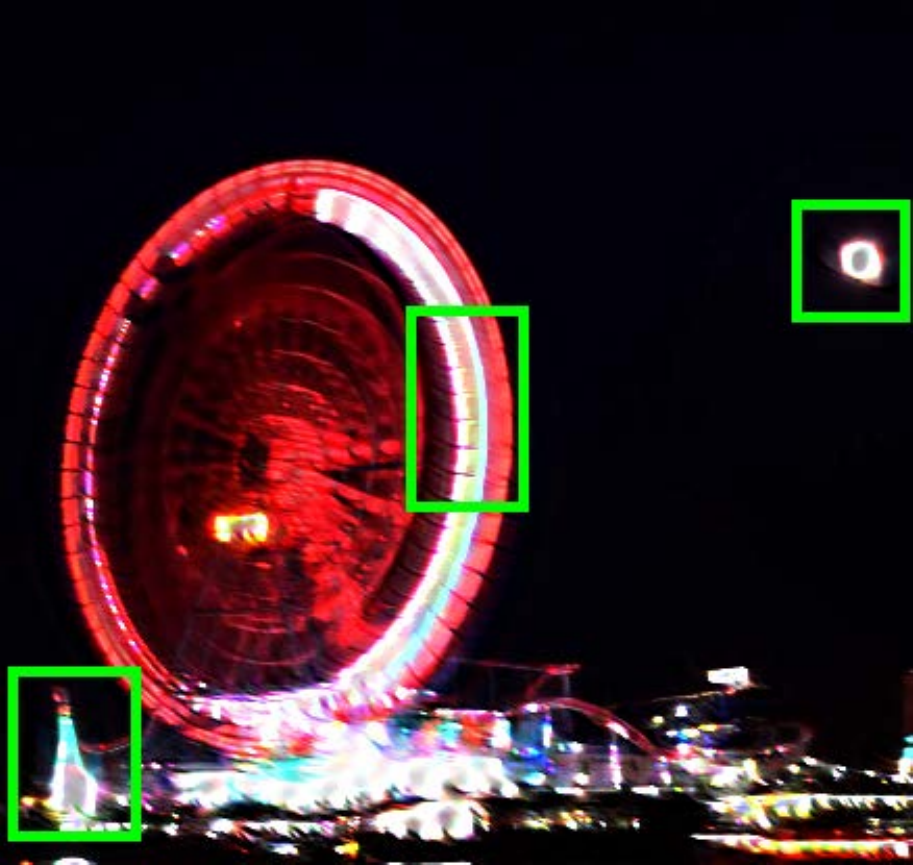}}
		\centerline{(b) Ours w/o $M$}
	\end{minipage}
	\begin{minipage}[b]{0.195\linewidth}
		\centering
		\centerline{
			\includegraphics[width =\linewidth]{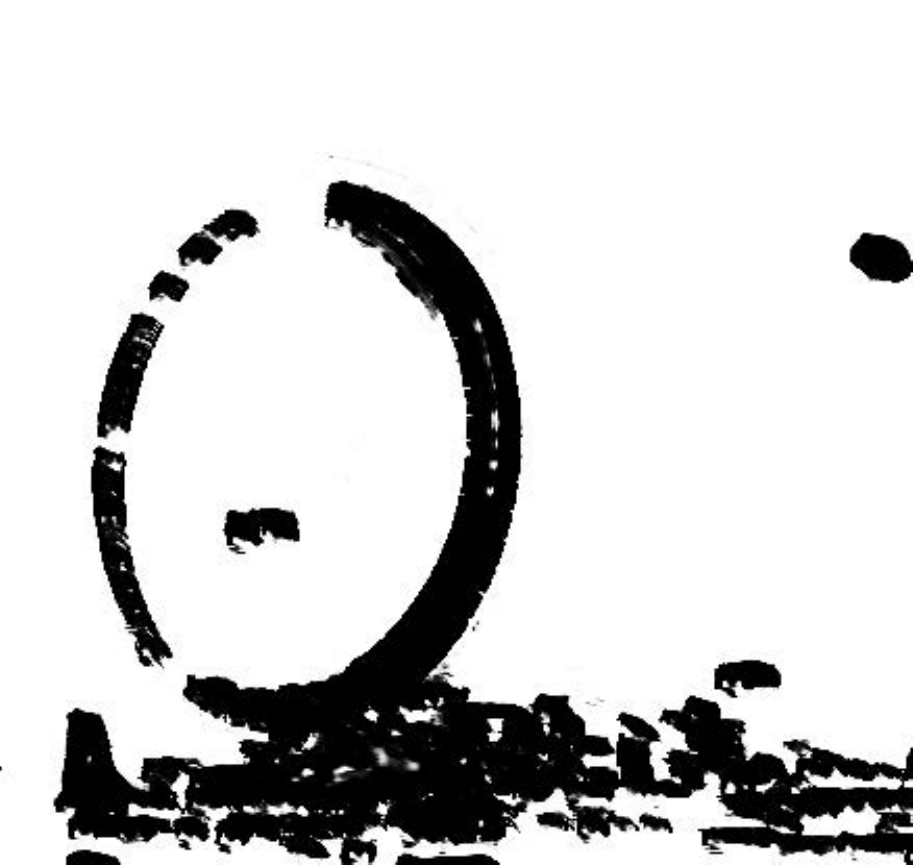}}
		\centerline{(c) Saturated pixels in~\cite{chen2021learning}}
	\end{minipage}
	\begin{minipage}[b]{0.195\linewidth}
		\centering
		\centerline{
			\includegraphics[width =\linewidth]{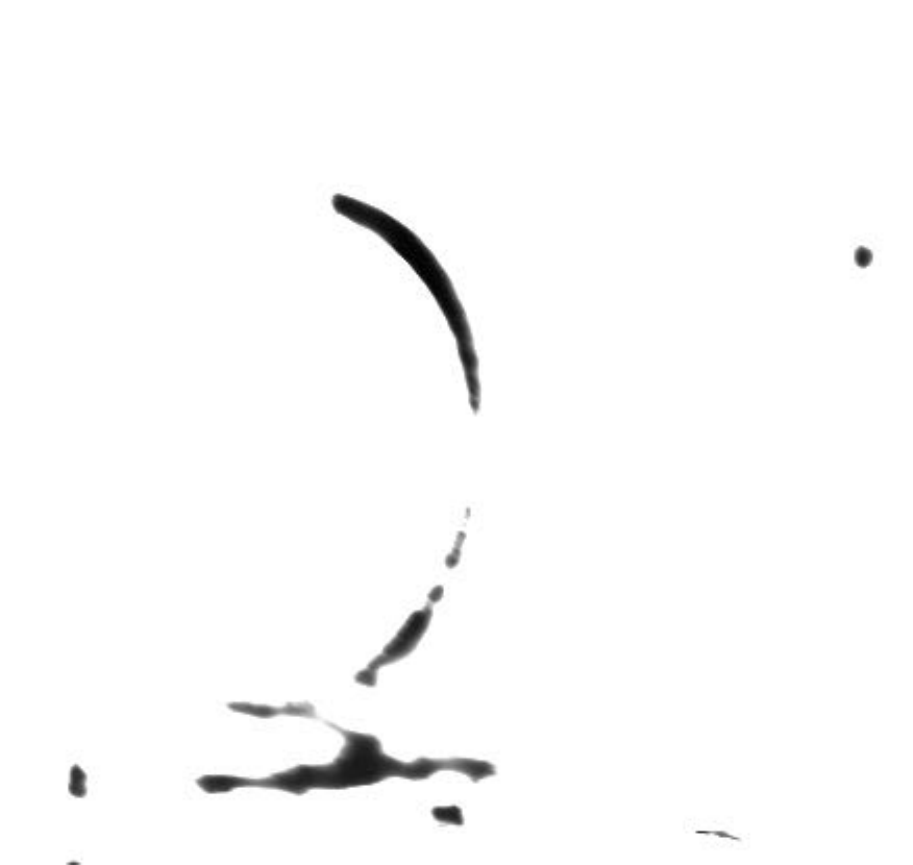}}
		\centerline{(d) $M$ from model in~\cite{chen2021blind}}
	\end{minipage}
	\begin{minipage}[b]{0.195\linewidth}
		\centering
		\centerline{
			\includegraphics[width =\linewidth]{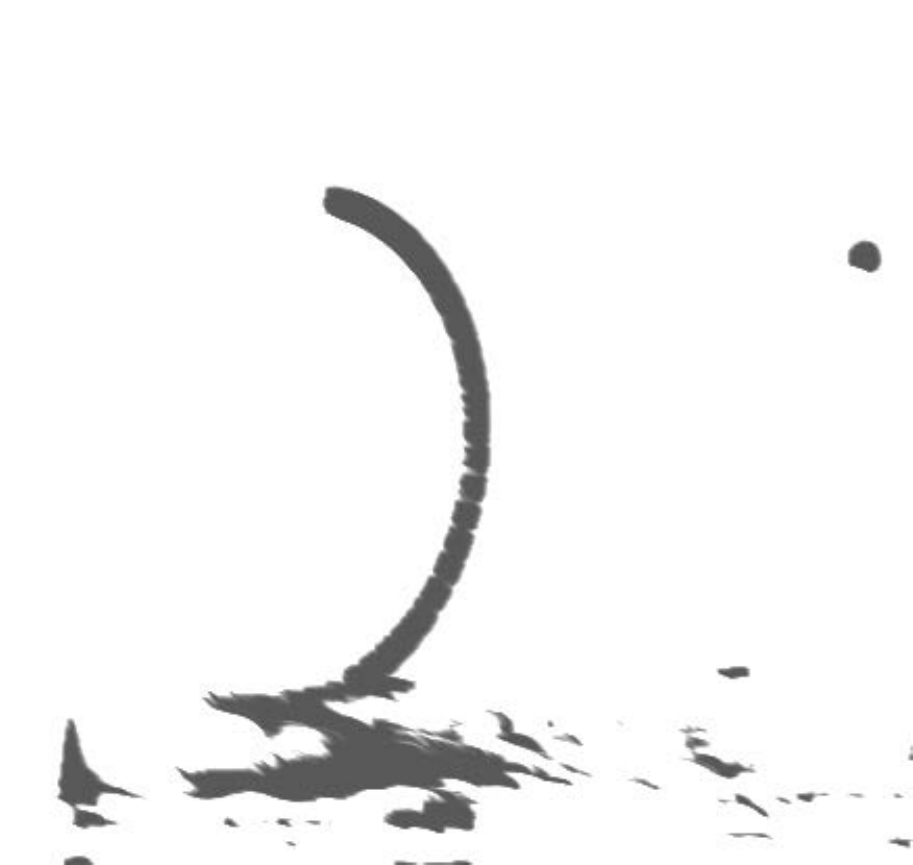}}
		\centerline{(e) Our latent map $M$}
	\end{minipage}
	\\
	\begin{minipage}[b]{0.195\linewidth}
		\centering
		\centerline{
			\includegraphics[width =\linewidth]{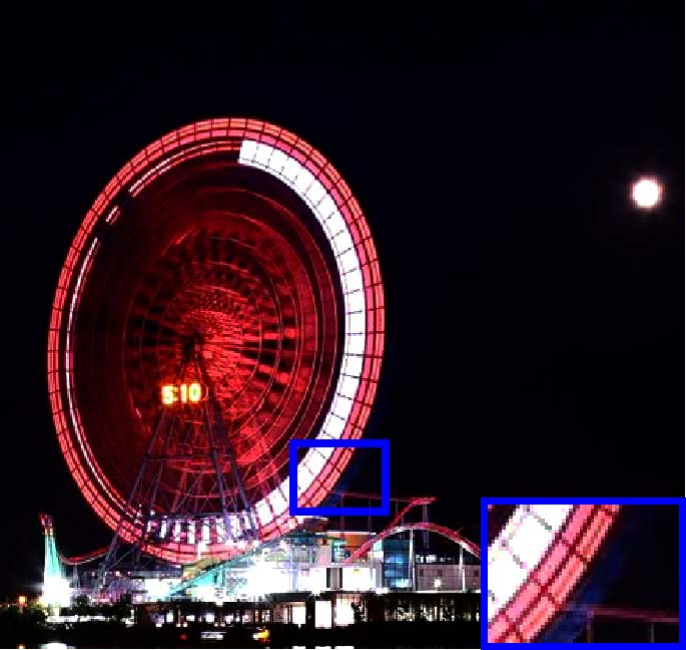}}
		\centerline{(f) GT}
	\end{minipage}
	\begin{minipage}[b]{0.195\linewidth}
		\centering
		\centerline{
			\includegraphics[width =\linewidth]{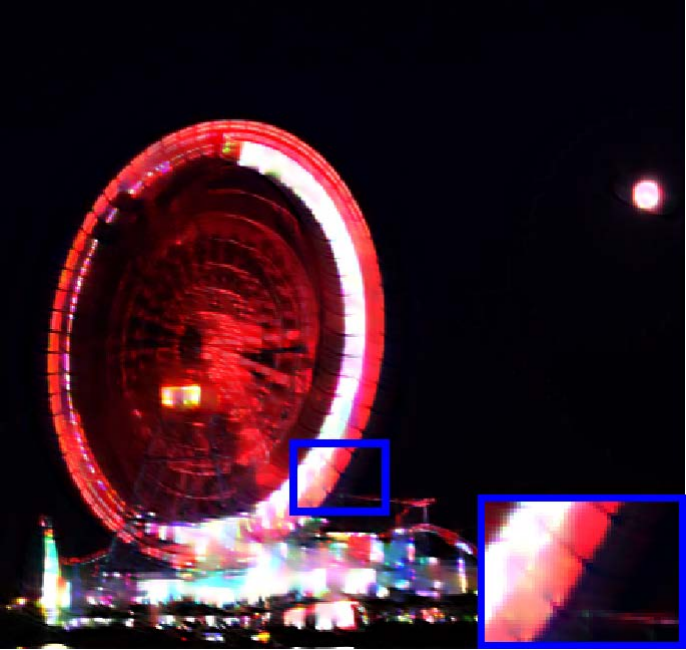}}
		\centerline{(g) Ours w/ model in~\cite{Whyte14deblurring}}
	\end{minipage}
	\begin{minipage}[b]{0.195\linewidth}
		\centering
		\centerline{
			\includegraphics[width =\linewidth]{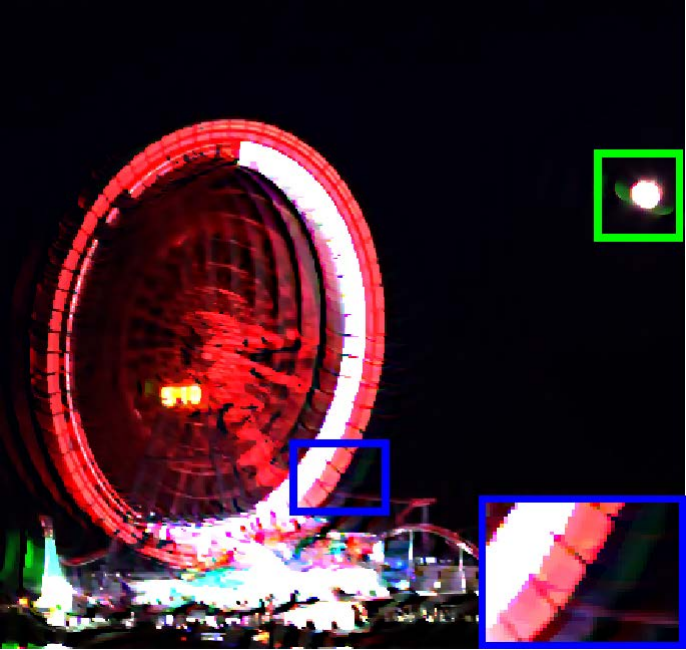}}
		\centerline{(h) Ours w/ model in~\cite{chen2021learning}}
	\end{minipage}
	\begin{minipage}[b]{0.195\linewidth}
		\centering
		\centerline{
			\includegraphics[width =\linewidth]{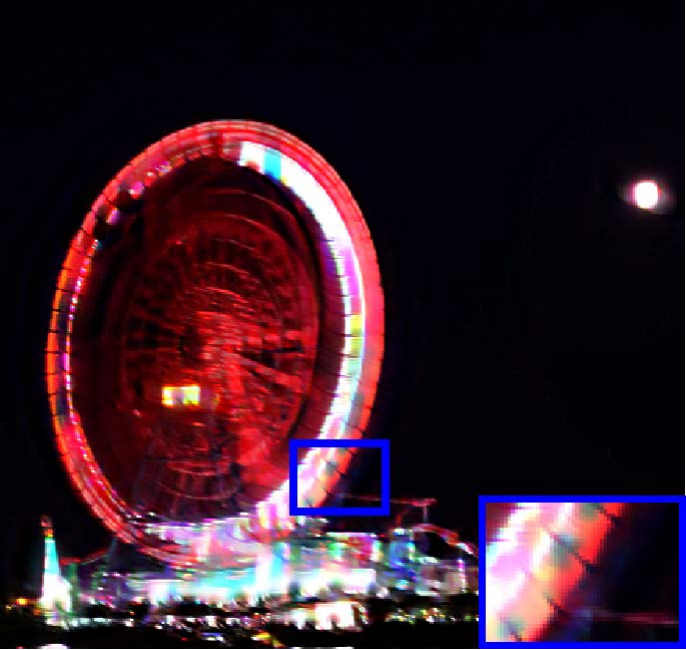}}
		\centerline{(i) Ours w/ model in~\cite{chen2021blind}}
	\end{minipage}
	\begin{minipage}[b]{0.195\linewidth}
		\centering
		\centerline{
			\includegraphics[width =\linewidth]{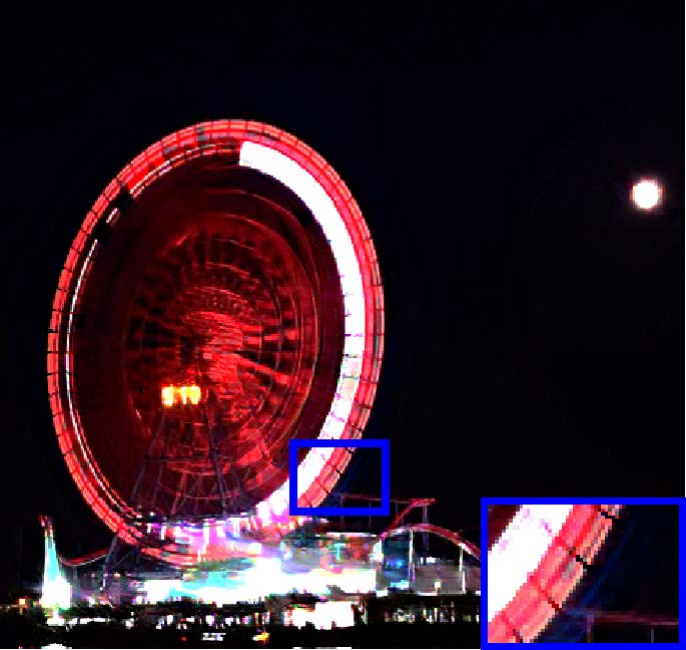}}
		\centerline{(j) Ours}
	\end{minipage}
	
	\caption{Comparisons between different blur models. Artifacts are presented in the blue and green boxes in (b), (g) - (i), while the result from our model contains fewer artifacts and is close to the GT image.
(c) is the detected saturated pixels from the blur model in NBDN~\cite{chen2021learning} which includes almost all informative pixels in the blurry image.
Darker pixels indicate smaller values in (c) - (e), where (d) and (e) are processed with the same gamma correction for better viewings. 
}
	\label{fig mapresults}
\end{figure*}

\subsection{Effectiveness of the Proposed Blur Model}
\label{sec effec}
\subsubsection{Compared to the Commonly-Used Blur Model.}
Different from other methods, we use a latent map $M$ to explicitly handle saturated pixels in the blurry image.
$M$ can be similarly considered as a replacement of the ideal clipping function in Eq.~\eqref{eq blur_clip}.
When the latent map is disabled (\eg setting $M=\textbf{1})$), our model reduces to the blur model in Eq.~\eqref{eq blur}, which is widely-adopted in many existing methods~\cite{Zhang_2017_CVPR,kaiZhang_2017_CVPR,gong2018learning}.
However, the restored image without the latent map often contains many artifacts in the saturated regions, as depicted in the green boxes of Fig.~\ref{fig mapresults}~(b).
For the learned latent map $M$ in Fig.~\ref{fig mapresults}~(d), we can observe that it has smaller values in the saturated regions and vice versa for unsaturated pixels.
With the help of the latent map, our method can generate a result with fewer ringings around the saturated regions, as shown in Fig.~\ref{fig mapresults}~(h).
To quantitatively compare our method with the strategy without the latent map, we conduct an ablation study on the proposed testing set. The results in Table~\ref{tab 3} show that the blur model with the proposed latent map can improve deblurring when saturated pixels are present.

\begin{figure*}
\small
\centering
	\begin{minipage}[b]{0.16\linewidth}
		\centering
		\centerline{
			\includegraphics[width =\linewidth]{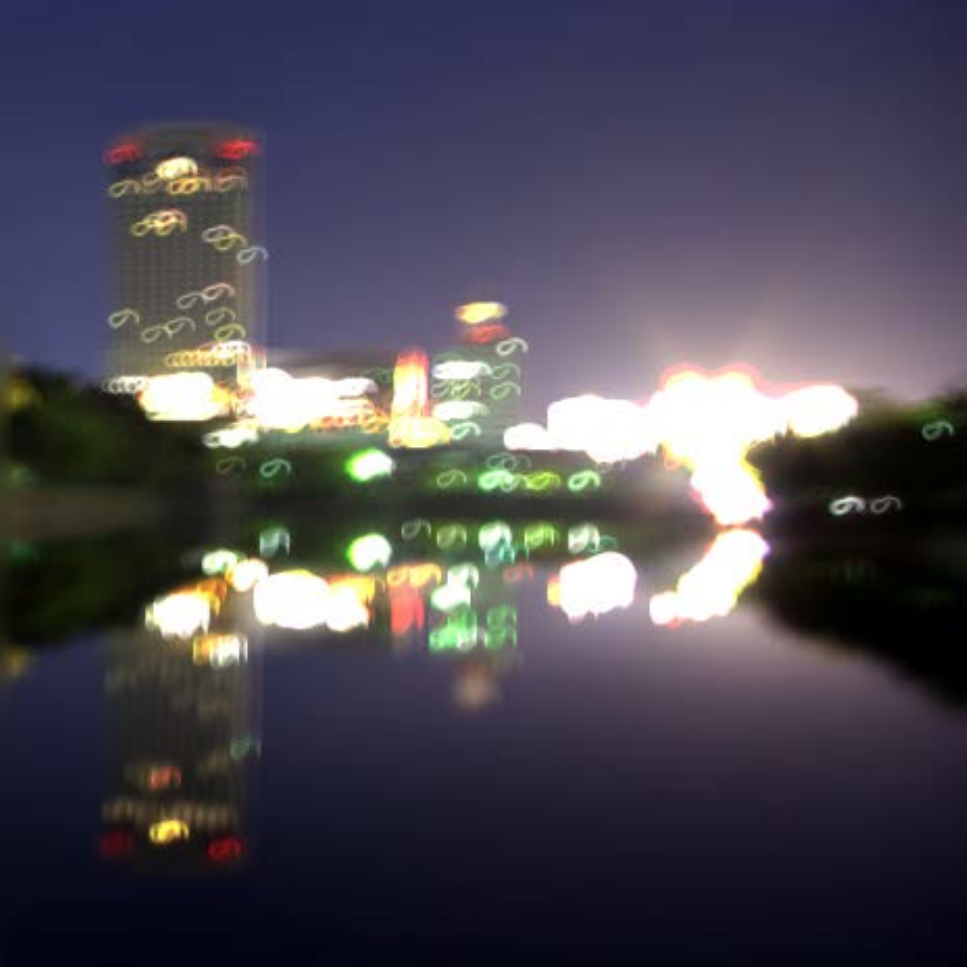}}
		\centerline{(a) Blurry image}
	\end{minipage}
	\begin{minipage}[b]{0.16\linewidth}
		\centering
		\centerline{
			\includegraphics[width =\linewidth, height=2.80 cm]{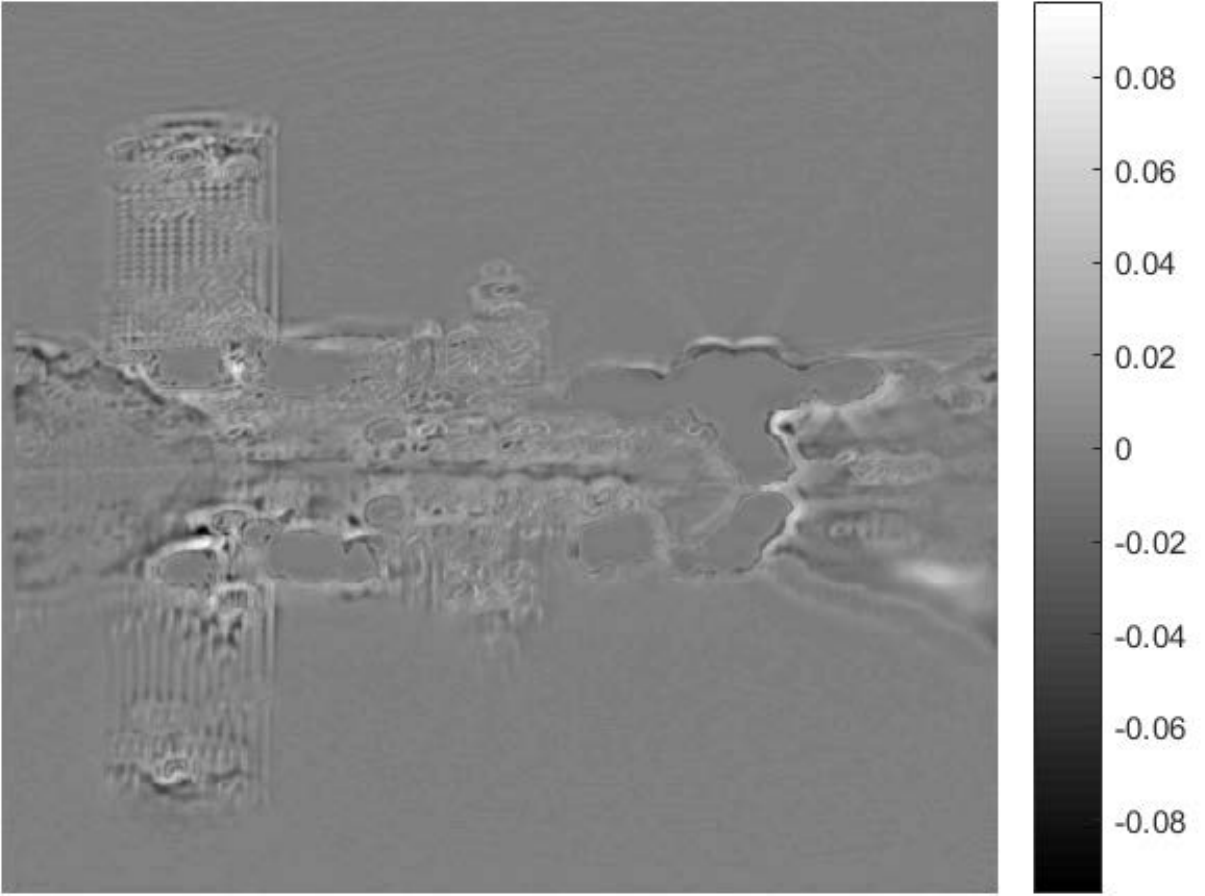}}
		\centerline{(b) $\lambda p'(I^{9})$}
	\end{minipage}
	\begin{minipage}[b]{0.16\linewidth}
		\centering
		\centerline{
			\includegraphics[width =\linewidth, height=2.8 cm]{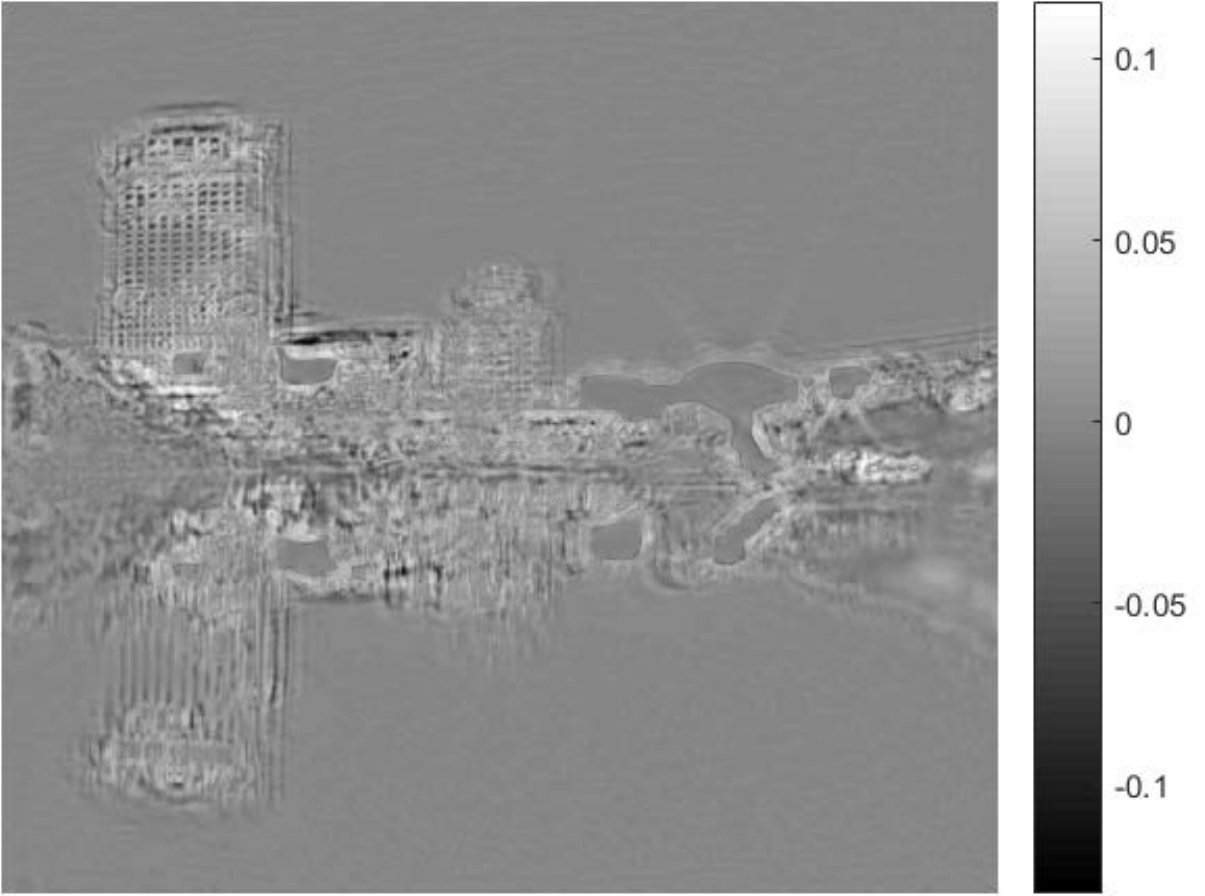}}
		\centerline{(c) $\lambda p'(I^{19})$}
	\end{minipage}
	\begin{minipage}[b]{0.16\linewidth}
		\centering
		\centerline{
			\includegraphics[width =\linewidth, height=2.8 cm]{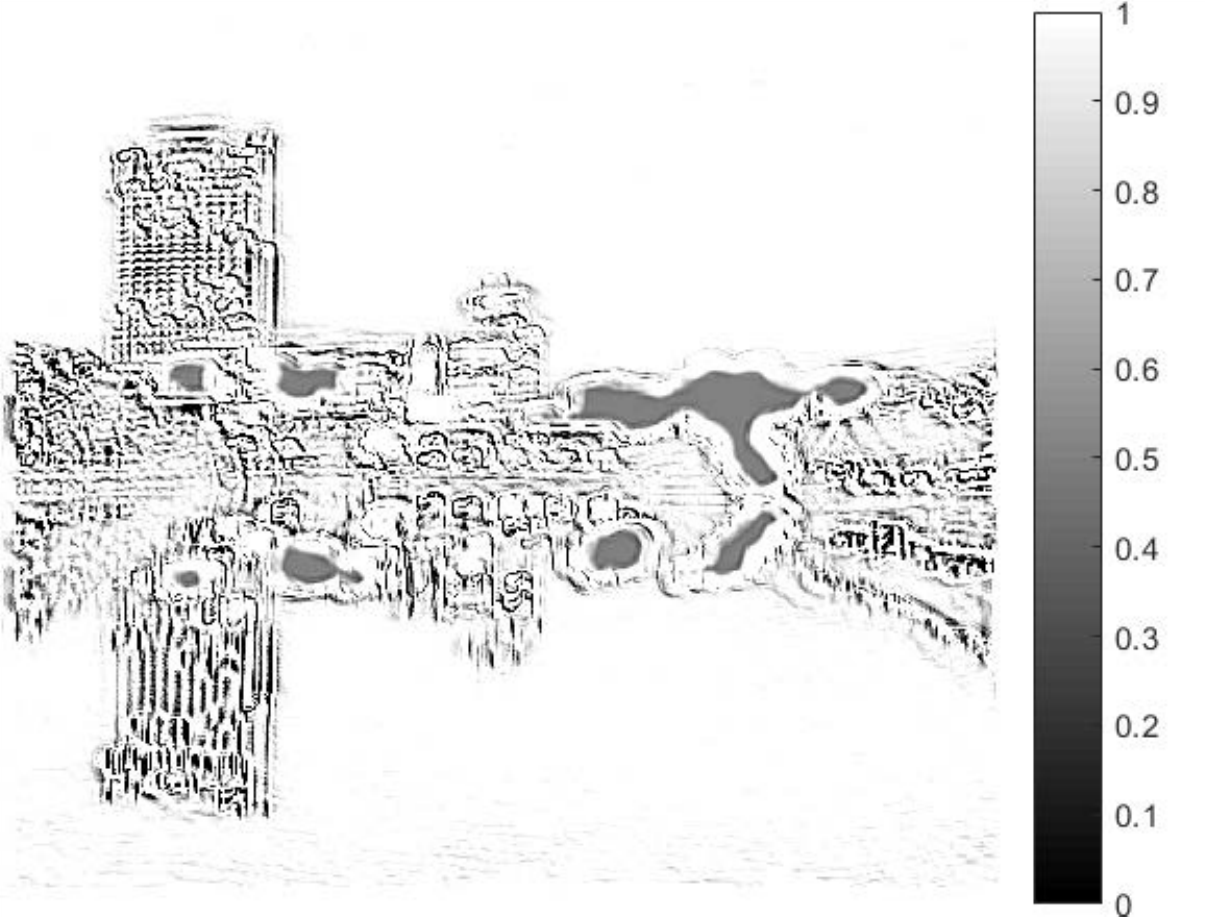}}
		\centerline{(d) $M$ in 9th iteration}
	\end{minipage}
	\begin{minipage}[b]{0.16\linewidth}
		\centering
		\centerline{
			\includegraphics[width =\linewidth, height=2.8 cm]{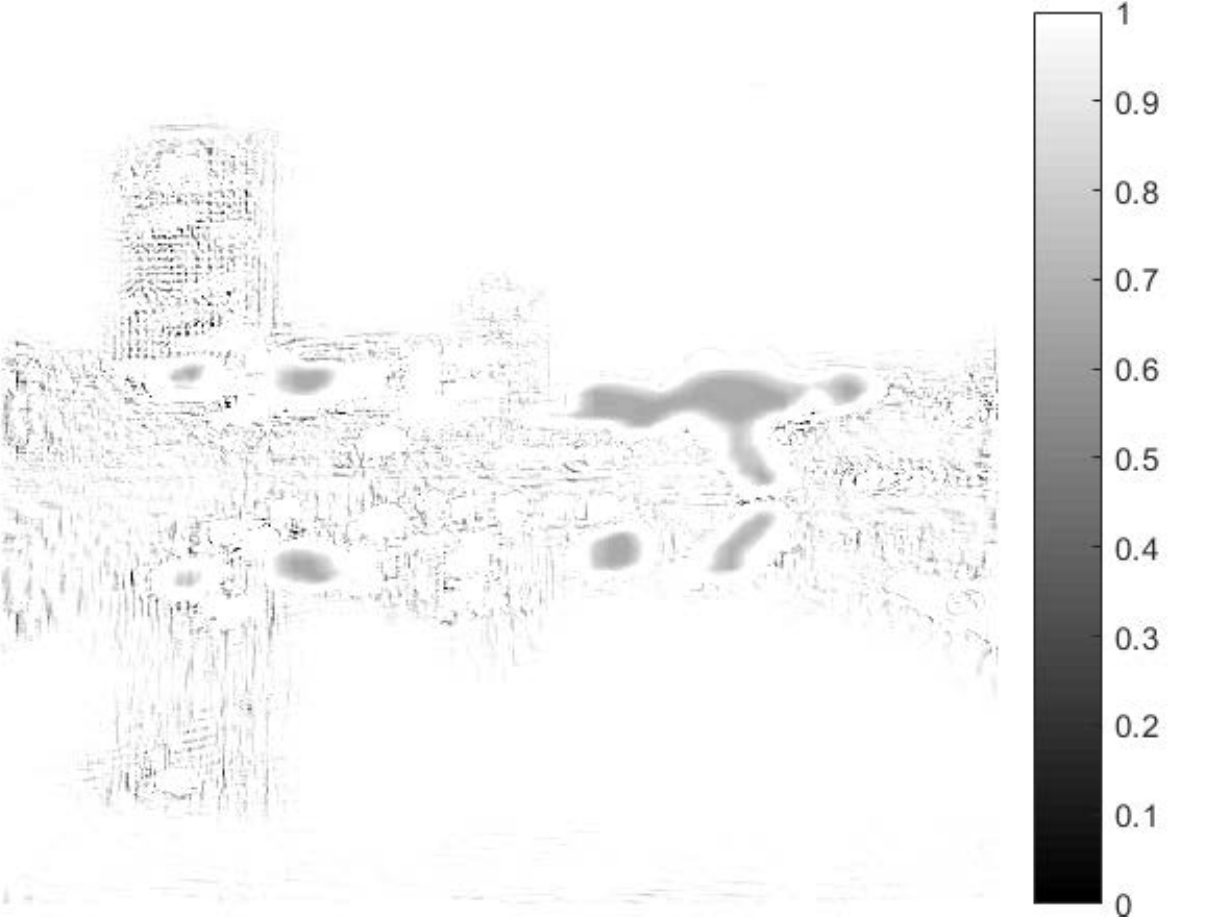}}
		\centerline{(e) $M$ in 19th iteration}
	\end{minipage}
	\begin{minipage}[b]{0.16\linewidth}
		\centering
		\centerline{
			\includegraphics[width =\linewidth, height=2.8 cm]{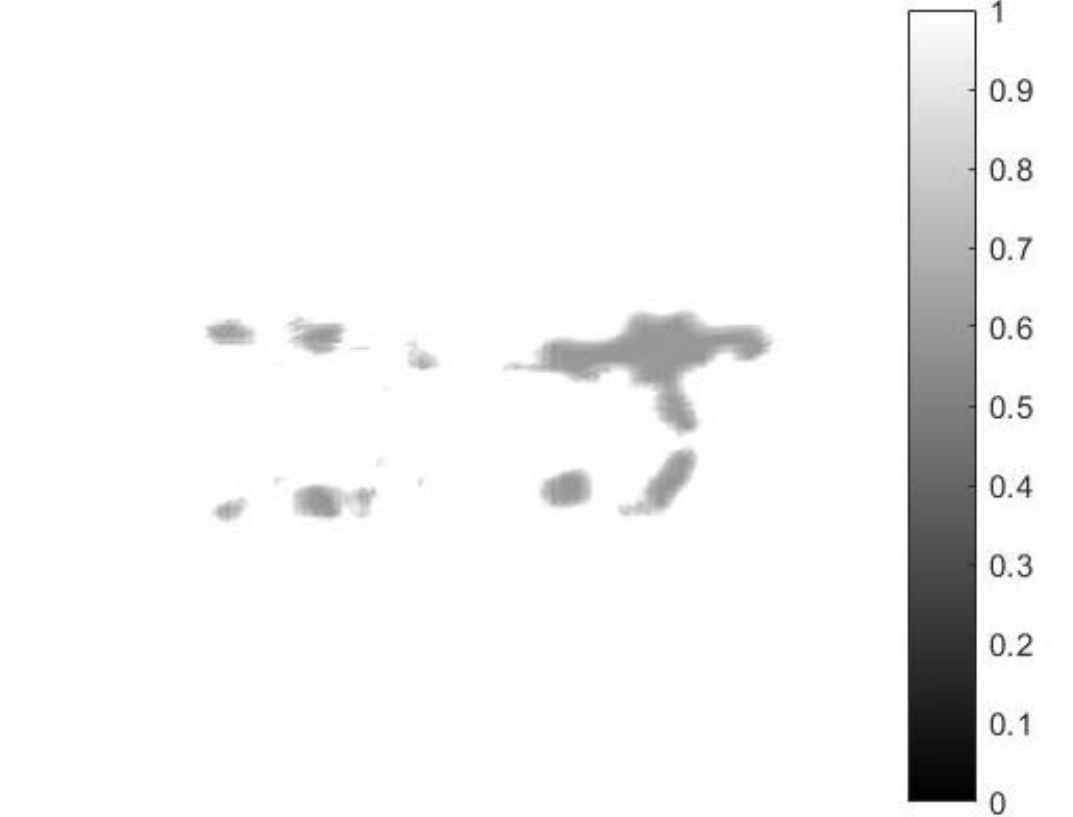}}
		\centerline{(f) GT $M$}
	\end{minipage}\\
	\begin{minipage}[b]{0.16\linewidth}
		\centering
		\centerline{
			\includegraphics[width =\linewidth]{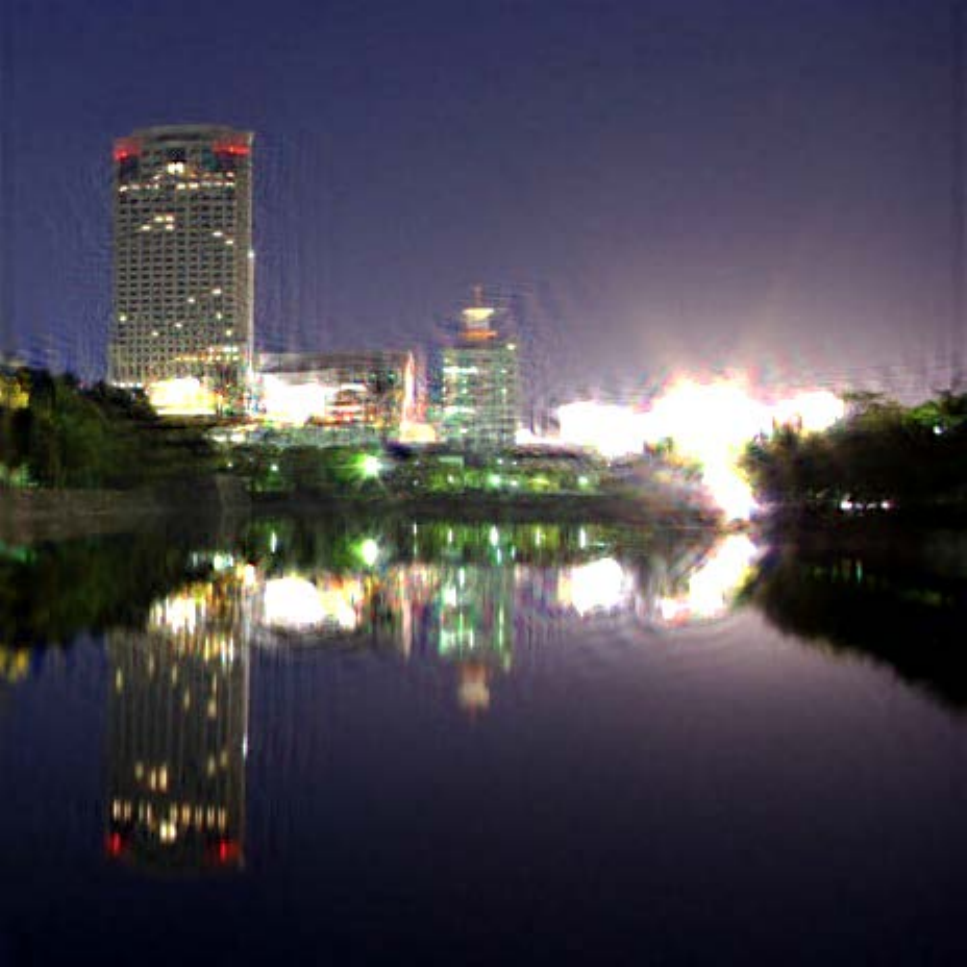}}
		\centerline{(g) Ours w/o PEN}
	\end{minipage}
	\begin{minipage}[b]{0.16\linewidth}
		\centering
		\centerline{
			\includegraphics[width =\linewidth]{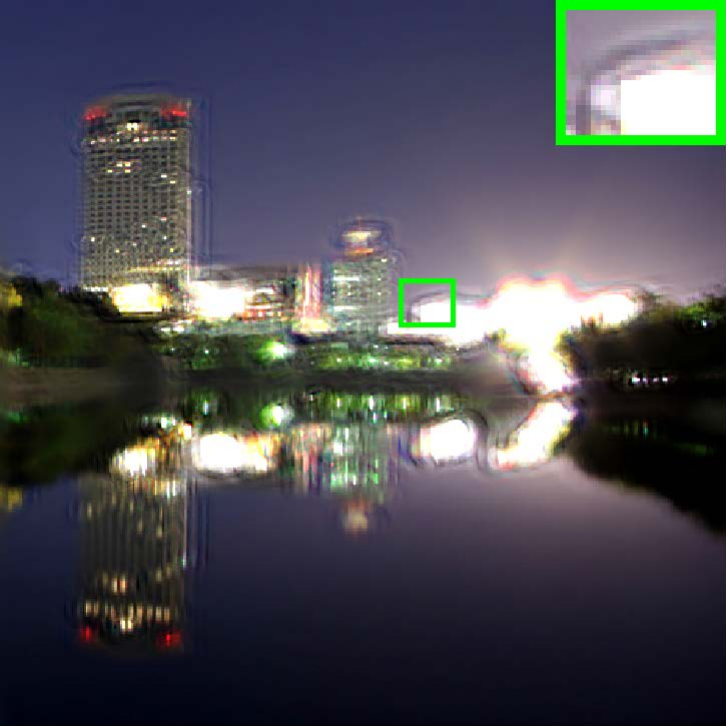}}
		\centerline{(h) $\bar I^{10}$}
	\end{minipage}
	\begin{minipage}[b]{0.16\linewidth}
		\centering
		\centerline{
			\includegraphics[width =\linewidth]{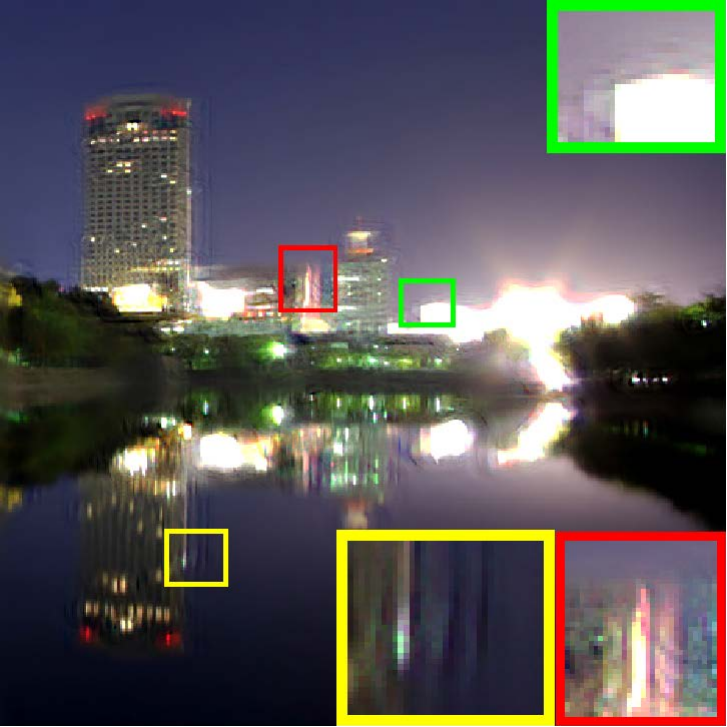}}
		\centerline{(i) $I^{10}$ }
	\end{minipage}
	\begin{minipage}[b]{0.16\linewidth}
		\centering
		\centerline{
			\includegraphics[width =\linewidth]{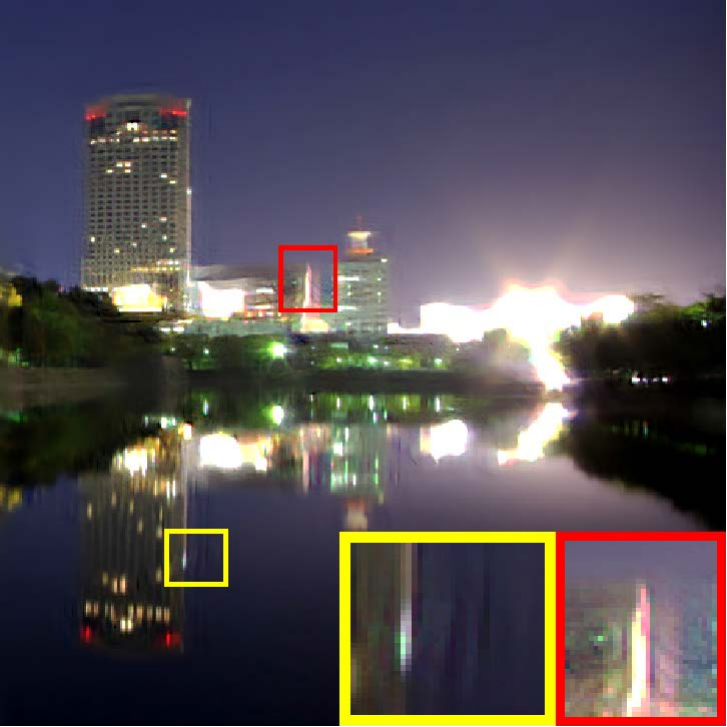}}
		\centerline{(j) $I^{20}$}
	\end{minipage}
	\begin{minipage}[b]{0.16\linewidth}
		\centering
		\centerline{
			\includegraphics[width =\linewidth]{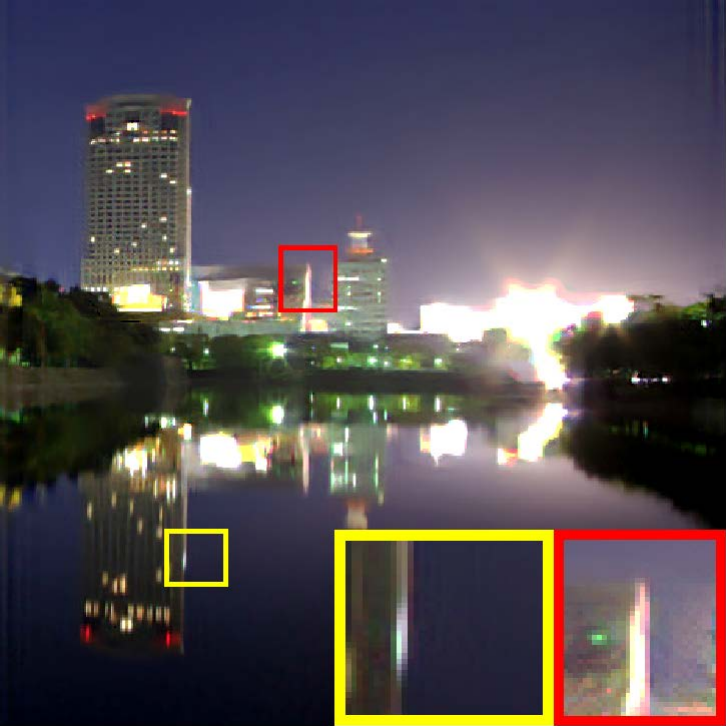}}
		\centerline{(k) Ours}
	\end{minipage}
	\begin{minipage}[b]{0.16\linewidth}
		\centering
		\centerline{
			\includegraphics[width =\linewidth]{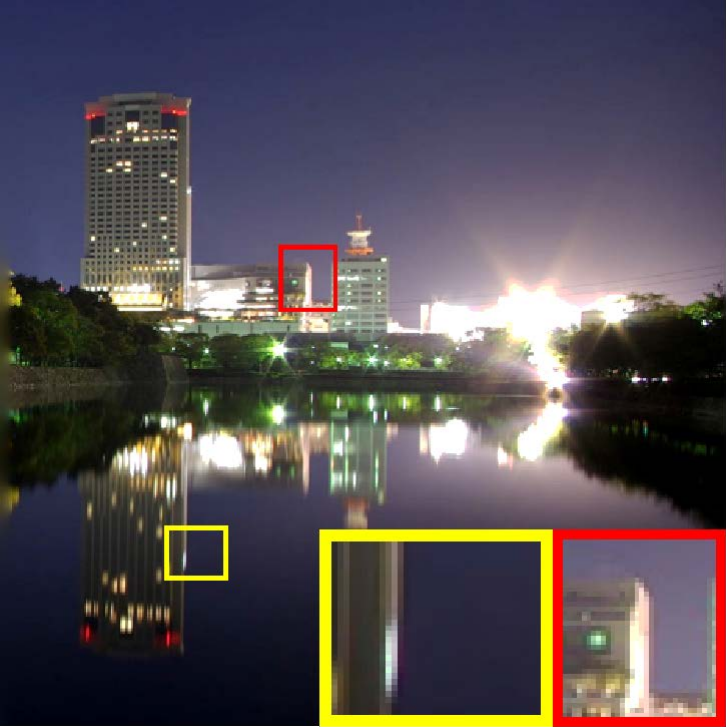}}
		\centerline{(l) GT}
	\end{minipage}
	\caption{Visualization of the output of PEN (\ie, $\lambda p'(I)$) and MEN (\ie $M$), where $I^{9}$, $I^{10}$, $I^{19}$, and $I^{20}$ denote the intermediate results at the 9th, 10th, 19th, and 20th iterations, respectively. Parts enclosed in the red and yellow boxes show that our model restores better results over iterations.
(i) is obtained by $\frac{\bar I^{t+1}}{\textbf{1}+\lambda p'(I^t)}$, and $\bar I^{t+1}$ is defined in Eq.~\eqref{eq barI}. Our result with PEN is more visually pleasing than that without it (\ie (j) vs. (g)), and intermediate results applied by PEN contain fewer ringings and noise than that before it (\ie green boxes in (i) vs. (h)).
Please zoom-in for a better view.
}
	\label{fig pen}
\end{figure*}

\subsubsection{Compared with the Blur Model by~\cite{Whyte14deblurring}.}
\label{sec whyte}
We also compare the proposed blur model with that from~\cite{Whyte14deblurring}. To solve the saturated deblurring problem,
Whyte~\etal designed the blur model based on Eq.~\eqref{eq blur_clip}. They use a specially-designed function~\cite{chen1996class} to approximate the clipping function.
However, their approximation function often requires heuristic parameters, and it is also cumbersome work to find the best parameters for different blurry images.
Differently, the clipping function is replaced by a learnable latent map in our model, which can be effectively estimated by a map estimation network (MEN).
We conduct an ablation study on the proposed testing set \wrt these two blur models.
To ensure a fair comparison between the proposed latent map-based model and the blur model from~\cite{Whyte14deblurring}, we train our method by replacing the latent map with the approximation function in~\cite{Whyte14deblurring} with their default parameter setting.
Evaluation results in Table \ref{tab 3} illustrate the effectiveness of the proposed model over that from~\cite{Whyte14deblurring}.
An example in Fig.~\ref{fig mapresults} (g) shows the limitation of the model in~\cite{Whyte14deblurring}, where the corresponding result contains artifacts and residual blur around the saturated pixels in the blue box.
In comparison, without heuristic settings, the saturated pixels can be well considered in the latent map (Fig.~\ref{fig mapresults} (d)) of our blur model. As a result, a higher quality result with fewer artifacts can be obtained as shown in Fig.~\ref{fig mapresults} (j).

\subsubsection{Compared with the Blur Model by NBDN~\cite{chen2021learning}.}
\label{sec nbdn}
Another blur model is from NBDN~\cite{chen2021learning} where Chen \etal propose to explicitly discard saturated regions by assigning small weights to the saturated pixels so that they are not involved in the deblurring process, and they further propose to use CNN to better identify the saturated regions \footnote{Taking no account of noise, the blur model in NBDN~\cite{chen2021learning} can be formulated as $\tilde{M}\circ B=\tilde{M}\circ(I\otimes K)$ s.t. $\tilde{M}\in\{0,1\}$, where pixels violate the linear blur model are assigned with small weights to make sure that they do not involve in the deblurring process. In our model, the blur model is $B=M\circ(I\otimes K)$ s.t. $M\in\left[0,1\right]$, and all pixels are considered during deblurring.}. This idea is first proposed in~\cite{cho2011outlier}. In their settings, the saturated pixels are located based on the residual of the blurry image and the convolution output of the latent image and the blur kernel (\ie $B-I \otimes K$).

However, this strategy has some intrinsic limitations. First, it is difficult to precisely separate saturated and unsaturated pixels using this model. The example shown in Fig.~\ref{fig mapresults} (c) suggests that some informative pixels (\ie the sharp edges around saturated regions), will be considered saturated and discarded during the deblurring process.
Second, because saturated regions are not involved in the deblurring process, artifacts around these regions often remain in the recovered result as shown in Fig.~\ref{fig mapresults} (h).
In comparison, our method does not require a saturated pixel discarding process because all pixels can be modeled by our blur model in Eq.~\eqref{eq our}, thus avoiding the limitations of this strategy.
To ensure a fair comparison, we reimplement our method with the blur model in NBDN~\cite{chen2021learning} and train it with the proposed training data.
Results presented in Table \ref{tab 3} further validate the effectiveness of our model against that from NBDN~\cite{chen2021learning}.

\subsubsection{Compared with the Blur Model by~\cite{chen2021blind}.}
\label{sec chen}
To comprehensively evaluate the effectiveness of the proposed model, we compare it with that from Chen \etal~\cite{chen2021blind}.
The method in~\cite{chen2021blind} is mainly designed for the blind deblurring task.
Although the same latent map is proposed to replace the clipping function, the latent map $M$ in their method is computed via a naive strategy: assuming a uniform thresh value $v$ in the image, for every pixel location $i$, $M_i=1$ if $(I\otimes K)_i<v$, or $M_i=v/(I\otimes K)_i$ if $(I\otimes K)_i>v$.
To compare with this model, we finetune our model using the map estimation strategy from~\cite{chen2021blind} and evaluate it using the proposed testing set.
Results are shown in Table \ref{tab 3} where our model outperforms that from~\cite{chen2021blind}.
The main reason is that the strategy in~\cite{chen2021blind} assumes the same thresh value among all blurry images (\ie they empirically use $v=0.9$ for all blurry samples). This strategy can work in their blind deblurring task where salient edges of the latent image are the most crucial part for estimating the blur kernel, and it does not require an accurate estimated latent image necessarily. However, their strategy may be inapplicable in the non-blind deblurring task, because the same thresh value may not be the best choice for different images.
Consequently, their computed latent map may be inaccurate in some examples. As shown in Fig.~\ref{fig mapresults} (d), the latent map from~\cite{chen2021blind} are not concordant with the saturated regions in the given example, and their deblurred result contain moderate artifacts (Fig.~\ref{fig mapresults} (i)). Compared to their model, our latent map $M$ is computed via a CNN, thus can avoid the naive settings and is suitable for most scenarios, and the recovered result is also shaper with fewer ringings (Fig.~\ref{fig mapresults} (j)).

\begin{table}[t]
    \centering
    \vspace{1mm}
    \caption{Comparisons on the given testing set with different image priors.} 
    \scalebox{1}{
    \begin{tabular}{c|ccc}\hline
      && PSNR & SSIM \\
      \hline
      \multirow{3}*{\shortstack{GT\\ blur\\ kernels}}
      & Without prior &23.88 &0.8322\\
      & hyper-Laplacian prior &24.30  & 0.8411\\
      & prior from PEN & 25.66 & 0.8595\\
      \hline
      \multirow{3}*{\shortstack{kernels\\ from\\~\cite{hu2014deblurring}}}
      & Without prior &23.34 &0.8156\\
      & hyper-Laplacian prior &23.79  &0.8202 \\
      & prior from PEN & 25.11 & 0.8367\\
      \hline
      \end{tabular}}
      \label{tab 4}
      \vspace{-0.3 cm}
\end{table}

\begin{table}
\centering
    \caption{Evaluations of the accuracies of the outputs of MEN and PEN on the proposed testing set.}
    \centering
    \vspace{1mm}
    \scalebox{1}{
    \begin{tabular}{cccc}
    \midrule
    & \multicolumn{3}{c}{Accuracy of $M$} \\
    \midrule
    iterations & 10 & 20 & 30\\
    \midrule
    MSE of $M$ & 0.0123 & 0.0091 & 0.0077\\
    \midrule
    &\multicolumn{3}{c}{Accuracy of $\lambda P'(I)$}\\
    \midrule
    iterations & 10 & 20 & 30 \\
    \midrule
    SSIM of $\bar{I}$ (\ie w/o $\lambda P'(I)$) & 0.8460 & 0.8582 & 0.8591 \\
    SSIM of $I$ (\ie w/ $\lambda P'(I)$) & 0.8467 &0.8588 &0.8595\\
    \bottomrule
    \end{tabular}}
    \label{tab acc}
\end{table}

\subsection{Effectiveness of PEN}
\label{sec ana_pen}
With only the fidelity term, the deblurring process often tends to amplify noise~\cite{dey2004deconvolution}.
Meanwhile, an accurate blur kernel is often inaccessible in real life, and it will cause significant artifacts and result in undesired image structures~\cite{yuan2008progressive}.
Thus, a decent prior term is required for the deblurring task~\cite{yuan2008progressive}.

\begin{table*}
\centering
\caption{Evaluations on the proposed testing set with different levels of noises.}
    \centering
    \scalebox{0.85}{
    \begin{tabular}{ccccccccccccc}
    \toprule
    &
    & Cho~\cite{cho2011outlier}
    & Hu ~\cite{hu2014deblurring}
    & Pan~\cite{pan2016robust}
    & Dong~\cite{dong2018learning}
    & Whyte~\cite{Whyte14deblurring}
    & FCNN~\cite{Zhang_2017_CVPR}
    & DMPHN~\cite{zhang2019deep}
    & MTRNN~\cite{park2020multi}
    & CPCR~\cite{eboli2020end}
    & DWDN~\cite{dong2020deep}
    & Ours\\
    \midrule
    &\multicolumn{12}{c}{Results with GT blur kernels}\\
    \midrule
    1$\%$ & PSNR & 20.78 &23.76 & 25.02 & 22.05 &23.55 &24.55 &22.34 &19.77 &20.79 &24.82 &\textbf{25.44}\\
    noise &SSIM & 0.7554 &0.7833 &0.8351 & 0.7760 &0.7889 &0.8310 &0.7156 &0.6432 &0.7519 &0.8362 &\textbf{0.8435}\\
    \midrule
    2$\%$ &PSNR & 20.55 &23.58 & 24.31 & 21.91 &23.04 &24.37 &22.26 &19.70 &20.55 &24.55 &\textbf{25.01}\\
    noise &SSIM & 0.7137 &0.7516 &0.8087 & 0.7564 &0.6948 &0.8003 &0.7056 &0.6225 &0.7022 &0.8103 &\textbf{0.8146}\\
    \midrule
    &\multicolumn{12}{c}{Results with blur kernels from~\cite{hu2014deblurring}}\\
    \midrule
    1$\%$ &PSNR & 20.59 &23.52 & 24.09 & 21.91 &23.43 &23.17 &22.34 &19.77 &20.65 &24.47 &\textbf{24.98}\\
    noise &SSIM & 0.7477 &0.7766 &0.8188 & 0.7646 &0.7914 &0.8082 &0.7156 &0.66432 &0.7493 &0.8198 &\textbf{0.8250}\\
    \midrule
    2$\%$&PSNR & 20.42 &23.41 & 23.62 & 21.80 &23.04 &22.84 &22.26 &19.70 &20.48 &24.19  &\textbf{24.66}\\
    noise &SSIM & 0.7181 &0.7530 &0.7949 & 0.7490 &0.7131 &0.7889 &0.7056 &0.6225 &0.7171 &0.7967  &\textbf{0.8006}\\
    \bottomrule
    \end{tabular}}
    \label{tab noise}
\end{table*}

\begin{table}
\centering
    \caption{Evaluations of our model trained with different data synthesizing strategies on the samples from~\cite{rim2020real}.}
    \centering
    \vspace{1mm}
    \scalebox{1}{
    \begin{tabular}{ccc}
    \midrule
    Data synthesized methods & NBDN~\cite{chen2021learning} & Ours \\
    \midrule
    Average PSNR & 27.89 & 28.13\\
    \bottomrule
    \end{tabular}}
    \label{tab 6}
\end{table}

The proposed PEN is used to obtain prior information for the MAP model.
Fig.~\ref{fig pen}~(b) and (c) show visualizations of the outputs of PEN at the $t$-th iteration (\ie $\lambda P'(I^t)$), which serve as discriminative
weights that facilitate the image restoration.
We can observe that in regions without edges, $\lambda P'(I^t)$ have small absolute values (close to 0), and it does not influence the optimization. In the edge regions, $\lambda P'(I^t)$ have larger absolute values so they can preserve the sharp details and remove artifacts.
Meanwhile, according to the iteration strategy in Eq.~\eqref{eq rl}, the intermediate result $\bar I^{t+1}$ defined by:
\begin{equation}
\label{eq barI}
\bar I^{t+1} \triangleq I^t\circ ((\frac{B}{I^t\otimes K} - M + \textbf{1}) \otimes \widetilde K),
\end{equation}
contains more artifacts (Fig.~\ref{fig pen}~(h)) than that after applying with PEN (Fig.~\ref{fig pen}~(i)).
The example validates that our PEN can remove ringings and noises in the deblurred result.
The comparison in Fig.~\ref{fig pen}~(g) and (k) further demonstrates that using PEN generates a better deblurred result.
Quantitative evaluations in Table~\ref{tab 4} also show that the proposed model trained without any prior information performs less effectively than that with it.

\begin{figure}
\centering
\scriptsize
\begin{minipage}[b]{0.45\linewidth}
		\centering
		\centerline{
			\includegraphics[width =\linewidth]{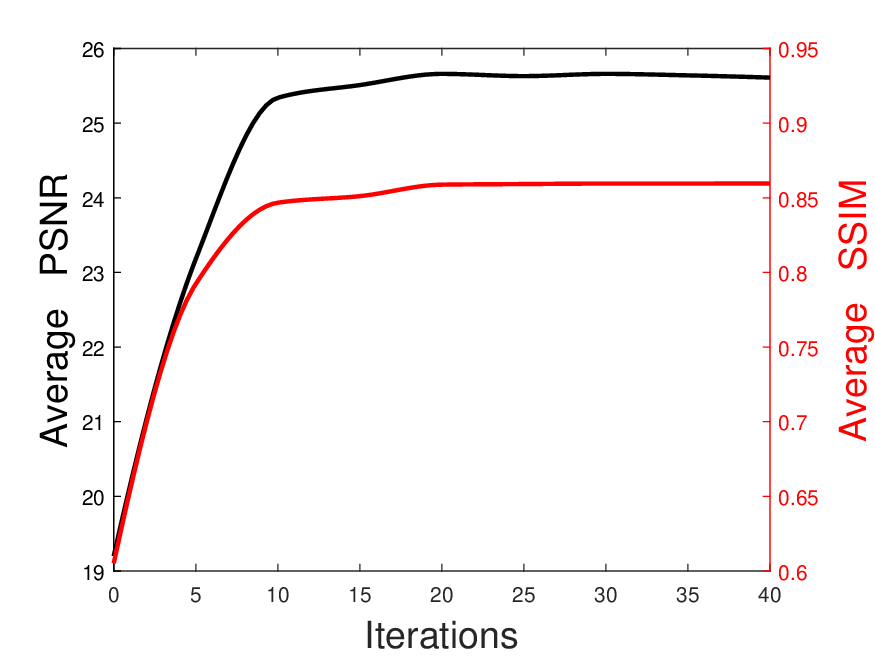}}
			\scriptsize{
			\centerline{(a) Results with GT kernels}}
\end{minipage}
\begin{minipage}[b]{0.45\linewidth}
		\centering
		\centerline{
			\includegraphics[width =\linewidth]{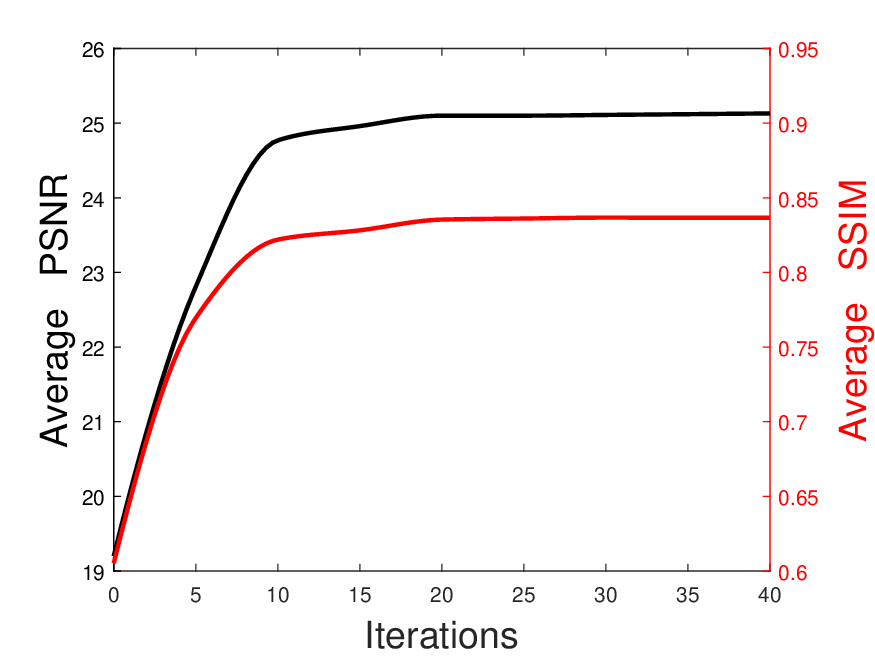}}
			\centerline{(b) Results with Kernels from~\cite{hu2014deblurring}}
\end{minipage}
\caption{Performances of the proposed method using different iteration steps.}
\label{fig convergence}
\vspace{-0.5 cm}
\end{figure}

Moreover, compared to the commonly-used hyper Laplacian prior~\cite{Levin07image}, our PEN  shows its advantages in the following aspects.
First, the hyper-Laplacian prior is effective at removing ringings.
But it can also result in fewer details.
In comparison, by learning from numerous data, the learned prior has shown its effectiveness in removing artifacts and obtaining details at the same time.
Second, the sparse prior often requires a heuristic setting for the weight parameter (\ie $\lambda$ in Eq.~\eqref{eq img_reform}), which is tedious work to manually tune the parameter for different images.
Differently, our PEN can directly learn the prior information (\ie $\lambda P'(I)$ in Eq.~\eqref{eq rl}).
To quantitatively evaluate the effectiveness of PEN, we conduct an ablation study on the proposed testing set by replacing PEN with the hyper-Laplacian prior and finetune this model.
Note the hyper-parameter $\lambda$ is set to be 0.003, which is the same as existing arts~\cite{pan2016robust,cho2011outlier}, throughout the training.
The results are shown in Table \ref{tab 4}. We note that our method with PEN performs better than that with the hyper-Laplacian prior. All these results demonstrate the effectiveness of the proposed PEN.

\begin{table*}
\centering
    \caption{Evaluations of our model with different kernel estimation methods on the samples from~\cite{rim2020real}.}
    \centering
    \scalebox{0.9}{
    \begin{tabular}{ccccccccccc}
    \midrule
    & Hu~\cite{hu2014deblurring} &Xu~\cite{xu2013unnatural} &Pan~\cite{pan2016blind} & DMPHN~\cite{zhang2019deep} &Nah~\cite{Nah17deep} &DeblurGAN~\cite{kupyn2018deblurgan} &SRN~\cite{tao2018scale} &DeblurGAN-V2~\cite{kupyn2019deblurgan} & Ours +~\cite{hu2014deblurring} &Ours +~\cite{chen2021blind} \\
    \midrule
    PSNR &26.41 &27.14 &27.22 &27.80 &27.87 &27.97 &28.56 &28.70 &28.13 &28.44\\
    SSIM &0.8028 &0.8303 &0.7901 &0.8472 &0.8274 &0.8343 &0.8674 &0.8662 &0.8497 &0.8538 \\
    \bottomrule
    \end{tabular}}
    \label{tab 7}
\end{table*}

\subsection{Accuracies of Outputs of MEN and PEN}
We first measure the estimation accuracy of MEN by computing the MSE values of $M$ in the intermediate updating steps, where the GT $M$ is computed by $\frac{B}{I\otimes K}$ with the GT $I$.
Results are shown in Table \ref{tab acc}, which shows that the network can output a high quality $M$, and it can gradually estimate better maps with more iterations.
Examples of interim results of $M$ are shown in Fig.~\ref{fig pen}. We can observe that the estimated $M$ in Fig.~\ref{fig pen} (d) and (e) are concordant with our model that saturated regions have smaller $M$ values, and $M$ is closer to the GT over iterations.

There is no ground truth label to examine the accuracy of the output of PEN (\ie $\lambda P'(I)$).
Thus, we use the SSIM value of $\bar{I}^{t+1}$ in Eq.~\eqref{eq barI} and $I^{t+1}$ in Eq.~\eqref{eq rl}, which are the intermediate result before and after considering PEN, to measure the accuracy of the output of PEN.
Results in Table \ref{tab acc} show the learned prior term can help restore better results during iterations.

\subsection{Convergence Analysis}
\label{sec convergence}
Visual examples in the red and yellow boxes of Fig.~\ref{fig pen} (i), (j), and (k) show that our method can restore better results over iterations.
To quantitatively evaluate the convergence property of our method, we conduct experiments on the proposed testing set and compute the average PSNR and SSIM values with different numbers of iteration stages.
Fig.~\ref{fig convergence} shows that our method converges well after 30 iterations and more iteration stages will not improve the results.

\begin{figure}
\centering
\begin{minipage}[b]{0.3\linewidth}
		\centering
		\centerline{
			\includegraphics[width =\linewidth]{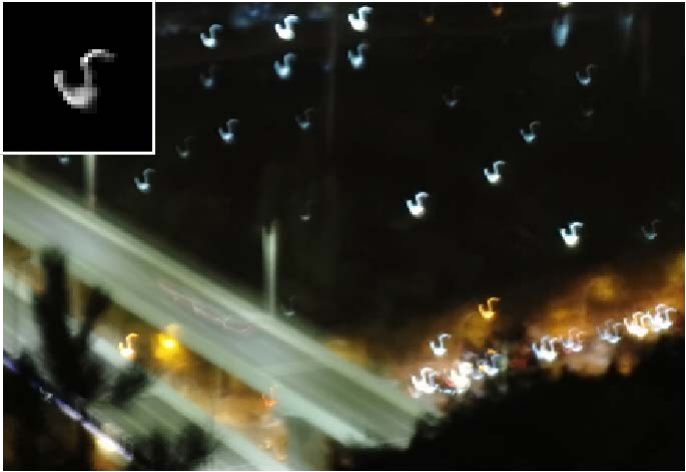}}
	\end{minipage}
	\begin{minipage}[b]{0.3\linewidth}
		\centering
		\centerline{
			\includegraphics[width =\linewidth]{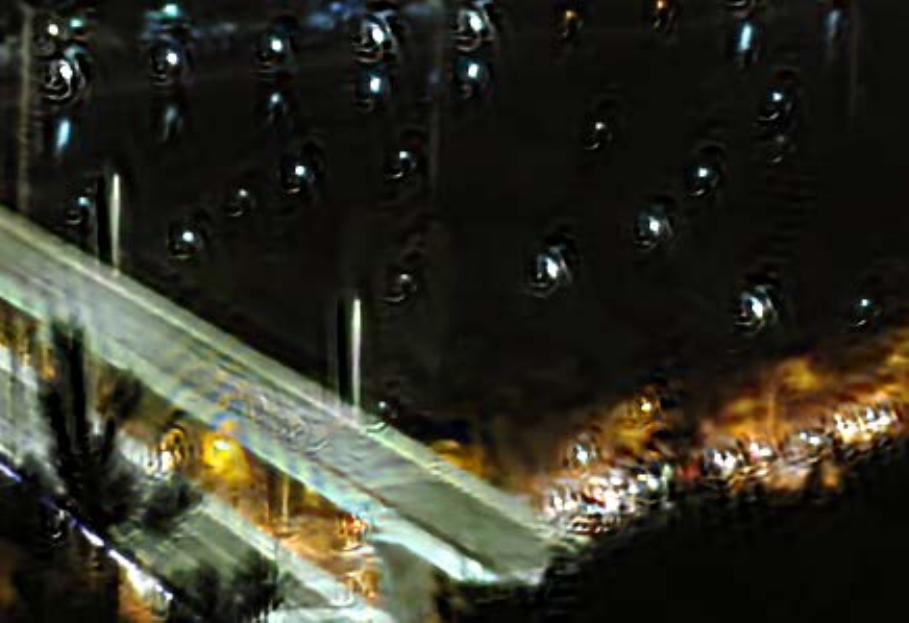}}
	\end{minipage}
	\begin{minipage}[b]{0.3\linewidth}
		\centering
		\centerline{
			\includegraphics[width =\linewidth]{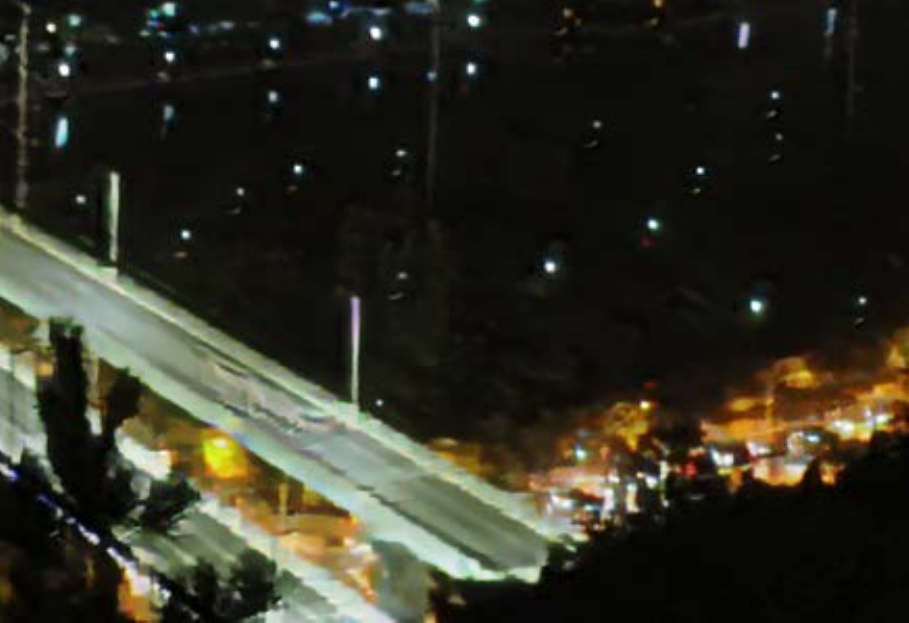}}
	\end{minipage}\\
	\vspace{0.05 cm}
	\begin{minipage}[b]{0.3\linewidth}
		\centering
		\centerline{
			\includegraphics[width =\linewidth]{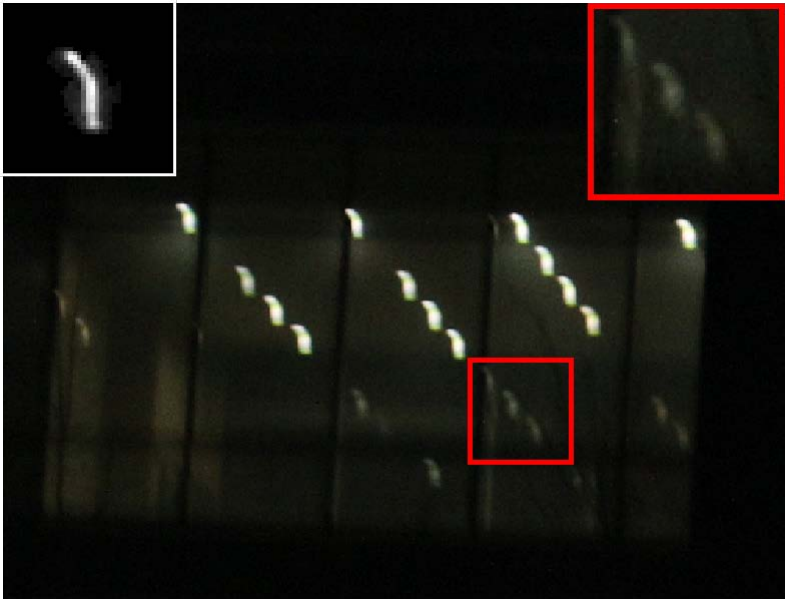}}
		\centerline{(a)}
	\end{minipage}
	\begin{minipage}[b]{0.3\linewidth}
		\centering
		\centerline{
			\includegraphics[width =\linewidth]{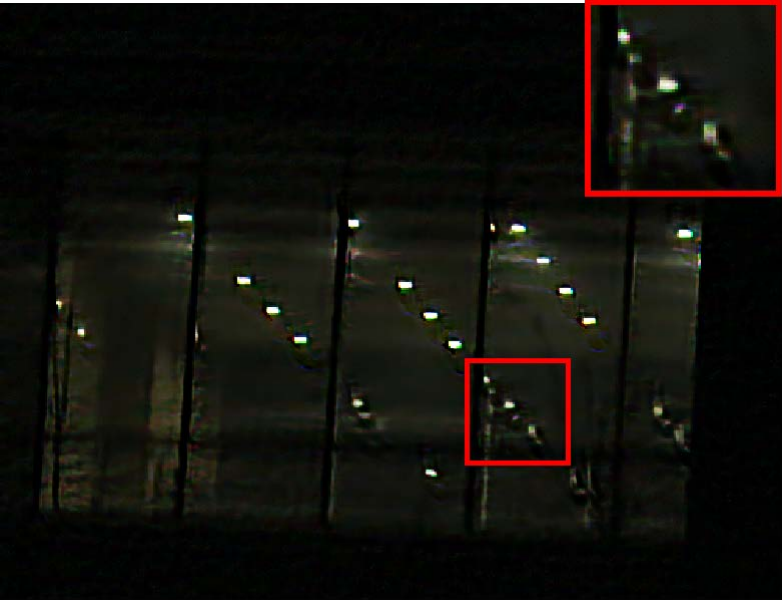}}
		\centerline{(b)}
	\end{minipage}
	\begin{minipage}[b]{0.3\linewidth}
		\centering
		\centerline{
			\includegraphics[width =\linewidth]{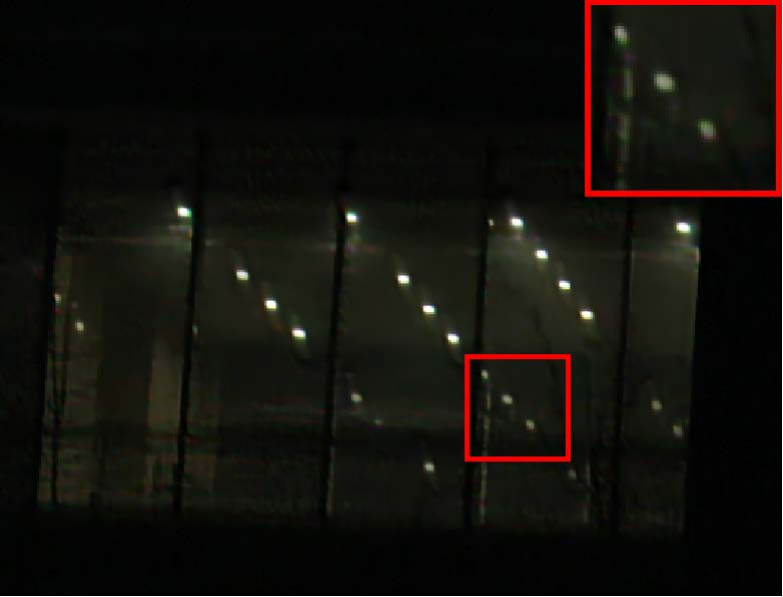}}
		\centerline{(c)}
	\end{minipage}
	\caption{Real examples recovered by our model trained with samples using different synthesizing strategies. (a) is the blurry input. (b) and (c) are outputs of our model trained with samples synthesized by~\cite{chen2021learning} and ours, respectively.}
	\label{fig diffsyn}
	\vspace{-0.4 cm}
\end{figure}

\subsection{Robustness to Noise}
Night blurry images may also contain noises. To evaluate our method on images with noises, we first train our method on
images with random Gaussian noises in a range of 0 to 3$\%$. Then, we verify the robustness of our method \wrt different levels of noise by adding 1$\%$ and 2$\%$ noises to the blurry images in the testing dataset.
Besides the optimization-based robust methods~\cite{cho2011outlier,hu2014deblurring,pan2016robust,dong2018learning,Whyte14deblurring}, we also compare our model with several leading arts that report to be robust to noise including FCNN~\cite{Zhang_2017_CVPR}, DMPHN~\cite{zhang2019deep}, MTRNN~\cite{park2020multi}, CPCR~\cite{eboli2020end}, and DWDN~\cite{dong2020deep}, where\cite{Zhang_2017_CVPR,zhang2019deep,dong2020deep} are finetuned using our training samples. Results from different methods are shown in Table \ref{tab noise}.
We note that the proposed method can generate competitive results among the compared methods, which also validates the robustness of our method to different levels of noise.

\subsection{Training Sample Synthesizing Strategy Comparison}
\label{sec syndata}
Our training sample synthesizing process is inspired by the strategy in~\cite{hu2014deblurring}, and it is different from that in~\cite{chen2021learning}.
Chen \etal suggest synthesizing blurry saturated images by enlarging the whole image with an enlarging factor before the convolving step.
Although their strategy can synthesize enough saturated pixels, enlarging the whole image will make the dynamic range of training samples smaller.
Considering that saturated images are often with a high dynamic range, we suggest only enlarging pixels above given thresh values, thus retaining or even increasing the dynamic range of the training samples.
To compare these two data synthesizing strategies, we train our model on samples synthesized with each of them respectively. The test data are uniform-blurred saturated images selected from the test set of the real-world dataset~\cite{rim2020real}, and the blur kernels are obtained from~\cite{hu2014deblurring}.
Evaluation results in Table~\ref{tab 6} demonstrate the effectiveness of our sample synthesizing strategy.
A real-world deblurring example in Fig.~\ref{fig diffsyn} also shows that our synthesizing strategy can better handle the real-world saturated blurry images.

\subsection{Evaluations with Different Kernel Estimation Methods}
\label{sec realblur}
As an important part of the overall image deblurring pipeline, kernel estimation methods can also affect the quality of the restored images. To evaluate whether our method works well when using different estimated kernels, we use two different kernel estimation methods~\cite{hu2014deblurring,chen2021blind} on the low-light blurry images (JPEG form) from the RealBlur dataset~\cite{rim2020real}.
Evaluated results are listed in Table~\ref{tab 7}. Results from the compared methods are directly cited from the official website of the RealBlur dataset \footnote{http://cg.postech.ac.kr/research/realblur/}. We observe that our method can obtain favorable performances among the state-of-the-arts with different blur kernel estimation methods. These results indicate that our method is robust to blur kernels to some extent.

\section{Conclusion}
In this paper, we propose a simple yet effective method for non-blind deblurring when saturated pixels are present in the blurry image.
To explicitly handle saturated pixels, we modify the widely-adopted linear blur model by introducing a learnable latent map to estimate it.
Based on the blur model, we formulate the deblurring task into a MAP problem and solve it by iteratively updating the latent image and map.
In specific, we develop a map estimation network (MEN) to directly estimate the latent map based on the current estimation of the latent image, and we obtain the revised latent image by conducting a Richardson-Lucy (RL)-based optimization scheme.
In addition, we develop an effective prior estimation network (PEN) to learn an image prior as the constraint of the latent image so that high-quality deblurring results can be better obtained under this learned prior.
Both quantitative and qualitative evaluations on synthetic datasets and real-world images demonstrate the effectiveness of the proposed method.

\section*{Acknowledgement}
F. Fang was supported by the National Key R\&D Program of China (2022ZD0161800), the NSFC-RGC (61961160734), and the Shanghai Rising-Star Program (21QA1402500). J. Pan was supported by the National Natural Science Foundation of China (Nos. U22B2049).

\section*{Appendix}
We present the detailed deviations for solving Eq.~\eqref{eq img_reform} in this appendix.
By reformulating Eq.~\eqref{eq img_reform} into a vectorized form, we can obtain:
\begin{equation}
\label{eq matrix}
\min_\textbf{I} \textbf{M}^\text{T}\textbf{KI} - \textbf{B}^\text{T}\log (diag(\textbf{M})\textbf{KI}) + \lambda \overline{\textbf{1}}^\text{T} P(\textbf{I})\overline{\textbf{1}},
\end{equation}
where $\textbf{M}$, $\textbf{B}$, and $\textbf{I}$ denote the vectorized forms of $M$, $B$ and $I$; $\textbf{K}$ is the Toeplitz matrix of $K$ \wrt $I$; $\overline{\textbf{1}}$ denotes a vector whose elements are all ones.
For the second term of Eq.~\eqref{eq matrix}, we denote it as $\mathbf{A}$ and its derivative \wrt $\textbf{I}$ is:
\begin{equation}
\small
\begin{aligned}
\frac{\partial \mathbf{A}}{\partial \textbf{I}}&=\frac{\partial diag(\textbf{M})\textbf{KI}}{\partial\textbf{I}}~\frac{\partial log(diag(\textbf{M})\textbf{KI})}{\partial diag(\textbf{M})\textbf{KI}}~\frac{\partial \textbf{B}^\text{T}\log (diag(\textbf{M})\textbf{KI})}{\partial log(diag(\textbf{M})\textbf{KI})},\\
&=(\textbf{K}^\text{T}diag(\textbf{M}))diag(\frac{1}{diag(\textbf{M})\textbf{KI}})\textbf{B},\\
&=\textbf{K}^\text{T}diag(\frac{1}{diag(\textbf{M})\textbf{KI}})(diag(\textbf{M})\textbf{B}),
\end{aligned}
\end{equation}
where the divide operation is element-wise.
Then we can solve Eq. \eqref{eq matrix} by setting its derivative to zero as:
\begin{equation}
\label{eq dev}
\textbf{K}^\text{T}\textbf{M} - \textbf{K}^\text{T}diag(\frac{1}{diag(\textbf{M})\textbf{KI}})(diag(\textbf{M})\textbf{B}) + \lambda P'(\textbf{I}) = 0.
\end{equation}
Reformulate the above formation into its matrix form, we have:
\begin{equation}
\label{eq 2}
M\otimes \widetilde{K} - \frac{M\circ B}{M\circ (I\otimes K)}\otimes \widetilde{K}  + \lambda P'_I(I) = 0.
\end{equation}
where $\widetilde{K}$ is the transpose of $K$ that flips the shape of $K$ upside down and left-to-right, $P'_I(I)$ is the first order derivative of $P_I(I)$ w.r.t. $I$.
Recall that the sum of the kernel equals to 1, \ie $\overline{\textbf{1}}^\text{T} \widetilde{\textbf{K}} = 1$, where $\widetilde{\textbf{K}}$ is the vectorized form of $\widetilde{K}$.
Thus, we further have,
\begin{equation}
\label{eq 3}
M\otimes \widetilde{K} - \frac{M\circ B}{M\circ (I\otimes K)}\otimes \widetilde{K}  + \lambda P'_I(I) + \textbf{1} - \textbf{1}\otimes\widetilde{K} = 0,
\end{equation}
\begin{equation}
\label{eq 3_1}
(\frac{B}{I\otimes K} - M + \textbf{1}) \otimes \widetilde{K} = \textbf{1}+\lambda P'_I(I),
\end{equation}
\begin{equation}
\label{eq 3_2}
I\circ(\frac{B}{I\otimes K} - M + \textbf{1}) \otimes \widetilde{K} = I\circ(\textbf{1}+\lambda P'_I(I)),
\end{equation}
where $\textbf{1}$ is an all-one image.

In order to solve Eq.~\eqref{eq 3_2}, we use the fixed point iteration scheme and rewrite it as:
\begin{equation}
\label{eq 4}
\frac{I^{t+1}}{I^t} =\textbf{1}= \frac{(\frac{B}{I^t\otimes K} - M + \textbf{1}) \otimes \widetilde{K}}{\textbf{1}+\lambda P'_I(I^t)}.
\end{equation}
Thus, we can finally get Eq.~\eqref{eq rl} in our manuscript:
\begin{equation}
\label{eq 5}
I^{t+1} = \frac{I^t \circ ((\frac{B}{I^t\otimes K} - M + \textbf{1}) \otimes \widetilde{K})}{\textbf{1}+\lambda P'_I(I^t)}.
\end{equation}

\bibliographystyle{plain}
{\footnotesize
\bibliography{deblur}
}

\end{document}